\documentclass[twocolumn]{svjour3}          
\smartqed  
\usepackage[usenames]{color}
\usepackage[numbers]{natbib}

\usepackage{breakurl}
\usepackage[table]{xcolor}
\usepackage{amssymb}
\usepackage{makecell}
\usepackage{multirow}
\usepackage{rotating}
\usepackage{array}
\usepackage{times}
\usepackage{graphicx}
 \usepackage{psfrag}
\usepackage{subfigure}
\usepackage{rotating}
\usepackage{url}
\usepackage{breakurl}
\usepackage{rotating, graphicx}
\usepackage{amsmath}
\usepackage{booktabs}
\usepackage{lscape}
\usepackage{verbatim}
\usepackage{geometry}
\usepackage[misc]{ifsym}
\papertype{generic article}
\definecolor{DarkBlue}{rgb}{0,0,0}

\usepackage[pagebackref=true,breaklinks=true,linkcolor=red,anchorcolor=blue, citecolor=green,letterpaper=true,colorlinks,bookmarks=false]{hyperref}
\usepackage{graphicx}

\definecolor{LightBlue}{rgb}{0.88,0.92,0.95}
\definecolor{Orange}{rgb}{1,0.75,0}

\begin{document}

\title{From BoW to CNN: Two Decades of Texture Representation for Texture Classification}
\subtitle{}

\author{Li Liu $^{1,2}$  \and
        Jie Chen $^{2}$ \and
        Paul Fieguth $^{3}$ \and \\
        Guoying Zhao $^{2}$ \and
        Rama Chellappa $^{4}$ \and
        Matti Pietik\"{a}inen $^{2}$
}

\authorrunning{Li Liu \emph{et al.}} 

\institute{ $\textrm{\Letter}$ Li Liu (li.liu@oulu.fi) \\
        Jie Chen (jie.chen@oulu.fi) \\
        Paul Fieguth (pfieguth@uwaterloo.ca) \\
        Guoying Zhao (guoying.zhao@oulu.fi) \\
        Rama Chellappa (rama@umiacs.umd.edu) \\
        Matti Pietik\"{a}inen (matti.pietikainen@oulu.fi) \\
       1 National University of Defense Technology, China \\
       2 University of Oulu, Finland \\
       3 University of Waterloo, Canada \\
       4 University of Maryland, USA
}

\date{Received: date / Accepted: date}

\maketitle

\begin{abstract}
Texture is a fundamental characteristic of many types of images, and texture representation is one of the essential and challenging problems in computer vision and pattern recognition which has attracted extensive research attention over several decades. Since 2000, texture representations based on Bag of Words (BoW) and on Convolutional Neural Networks (CNNs) have been extensively studied
with impressive performance. Given this period of remarkable evolution, this paper aims to present a comprehensive survey of advances in texture representation over the last two decades.
More than 250 major publications are cited in
this survey covering different aspects of the research, including benchmark datasets and state of the art results. In retrospect of what has been achieved so far,
the survey discusses open challenges and directions for future research.


\end{abstract}

\section{Introduction}
\label{sec:introduction}
\begin {figure*}[ht]
\centering
\includegraphics[width=0.95\textwidth]{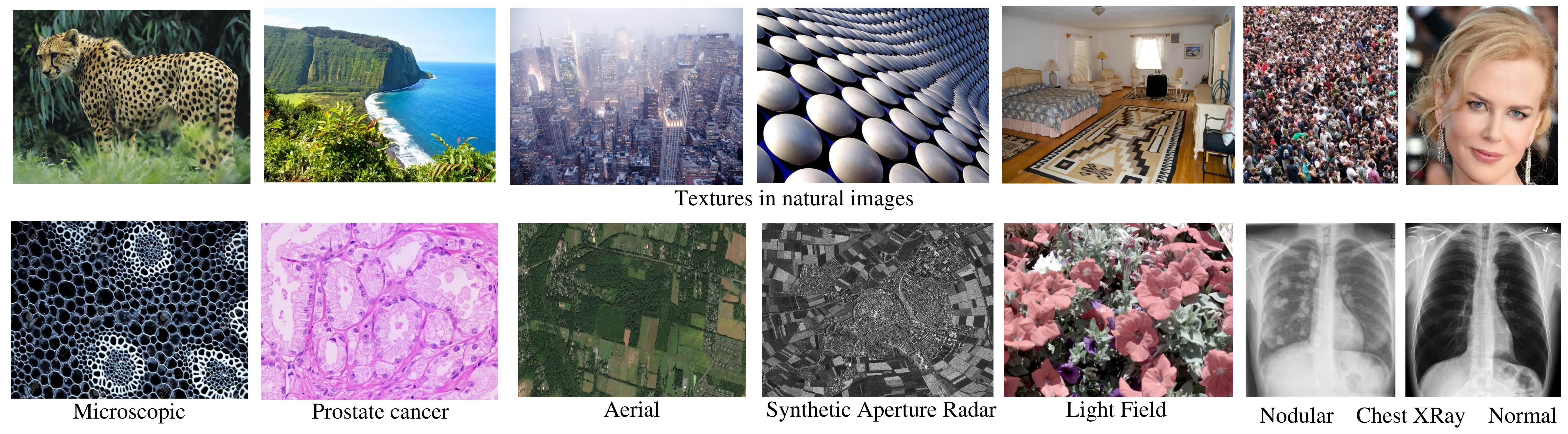}
\caption{Texture is an important characteristic of many types of images.}
\label{fig:TexturesWidelyExist}
\end {figure*}

Our visual world is richly filled with a great variety of
textures, present in images ranging from multispectral satellite data to microscopic images of tissue samples (see Fig.~\ref{fig:TexturesWidelyExist}). As a powerful visual cue, like color, texture provides useful information in identifying objects or regions of interest in images. Texture is different from color in that it refers to the spatial organization of a set of basic elements or primitives (\emph{i.e.,} textons), the fundamental microstructures in natural images and the atoms of preattentive human visual perception \cite{julesz1981textons}. A textured region will obey some statistical properties, exhibiting periodically repeated textons with some degree of variability in their appearance and relative position \cite{Forsyth2012}. Textures may range from purely stochastic to perfectly regular and everything in between (see Fig.~\ref{fig:TexturesWidelyExist}).

As a longstanding, fundamental and challenging problem in the fields of computer vision and pattern recognition, texture analysis has been a topic of intensive research since the 1960's \cite{julesz1962visual} due to its significance both in understanding how the texture perception process works in human vision as well as in the important role it plays in a wide variety of applications.
The analysis of texture traditionally embraces several problems including classification, segmentation, synthesis and shape from texture \cite{tuceryan1993texture}. Significant progress has been made since the 1990's in the first three areas, with shape from texture receiving comparatively less attention. Typical applications of texture analysis include medical image analysis \cite{BiomedicalTexture,Nanni10,Peikari2016Triaging}, quality
inspection \cite{Xie2007Texems}, content based image retrieval \cite{Manjunath1996Texture,Sivic2003,Zheng2017SIFT}, analysis of satellite or aerial imagery \cite{Kandaswamy2005Efficient,He2013Texture}, face analysis \cite{Ahonen2006Face,Ding2016Multi,Simonyan2013Fisher,Zhao07}, biometrics \cite{Ma2003Personal,Pietikainen11}, object recognition \cite{Shotton2009textonboost,Oyallon2015Deep,Zhang07}, texture synthesis for computer graphics and image compression \cite{Gatys2015Texture,Gatys2016Image}, and robot vision and autonomous navigation for
unmanned aerial vehicles. The ever-increasing amount of image and video data due to surveillance, handheld devices, medical imaging, robotics etc. offers an endless potential for further applications of texture analysis.

Texture representation, \emph{i.e.,} the extraction of features that describe texture
information, is at the core of texture analysis. After over five decades of continuous research, many kinds of theories and algorithms have emerged, with major surveys and some representative work as follows. The majority of texture features
 before 1990 can be found in surveys and comparative
studies \cite{Conners1980Theoretical,Haralick1979Statistical,Ohanian1992Performance,reed1993review,
tuceryan1993texture,Van1985Texture,Weszka1976Comparative}. Tuceryan and Jain \cite{tuceryan1993texture} identified five major categories of features for texture discrimination: statistical, geometrical, structural, model based, and filtering based features. In 1996, Ojala \emph{et al.} \cite{Ojala1996Comparative} carried out a comparative
study to evaluate the classification performance of several texture features. In 1999, Randen and Hus{\o}y~\cite{randen1999filtering} reviewed most major filtering based texture features and performed a comparative performance evaluation for texture segmentation. In 2002, Zhang and Tan~\cite{zhang2002brief} reviewed invariant texture feature extraction methods. In 2007, Zhang \emph{et al.} \cite{Zhang07} evaluated the performance of several major invariant local texture descriptors. The 2008 book ``Handbook of Texture Analysis'' edited by Mirmehdi \emph{et al.} \cite{Xie08} contains representative work on texture analysis --- from 2D to 3D, from feature extraction to synthesis, and from texture image acquisition to
classification. The book ``Computer Vision Using Local Binary Patterns'' by Pietik\"{a}inen \emph{et al.}~\cite{Pietikainen11} in 2011 provides an excellent overview of the theory of Local Binary Patterns (LBP)
and the use in solving various kinds of problems in computer vision, especially in biomedical applications and biometric recognition systems. Huang \emph{et al.}~\cite{Huang2011local} presented a review of the LBP variants in the application area of facial image analysis. The book ``Local Binary Patterns: New Variants and
Applications''  by Brahnam \emph{et al.} \cite{Brahnam2014Local} in 2014 is a collection of several new LBP variants and their applications to face recognition. More recently, Liu \emph{et al.} \cite{liu2017local} conducted a taxonomy of recent LBP variants and performed a large scale performance evaluation of forty texture features. Researchers \cite{Raad2017Survey,Akl2018Survey} presented a review of exemplar based texture synthesis approaches.

The published surveys \cite{Conners1980Theoretical,Haralick1979Statistical,Ohanian1992Performance,
Reed1990Segmentation,reed1993review,Ojala1996Comparative,Pichler1996Comparison,
tuceryan1993texture,Van1985Texture} mainly reviewed or compared methods prior to 1995. Similarly, the articles \cite{randen1999filtering,zhang2002brief} only covered approaches before 2000. There are more recent surveys \cite{Brahnam2014Local,Huang2011local,liu2017local,Pietikainen11}, however they focused exclusively on texture features based on LBP. The emergence of many powerful texture analysis techniques has given rise to a further increase in research activity in texture research since 2000, however none of these published surveys provides an extensive survey over that time. Given recent developments, we believe that there is a need for an updated survey, motivating this present work.
A thorough review and survey of existing work, the focus of this paper, will contribute to more progress in texture analysis. Our goal is to overview the core tasks
and key challenges in texture representation approaches, to define taxonomies of representative
approaches, to provide a review of texture datasets, and to summarize the performance of the
state of the art on publicly available datasets.
According to the different visual representations, this survey
categorizes the texture representation literature into three broad types: BoW-based, CNN-based, and attribute-based.
The BoW-based methods are organized according to their key components.
The CNN-based methods are categorized into one of pretrained CNN models, finetuned
CNN models, or handcrafted deep convolutional networks.

The remainder of this paper is organized as follows. Related background, including the problem and its applications, the progress made during the past decades, and the challenges of the problem, are summarized in Section~\ref{sec:Background}. From Sections \ref{sec:BoW} to~\ref{sec:Attributes} we give a detailed review of texture representation techniques for texture classification by providing a taxonomy to more clearly group the prominent alternatives. A summarization of benchmark texture databases and state of the art performance is given in Section~\ref{sec:Texturedatasets}. Section~\ref{sec:Discussion} concludes the paper with a discussion of
promising directions for texture representation.

\begin {figure*}[!t]
\centering
\includegraphics[width=0.9\textwidth]{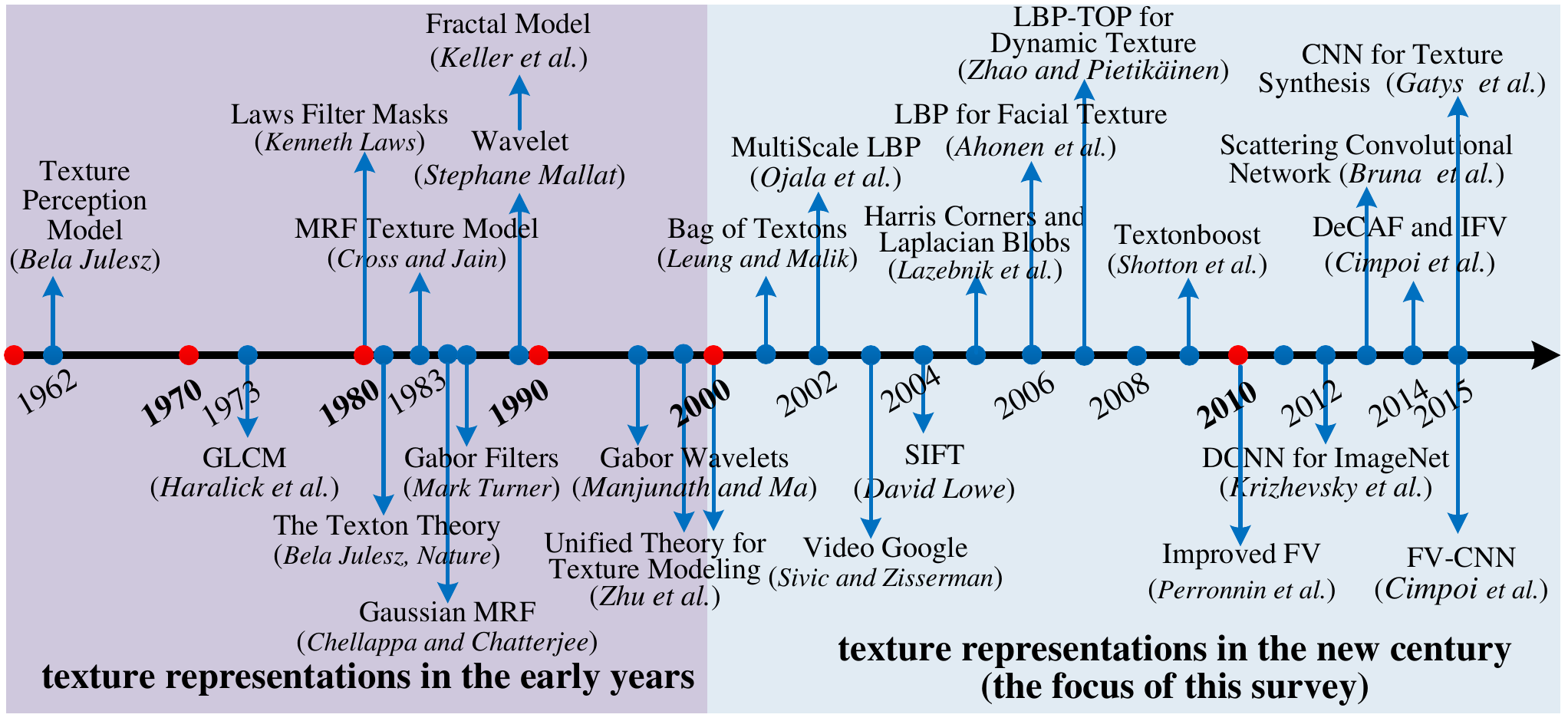}
\caption{The evolution of texture representation over the past decades (see discussion in Section~\ref{sec:summary}).}
\label{fig:milestones}
\end {figure*}

\section{Background}
\label{sec:Background}
\subsection{\textbf{The Problem}}
\label{sec:Problem}
Texture analysis can be divided into four areas: classification, segmentation, synthesis, and shape from texture \cite{tuceryan1993texture}. Texture classification \cite{Lazebnik05,LiuFieguthPAMI,tuceryan1993texture,Varma05,Varma09} deals with designing algorithms for declaring a given  texture region or image as belonging to one of a set of known texture categories of which training samples have been provided. Texture classification may also be a binary hypothesis testing problem, such as differentiating a texture as being within or outside of a given class, such as distinguishing between healthy and pathological tissues in medial image analysis. The goal of texture segmentation is to partition a given image into disjoint regions of homogeneous texture \cite{Jain1991Unsupervised,Manjunath1991Unsupervised,Reed1990Segmentation,Shotton2009textonboost}. Texture synthesis is the process of generating new texture images which are perceptually
equivalent to a given texture sample \cite{Efros1999Texture,Gatys2015Texture,portilla2000parametric,Raad2017Survey,Wei2000Fast,Zhu1998Filters}. As textures provide powerful shape cues, approaches for shape from texture attempt to recover the three dimensional shape of a textured object from its image. It should be noted that the concept of ``texture'' may have different connotations or definitions depending on the given objective. Classification, segmentation, and synthesis are closely related and widely studied, with shape from texture receiving comparatively less attention.
Nevertheless, texture representation is at the core of these four problems.
Texture representation, together with texture classification, will form the primary focus of this survey.

As a classical pattern recognition problem, texture classification primarily consists of two critical subproblems: texture representation and classification \cite{Jain2000statistical}. It is generally agreed that the extraction of powerful texture features plays a relatively more important role, since if poor features are used even the best classifier will fail to achieve good results. While this survey is not explicitly concerned with texture synthesis, studying synthesis can be instructive, for example, classification of textures via \emph{analysis by synthesis} \cite{Gatys2015Texture} in which a model is first constructed for synthesizing textures and then inverted for the purposes of classification. As a result, we will include representative texture modeling methods in our discussion.

\subsection{\textbf{Summary of Progress in the Past Decades}}
\label{sec:summary}
Milestones in texture representation over the past decades are listed in Fig.~\ref{fig:milestones}. The study of texture analysis can be traced back to the earliest work of Julesz \cite{julesz1962visual} in 1962, who studied the theory of human visual perception of texture and suggested that texture might be modelled using \emph{k}th order statistics --- the cooccurrence statistics for intensities at \emph{k}-tuples of pixels. Indeed, early work on texture features in the 1970s, such as the well known Gray Level Cooccurrence Matrix (GLCM) method \cite{Haralick1973Textural,Haralick1979Statistical},
were mainly driven by this perspective. Aiming at seeking essential ingredients
in terms of features and statistics in human texture perception, in the early 1980s Julesz \cite{julesz1981textons,Julesz1983Human} proposed the texton theory to explain texture preattentive discrimination, which states that textons (composed of local conspicuous features such as corners, blobs, terminators and crossings) are the elementary units of preattentive human texture perception and only the {\em first} order statistics of textons have perceptual significance: textures having the same texton densities could not be discriminated. Julesz's texton theory has been widely studied and
has largely influenced the development of texture analysis methods.

\begin {figure*}[!t]
\centering
\includegraphics[width=0.9\textwidth]{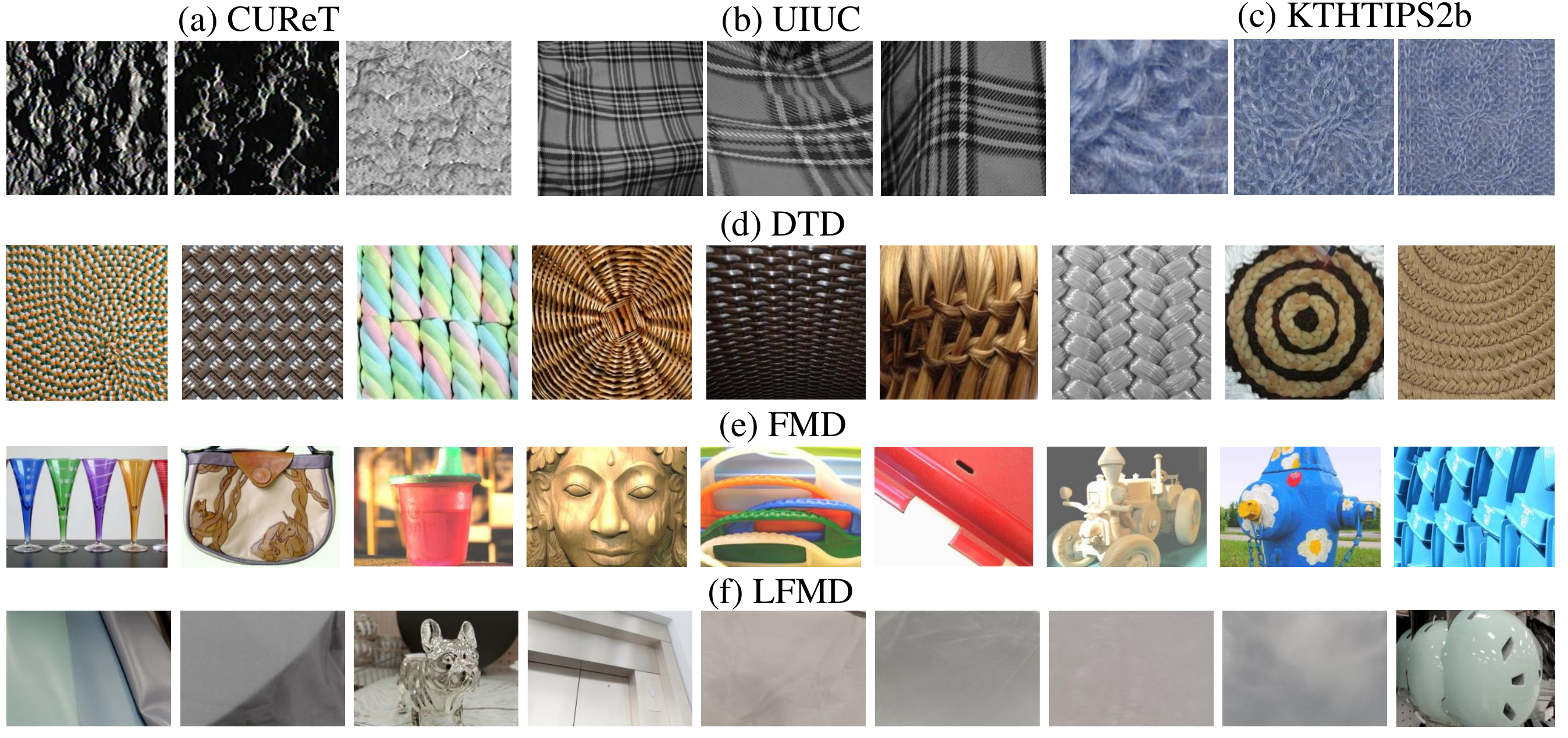}
\caption{Illustrations of challenges in texture recognition. Dramatic intraclass
variations: (a) Illumination variations, (b) View point and local nonrigid deformation, (c) Scale variations, and (d) Different instances from the same category. Small interclass variations make the problem harder still: (e) Images from the FMD database, and (f) Images from the LFMD database (photographed with a light-field camera). The reader is invited to identify the material category of the foreground surfaces in each image in (e) and (f). The correct answers are (from left to right): (e) glass, leather, plastic, wood, plastic, metal, wood, metal and plastic; (f) leather, fabric, metal, metal, paper, leather, water, sky and plastic. Section~\ref{sec:Texturedatasets} gives details regarding texture databases.}
\label{fig:TextureSamples}
\end {figure*}

Research on texture features in the late 1980s and the early 1990s mainly focused on two well-established areas:
\begin{enumerate}
\item {\em Filtering} approaches, which convolve an image with a bank of filters followed by some nonlinearity. One pioneering approach was that of Laws \cite{Laws1980Rapid},
where a bank of separable filters was applied, with subsequent filtering methods including Gabor filters \cite{Bovik1990Multichannel,Jain1991Unsupervised,Turner1986Texture}, Gabor wavelets \cite{Manjunath1996Texture}, wavelet pyramids \cite{Freeman1991Design,Mallat1989Theory}, and simple linear filters like Differences of Gaussians \cite{Malik1990Preattentive}.
\item {\em Statistical modelling}, which characterizes texture images as arising from probability distributions on random fields, such as a Markov Random Field (MRF) \cite{Cross1983Markov,Mao1992Texture,Chellappa1985Classification,Li2009Markov} or fractal models \cite{Keller1989Texture,Mandelbrot1983Fractal}.
\end{enumerate}
At the end of the last century there was a renaissance of texton-based approaches, including Zhu \emph{et al.} \cite{Wu2000Equivalence,Xie2015Learning,Zhu1998Filters,Zhu2000Exploring,Zhu2003Statistical,Zhu2005Textons} on the mathematical modelling of textures and textons. A notable stride was the Bag of Textons (BoT) \cite{leung2001representing} and later Bag of Words (BoW) \cite{Csurka2004,Sivic2003,Vasconcelos2000Probabilistic} approaches, where a dictionary of textons is generated, and images are represented statistically as orderless histograms over the texton dictionary.

In the 1990s, the need for invariant feature representations was recognized, to reduce or eliminate sensitivity to variations such as illumination, scale, rotation, view point etc. This gave rise to the development of local invariant descriptors, particularly milestone texture features such as Scale Invariant Feature Transform (SIFT) \cite{lowe2004distinctive}, Speeded Up Robust Features (SURF) \cite{Bay2006SURF} and LBP \cite{Ojala02}. Such local handcrafted texture descriptors dominated many domains of computer vision until the turning point in 2012 when deep Convolutional Neural Networks (CNN) \cite{Krizhevsky12} achieved record-breaking image classification accuracy. Since that time the research focus has been on deep learning methods for many problems in computer vision, including texture analysis \cite{Cimpoi14,Cimpoi15,Cimpoi2016deep}.

The importance of texture representations (such as Gabor filters \cite{Manjunath1996Texture}, LBP \cite{Ojala02}, BoT \cite{leung2001representing}, Fisher Vector (FV) \cite{Sanchez13}, and wavelet Scattering Convolution Networks (ScatNet) \cite{Bruna13}) is that they were found to be well applicable to other problems of image understanding and computer vision, such as object recognition \cite{Everingham2015,russakovsky2015imagenet}, scene classification \cite{Bosch2008Scene,Cimpoi2016deep,Kwitt2012Scene,Renninger2004Scene} and facial image analysis \cite{Ahonen06,Simonyan2013Fisher,Zhao07}. For
instance, recently many of the best object recognition approaches in challenges such as PASCAL VOC \cite{Everingham2015} and ImageNet ILSVRC \cite{russakovsky2015imagenet} were based on variants of texture representations.  Beyond BoT \cite{leung2001representing} and FV \cite{Sanchez13}, researchers developed Bag of Semantics (BoS) \cite{Dixit2015Scene,Dixit2016Object,Kwitt2012Scene,Li2014Object,Rasiwasia2012Holistic} which requires classifying image patches using BoT or CNN and considers the class posterior probability vectors as locally extracted semantic descriptors.  On the other hand, texture representations optimized for objects were also found to perform well for  texture-specific problems \cite{Cimpoi14,Cimpoi15,Cimpoi2016deep}. As a result, the division between texture descriptors and more generic image or video descriptors has been narrowing. The study of texture representation continues to play an important role in computer vision and pattern recognition.

\subsection{\textbf{Key Challenges}}
\label{subsec:Difficulties}

In spite of several decades of development, most texture features have not been capable of performing at a level sufficient for real-world textures and are computationally too complex to meet the real-time requirements of many computer vision applications. The inherent difficulty in obtaining powerful texture representations
lies in balancing two competing goals: \emph{high quality representation} and \emph{high efficiency}.

\textbf{High Quality} related challenges mainly arise due to the large intraclass appearance variations caused by changes in illumination, rotation, scale, blur, noise, occlusion, etc.\ and potentially small interclass appearance differences, requiring texture representations to be of high robustness and distinctiveness. Illustrative examples are shown in Fig.~\ref{fig:TextureSamples}. A further difficulty is in obtaining
sufficient training data in the form of labeled examples, which are frequently available only in
limited amounts due to collection time or cost.

\textbf{High Efficiency} related challenges include the potentially large number of different texture categories and their high dimensional representations. Here we have polar opposite motivations:  that of big data, with associated grand challenges and the scalability/complexity of huge problems, and that of tiny devices, the growing need for deploying highly compact and efficient texture representations on resource-limited platforms such as embedded and handheld devices.

\begin {figure}[!t]
\centering
\includegraphics[width=0.45\textwidth]{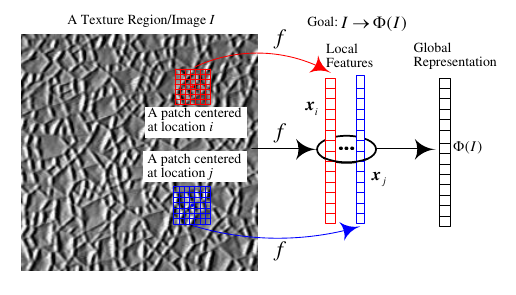}
\caption{The goal of texture representation is to transform the input texture image into a feature vector that describes the properties of the texture, facilitating subsequent tasks such as texture recognition. Usually a texture image is first transformed into a pool of local features, which are then aggregated into a global representation for an entire image or region.}
\label{fig:TextureRepresentation}
\end {figure}

\section{Bag of Words based Texture Representation}
\label{sec:BoW}

The goal of texture representation or texture feature extraction is to transform the input texture image into a feature vector that describes the properties of a texture, facilitating subsequent tasks such as texture classification, as illustrated in Fig.~\ref{fig:TextureRepresentation}. Since texture is a spatial phenomenon, texture representation cannot be based on a single pixel, and generally requires the analysis of patterns over local pixel neighborhoods. Therefore, a texture image is first transformed to a pool of local features, which are then aggregated into a global representation for an entire image or region. Since the properties of texture are usually translationally invariant, most texture representations are based on an orderless aggregation of local texture features, such as a sum or max operation.

Early in 1981, Julesz \cite{julesz1981textons} introduced ``textons'', which refer to basic image features such as elongated blobs, bars, crosses, and terminators, as the elementary units of preattentive human texture perception. However Julesz's texton studies were limited by their exclusive focus on artificial texture patterns rather than natural textures. In addition, Julesz did not provide a rigorous definition for textons. Subsequently, texton theory fell into disfavor as a model of texture discrimination until the influential work by Leung and Malik \cite{leung2001representing} who revisited textons and gave an operational definition of a texton as a cluster center in filter response space. This not only enabled textons to be generated automatically from an image, but also opened up the possibility of learning a universal texton dictionary for all images. Texture images can be statistically represented as histograms over a texton dictionary, referred to as the Bag of Textons (BoT) approach.  Although BoT was initially developed in the context of texture recognition \cite{leung2001representing,Malik1999Textons}, it was introduced / generalized to image retrieval \cite{Sivic2003} and classification \cite{Csurka2004}, where it was referred to as Bag of Features (BoF) or, more commonly, Bag of Words (BoW). The research community has since witnessed the prominence of the BoW model for over a decade during which many improvements were proposed.
\begin {figure*}[!t]
\centering
\includegraphics[width=0.98\textwidth]{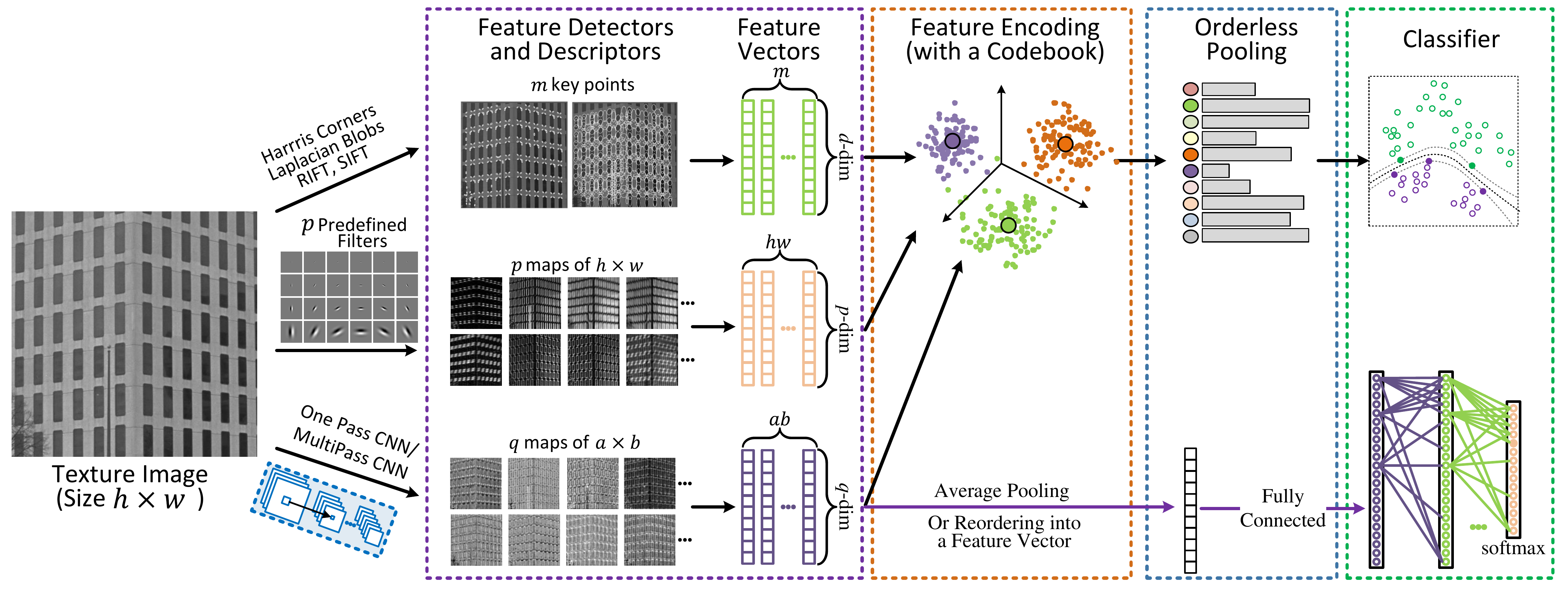}
\caption{General pipeline of the BoW model. See Table \ref{Tab:BoW}, and also refer to Section \ref{sec:BoW} for detail discussion. Features are computed from handcrafted detectors for descriptors like SIFT and RIFT,
and densely applied local texture descriptors like handcrafted filters or CNNs. The CNN features can also be computed in an end-to-end manner using finetuned CNN models.
These local features are quantized to visual words in a codebook.}
\label{fig:BoW}
\end {figure*}

\subsection{\textbf{The BoW Pipeline}}
The BoW pipeline is sketched in Fig.~\ref{fig:BoW}, consisting of the following basic steps:

\textbf{1.\ Local Patch Extraction.} For a given image, a pool of $N$ image patches is extracted over a sparse set of points of interest \cite{Lazebnik05,Zhang07}, over a fixed grid \cite{Kong2012Multi,Marszalek2007Learning,sharan2013recognizing}, or densely at each pixel position \cite{Ojala02,Varma05,Varma09}.

\textbf{2.\ Local Patch Representation.} Given the extracted $N$ patches, local texture descriptors are applied to obtain a set or pool of texture features of $D$ dimension. We denote the local features of $N$ patches in an image as $\{{\textbf{\emph{x}}}_i\}_{i=1}^{N}$, ${\textbf{\emph{x}}}_i\in\mathbb{R}^D$.  Ideally, local descriptors should be distinctive and at the same time robust to a variety of possible image transformations, such as scale, rotation, blur, illumination, and viewpoint changes. High quality local texture descriptors play a critical role in the BoW pipeline.

\textbf{3.\ Codebook Generation.} The objective of this step is to generate a codebook (\emph{i.e.,} a texton dictionary) with $K$ codewords $\{{\textbf{\emph{w}}}_i\}_{i=1}^{K}$, ${\textbf{\emph{w}}}_i\in\mathbb{R}^D$ based on training data.  The codewords may be learned (\emph{e.g.,} by \emph{k}means \cite{lazebnik2003sparse,Varma05}) or in a predefined way (such as LBP \cite{Ojala02}). The size and nature of the codebook affects the representation followed and thus the discrimination power. The key here is how to generate a compact and discriminative codebook so as to enable accurate and efficient classification.

\textbf{4.\ Feature Encoding.} Given the generated codebook and the extracted local texture features $\{{\textbf{\emph{x}}}_i\}$ from an image, feature encoding represents each local feature ${\textbf{\emph{x}}}_i$ with the codebook, usually by mapping each ${\textbf{\emph{x}}}_i$ to one or a number of codewords, resulting a feature coding vector ${\textbf{\emph{v}}}_i$
(\emph{e.g.} ${\textbf{\emph{v}}}_i\in\mathbb{R}^K$).  Of all the steps in the BoW pipeline, feature encoding is a core component which links local representation and feature pooling, greatly influencing texture classification in terms of both accuracy and speed. Thus, many studies have focused on developing powerful feature encoding, such as vector quantization / \emph{k}means, sparse coding \cite{mairal2008discriminative,mairal2009supervised,peyre2009sparse}, Locality constrained Linear Coding (LLC) \cite{wang2010locality}, Vector of Locally Aggregated Descriptors (VLAD) \cite{Jegou2012Aggregating}, and Fisher Vector (FV) \cite{Cimpoi2016deep,Perronnin10,Sanchez13}.

\textbf{5.\ Feature Pooling.} A global feature representation ${\textbf{\emph{y}}}$ is produced by using a feature pooling strategy to aggregate the coded feature vectors $\{{\textbf{\emph{v}}}_i\}$. Classical pooling methods include average pooling, max pooling, and Spatial Pyramid Pooling (SPM) \cite{lazebnik2006beyond,timofte2012training}.

\textbf{6.\ Feature Classification.} The global feature is used as the basis for  classification, for which many approaches are possible \cite{Jain2000statistical,Webb2011}:  Nearest Neighbor Classifier (NNC), Support Vector Machines (SVM), neural networks, and random forests. SVM is one of the most widely used classifiers for the BoW based representation.

The remainder of this section will introduce the methods in each component, as summarized in Table~\ref{Tab:BoW}.

\begin{table*}[t]
\caption {A summary of components in the BoW representation pipeline, as was sketched in Fig.~\ref{fig:BoW}.}\label{Tab:BoW}
\centering
\renewcommand{\arraystretch}{1}
\setlength\arrayrulewidth{0.2mm}
\setlength\tabcolsep{2pt}
\resizebox*{18cm}{!}{
\begin{tabular}{c|lp{10cm}}
\Xhline{1pt}
\scriptsize \textbf{Step} &\scriptsize \textbf{Approach} & \scriptsize \textbf{Highlights} \\
\Xhline{1pt}
\cline{2-3}
\multirow{16}{*}{\rotatebox{90}{\scriptsize \shortstack [c] {\textbf{Local Texture Descriptors}  \\ \textbf{(Section \ref{Sec:LocalDes})}} }}&\scriptsize   Sparse Descriptors & \scriptsize     \\
&\scriptsize  $\quad\bullet$ (Harris+Laplacian)(RIFT+SPIN) \cite{Lazebnik05}& \scriptsize   Keypoint detectors plus novel descriptors SPIN and RIFT \\
&\scriptsize  $\quad\bullet$ (Harris+Laplacian)(RIFT+SPIN+SIFT) \cite{Zhang07}& \scriptsize A comprehensive
evaluation of multiple keypoint detectors, feature descriptors, and
classifier kernels. \\
\cline{2-3}
&\scriptsize  Dense Descriptors & \scriptsize     \\
&\scriptsize  $\quad\bullet$ Gabor Wavelets  & \scriptsize  Joint optimum resolution in time and frequency; Multiscale and multiorientation analysis. \\
&\scriptsize  $\quad\bullet$ LMfilters \cite{leung2001representing}& \scriptsize First to propose Bag of Texton (BoT) model (\emph{i.e.} the BoW model) \\
&\scriptsize  $\quad\bullet$ Schmid Filters  & \scriptsize Gabor like filters; Rotation invariant. \\
&\scriptsize  $\quad\bullet$ MR8 \cite{Varma05}& \scriptsize Rotationally invariant filters and low-dimensional filter response space. \\
&\scriptsize  $\quad\bullet$ Patch Intensity \cite{Varma09}& \scriptsize Challenge the dominant role of filter descriptors and propose image raw intensity feature. \\
&\scriptsize  $\quad\bullet$ LBP \cite{Ojala02}& \scriptsize Fast binary features with gray scale invariance; Predefined codebook. \\
&\scriptsize  $\quad\bullet$ Random Projection \cite{LiuFieguthPAMI}& \scriptsize First to introduce compressive sensing and random projection into texture classification. \\
&\scriptsize  $\quad\bullet$ Sorted Random Projection \cite{liu2011sorted}& \scriptsize Efficient and effective approach for random projection to achieve rotation invariance. \\
&\scriptsize  $\quad\bullet$ Basic Image Features (BIFs) \cite{Crosier10} & \scriptsize Introduce BIFs of Griffin and
Lillholm into texture classification; Predefined codebook.\\
&\scriptsize  $\quad\bullet$ Weber Local Descriptor (WLD) \cite{Crosier10} & \scriptsize A descriptor based on Weber's Law. \\
\cline{2-3}
&\scriptsize  Fractal Based Descriptors & \scriptsize   \\
&\scriptsize  $\quad\bullet$ MultiFractal Spectrum \cite{xu2009viewpoint} & \scriptsize Invariant under the bi-Lipschitz mapping. \\
\hline
\multirow{4}{*}{\rotatebox{90}{\scriptsize \shortstack [c] {\textbf{Codebook}  \\ \textbf{Generation} \\ (\textbf{Section \ref{Sec:CodebookGene}})} }}&\scriptsize   Predefined \cite{Crosier10,Ojala02} & \scriptsize No codebook learning step; Computationally efficient.  \\
&\scriptsize   \emph{k}means clustering \cite{Csurka2004,leung2001representing} & \scriptsize Most commonly used method; Cannot capture overlapping distributions in the feature space. \\
&\scriptsize   GMM modeling \cite{Cimpoi2016deep,Perronnin10,sharma2016local}
& \scriptsize   Considers both cluster centers and covariances
which describe the spreads of clusters.  \\
&\scriptsize   Sparse Coding based learning \cite{peyre2009sparse,Skretting2006Texture}  & \scriptsize Sparse representation based; Minimize reconstruction error of data; Computationally expensive.  \\
\hline
\multirow{10}{*}{\rotatebox{90}{\scriptsize \shortstack [c] {\textbf{Feature Encoding}\\ \textbf{(Section \ref{Sec:FeatEncode})}} }}&\scriptsize  Voting Based Methods & \scriptsize  Require a large codebook (usually learned by \emph{k}means); Usually combine with nonlinear SVM. \\
&\scriptsize $\quad\bullet$ Hard Voting \cite{leung2001representing,Varma05} & \scriptsize  Quantize each feature to nearest codeword; Fast to compute; Codes are sparse and high dimensional. \\
&\scriptsize $\quad\bullet$ Soft Voting \cite{Ahonen2007Soft,Ren2013Noise,Van2008Kernel} & \scriptsize  Assigns each feature to multiple codewords; Does not minimize reconstruction error.  \\
\cline{2-3}
&\scriptsize  Fisher Vector (FV) Based Methods & \scriptsize Require a small codebook; Very high dimension; Combines with efficient linear SVM. \\
&\scriptsize $\quad\bullet$ FV \cite{Perronnin2007Fisher} & \scriptsize GMM-based;  Encodes higher order statistics; Efficient to compute. \\
&\scriptsize $\quad\bullet$ Improved FV (IFV) \cite{Cimpoi14,Perronnin10,sharma2016local} & \scriptsize Uses signed square rooting and $L_2$ normalization; State of the art performance in texture classification. \\
&\scriptsize $\quad\bullet$  VLAD \cite{Jegou2012Aggregating,Cimpoi14} & \scriptsize A simplified version of FV.   \\
\cline{2-3}
&\scriptsize  Reconstruction Based Methods & \scriptsize Enforce sparse representation; Explores the manifold structure of data; Minimize reconstruction error. \\
&\scriptsize $\quad\bullet$ Sparse Coding \cite{peyre2009sparse,Skretting2006Texture,Yang2009Linear} & \scriptsize Leverage that fact that natural images are sparse; Optimization is computationally expensive.  \\
&\scriptsize $\quad\bullet$ Local constraint Linear Coding (LLC) \cite{Cimpoi14,wang2010locality} & \scriptsize
Local smooth sparsity; Fast computation through approximated LLC. \\
\hline
\multirow{3}{*}{\rotatebox{90}{\scriptsize \shortstack [c] {\textbf{Feature}  \\ \textbf{Pooling}\\ \textbf{(Sec. \ref{Sec:FeatPool})}} }}&\scriptsize  Average Pooling & \scriptsize The most widely used pooling scheme in texture representation.  \\
& \scriptsize Max Pooling & \scriptsize  Usually used in combination with sparse coding and LLC. \\
&\scriptsize Spatial Pyramid Pooling (SPM) & \scriptsize Preserving more spatial information; Higher feature dimensionality.  \\
\hline
\multirow{3}{*}{\rotatebox{90}{\scriptsize \shortstack [c] {\textbf{Classifier} \\ \textbf{(Sec. \ref{Sec:FeatPool})}} }} &\scriptsize Nearest Neighbor Classifier (NNC) \cite{LiuFieguthPAMI,Varma05} & \scriptsize Simple and elegant nonparametric classifier; Popular in texture classification. \\
&\scriptsize  Kernel SVM \cite{Zhang07}  & \scriptsize Usually in combination with Chi Square for BoW based representation.  \\
&\scriptsize Linear SVM \cite{Cimpoi2016deep} & \scriptsize Suitable for high-dimensional feature representation like FV and VLAD.  \\
\Xhline{1pt}
\end{tabular}
}
\end{table*}

\subsection{\textbf{Local Texture Descriptors}}
\label{Sec:LocalDes}

All local texture descriptors aim to provide local representations invariant to contrast, rotation, scale, and possibly other criteria.  The primary categorization is whether the descriptor is applied densely, at every pixel, as opposed to sparsely, only at certain locations of interest.

\subsubsection{Sparse Texture Descriptors}
\label{subsubsec:Sparse}

To develop a sparse texture descriptor, a region of interest detector must be designed which is able to reliably detect a sparse set of regions, reliably and stably, under various
imaging conditions. Typically, the detected regions undergo a geometric normalization, after which local descriptors are applied to encode the image content. A series of region detectors and local descriptors has been proposed, with excellent surveys \cite{mikolajczyk2005performance,mikolajczyk2005comparison,tuytelaars2008local}. The sparse approach was introduced to texture recognition by Lazebnik \emph{et al.} \cite{lazebnik2003sparse,Lazebnik05} and followed by Zhang \emph{et al.} \cite{Zhang07}.

In \cite{Lazebnik05} two types of complementary region detectors, the Harris affine
detector of Mikolajczyk and Schmid \cite{Mikolajczyk2002Affine} and the
Laplacian blob detector of G{\aa}rding and Lindeberg \cite{Garding1996Direct}, were used to detect affine covariant regions, meaning that the region content is affine invariant. Each detected region can be thought of as a texture element having a characteristic elliptic shape and a distinctive appearance pattern. In order to achieve affine invariance, each elliptical region was normalized and then two rotation invariant descriptors, the spin image (SPIN) and the Rotation Invariant Feature Transform (RIFT) descriptor, were applied. As a result, for each texture image four feature channels were extracted (two detectors $\times $ two descriptors), and for each feature channel \emph{k}means clustering is performed to form its signature. The Earth Mover's Distance (EMD) \cite{Rubner2000Earth} was used for measuring the similarity between image signatures and NNC was used for classification. The Harris affine regions and Laplacian blobs in combination with SPIN and RIFT descriptors (\emph{i.e.} the  (H+L)(S+R) method) have demonstrated good performance (listed in Table~\ref{Tab:Results}) in classifying textures with significant affine variations, evidenced by the classification rate $96.0\%$ on UIUC with a NNC classifier. Although this approach achieve affine invariance, they lack distinctiveness since some spatial information is lost due to their feature pooling schemes.

Following Lazebnik \emph{et al.} \cite{Lazebnik05}, Zhang \emph{et al.} \cite{Zhang07} presented an evaluation of multiple region detector types, levels of geometric invariance, multiple local texture descriptors, and SVM classifier with kernels based on two effective measures for comparing distributions (signatures and EMD distance vs.\ standard BoW and the Chi Square distance) for texture and object recognition. Regarding local description, Zhang \emph{et al.} \cite{Zhang07} also used the SIFT descriptor\footnote{Originally, SIFT is
comprised of a detector and descriptor, but which are used
in isolation now; in this survey, if not specified, SIFT refers to the descriptor, a common practice in the
community. } in addition to SPIN and RIFT. With SVM classification, Zhang \emph{et al.} \cite{Zhang07} showed significant performance improvement over that of Lazebnik \emph{et al.} \cite{Lazebnik05}, and reported classification rates of $95.3\%$ and $98.7\%$ on CUReT and UIUC respectively. They recommended that practical texture recognition should seek to incorporate multiple types of complementary features, but with local invariance properties not exceeding those absolutely required for a given application. Other local region detectors have also been used for texture description, such as the Scale Descriptors which measure the scales of salient textons \cite{Kadir2002Scale}.

\subsubsection{Dense Texture Descriptors}
\label{subsubsec:Dense}

The number of features derived from a sparse set of interesting points is much smaller than the total number of image pixels, resulting a compact feature space. However, the sparse approach can be inappropriate for many texture classification tasks:
\begin{itemize}
\renewcommand{\labelitemi}{$\circ$}
\item  Interest point detectors typically produce
a sparse output and could miss important texture elements.
\item  A sparse output in a small image might not
produce sufficient regions for robust statistical characterization.
\item  There are issues regarding the repeatability of the detectors, the stability of the selected regions and the instability of orientation estimation \cite{mikolajczyk2005comparison}.
\end{itemize}
 As a result, extracting local texture features {\em densely} at each pixel is the more popular representation, the subject of the following discussion.

\begin {figure}[!t]
\centering
\includegraphics[width=0.24\textwidth]{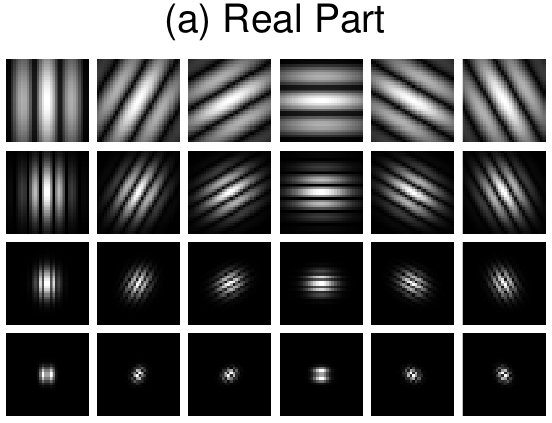}
\includegraphics[width=0.24\textwidth]{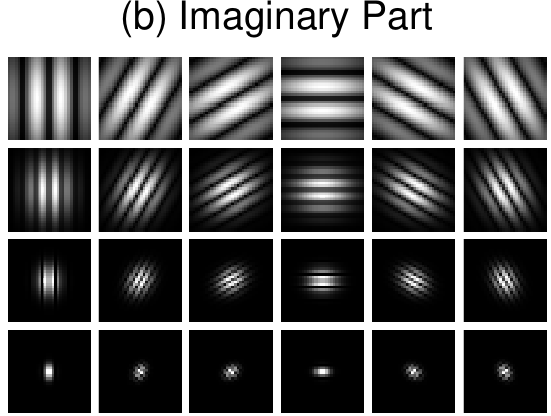}
\caption{Illustration of the Gabor wavelets used in \cite{Manjunath1996Texture}.}
\label{fig:Gabor}
\end {figure}

\textbf{(1) Gabor Filters} are one of the most popular texture descriptors, motivated by their relation to models of early visual systems of mammals as well as their joint optimum resolution in time and frequency \cite{Jain1991Unsupervised,Lee1996Image,Manjunath1996Texture}. As illustrated in Fig.~\ref{fig:Gabor}, Gabor filters can be considered as orientation and scale tunable edge and bar detectors.  The Gabor wavelets are generated by appropriate rotations and dilations from the following product of an elliptical Gaussian
and a complex plane wave:
\begin{eqnarray*}\label{eqn:Gabor}
    \phi(x,y)=\frac{1}{2\pi\sigma_x\sigma_y}\textrm{exp}{\left[-\left(\frac{x^2}{2\sigma_x^2}+\frac{y^2}{2\sigma_y^2}\right)\right]}
    \textrm{exp}{(j2\pi\omega)},
\end{eqnarray*}
whose Fourier transform is
\begin{eqnarray*}\label{eqn:Gabor}
    \hat{\phi}(x,y)=\textrm{exp}{\left[-\left(\frac{(u-\omega)^2}{2\sigma_u^2}+\frac{v^2}{2\sigma_v^2}\right)\right]},
\end{eqnarray*}
where $\omega$ is the radial center frequency of the filter in the frequency domain, $\sigma_x$ and $\sigma_y$ are the standard deviations of the elliptical Gaussian along $x$ and $y$.

Thus, a Gabor filter bank is defined by its parameters including frequencies, orientations and the parameters of the Gaussian envelope. In the literature, different parameter settings have been suggested, and filter banks created by these parameter settings work well in general. Details for the derivation of Gabor wavelets and parameter selection can be found in \cite{Lee1996Image,Manjunath1996Texture,Petrou2006Image}. Invariant Gabor representations can be accessed in \cite{Han2007Rotation}. According to the experimental study in \cite{Kandaswamy2011,Zhang07}, Gabor features \cite{Manjunath1996Texture} fail to meet the expected level of performance in the presence of rotation, affine and scale variations. However, Gabor filters encode structural features from multiple orientations and over a broader range of scales. It has been shown \cite{Kandaswamy2011} that for large datasets, under varying illumination conditions,
Gabor filters can serve as a preprocessing method and combine with LBP \cite{Ojala02} to obtain texture features with reasonable robustness \cite{Pietikainen11,Zhang2005Local}.

\begin {figure}[!t]
\centering
\includegraphics[width=0.48\textwidth]{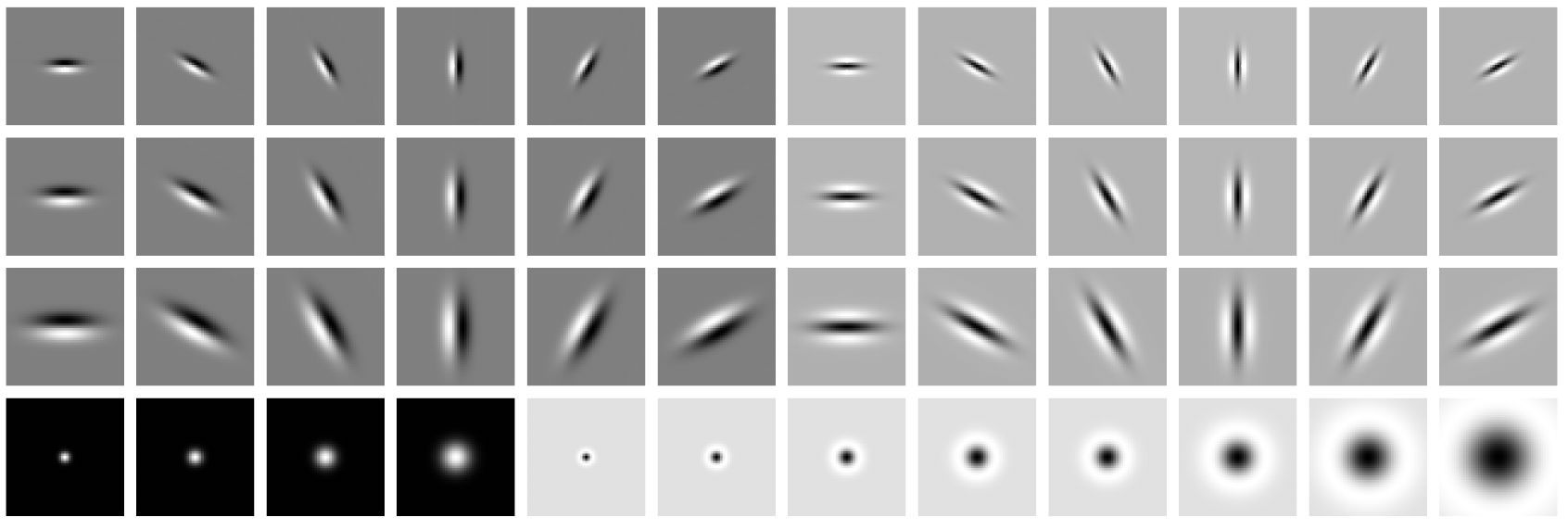}
\caption{The LMfilter bank has a mix of edge, bar and spot filters at multiple scales and orientations. It has a total of 48 filters: 2 Gaussian derivative filters at 6 orientations and 3 scales, 8 Laplacian of Gaussian filters and 4 Gaussian filters.}
\label{fig:LMfilters}
\end {figure}

\begin {figure}[!t]
\centering
\includegraphics[width=0.48\textwidth]{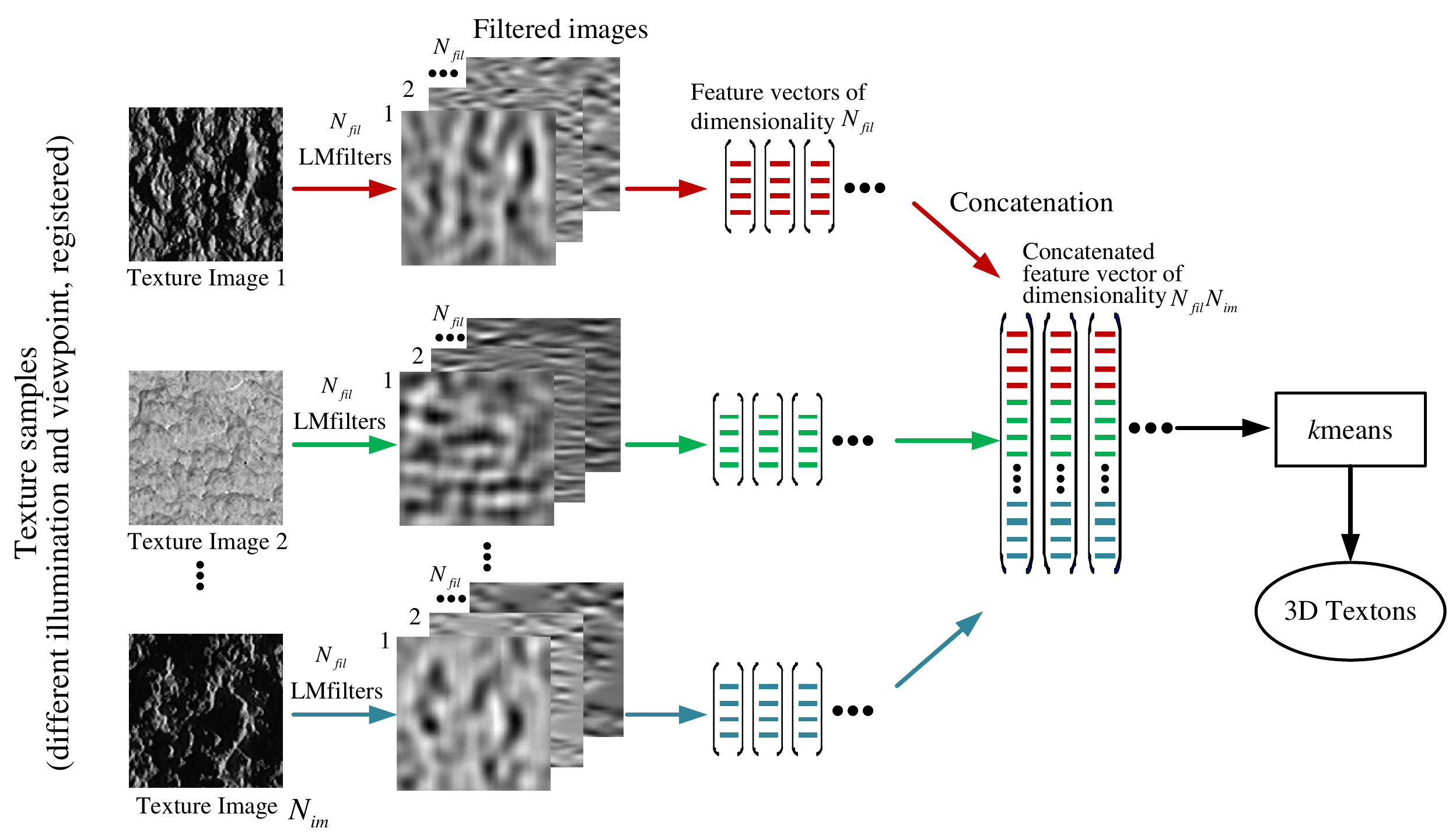}
\caption{Illustration of the process of 3D texton dictionary learning proposed by Leung and Malik \cite{leung2001representing}. Each image at different lighting and viewing directions is filtered using the filter bank illustrated in Fig. \ref{fig:LMfilters}. The response vectors are concatenated together to form data vectors of length $N_{fil}N_{im}$. These data vectors are clustered using the \emph{k}means algorithm to obtain the 3D textons.}
\label{fig:LM3DTextonLearning}
\end {figure}

\textbf{(2) Filters by Leung and Malik (LM Filters)} \cite{leung2001representing,Malik1999Textons} pioneered the problem of classifying textures under
varying viewpoint and illumination. The LM filters used for local texture feature extraction are illustrated in Fig.~\ref{fig:LMfilters}. In particular, they marked a milestone by giving an operational
definition of textons: the cluster centers of the filter response vectors. Their work has been widely followed by other researchers \cite{Csurka2004,Lazebnik05,Shotton2009textonboost,Sivic2003,Varma05,Varma09}. To handle 3D effects caused by imaging, they proposed 3D textons which were cluster centers of filter responses over a stack of images with representative viewpoints and lighting, as illustrated in Fig.~\ref{fig:LM3DTextonLearning}.
In their texture classification algorithm, 20 images of each texture were geometrically registered and transformed into 48D local features with the LM Filters. Then the 48D filter response vectors of 20 selected images of the same pixel were concatenated to obtain a 960D feature vector as the local texture representation, subsequently input into a BoW pipeline for texture classification.
A downside of the method is that it is not suitable for classifying a single texture image under \emph{unknown} imaging conditions, which usually arises in practical applications.

\begin {figure}[!t]
\centering
\includegraphics[width=0.4\textwidth]{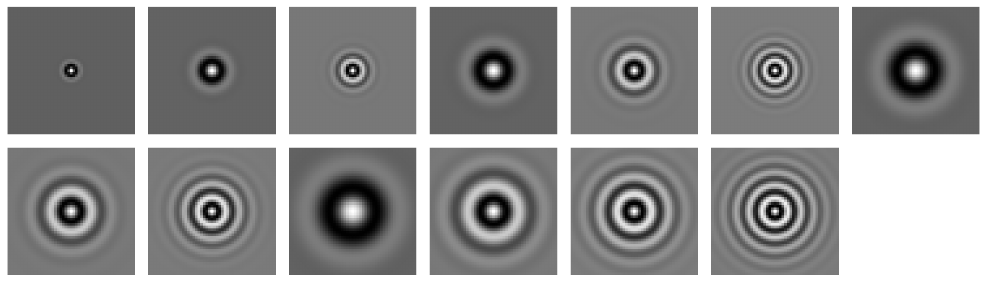}
\caption{Illustration of the rotationally invariant Gabor-like Schmid filters used in \cite{Schmid2001constructing}. The parameter $(\sigma,\beta)$ pair takes values (2,1), (4,1), (4,2),
(6,1), (6,2), (6,3), (8,1), (8,2), (8,3), (10,1), (10,2), (10,3) and (10,4).}
\label{fig:SchmidFilters}
\end {figure}

\textbf{(3) The Schmid Filters (S Filters)} \cite{Schmid2001constructing} consist of 13
rotationally invariant Gabor-like filters of the form
\begin{eqnarray*}\label{eqn:SFilters}
    \phi(x,y)=\textrm{exp}{\left[-\left(\frac{x^2+y^2}{2\sigma^2}\right)\right]}cos\left(\frac{\pi\beta\sqrt{x^2+y^2}}{\sigma}\right),
\end{eqnarray*}
where $\beta$ is the number of cycles of the harmonic function
within the Gaussian envelope of the filter. The filters are shown in Fig.~\ref{fig:SchmidFilters}; as can be seen, all of the filters have rotational symmetry. The rotation-invariant S Filters were shown to outperform the rotation-variant LM Filters in classifying the CUReT textures \cite{Varma05}, indicating that rotational invariance is necessary in practical applications.

\begin {figure}[!t]
\centering
\includegraphics[width=0.3\textwidth]{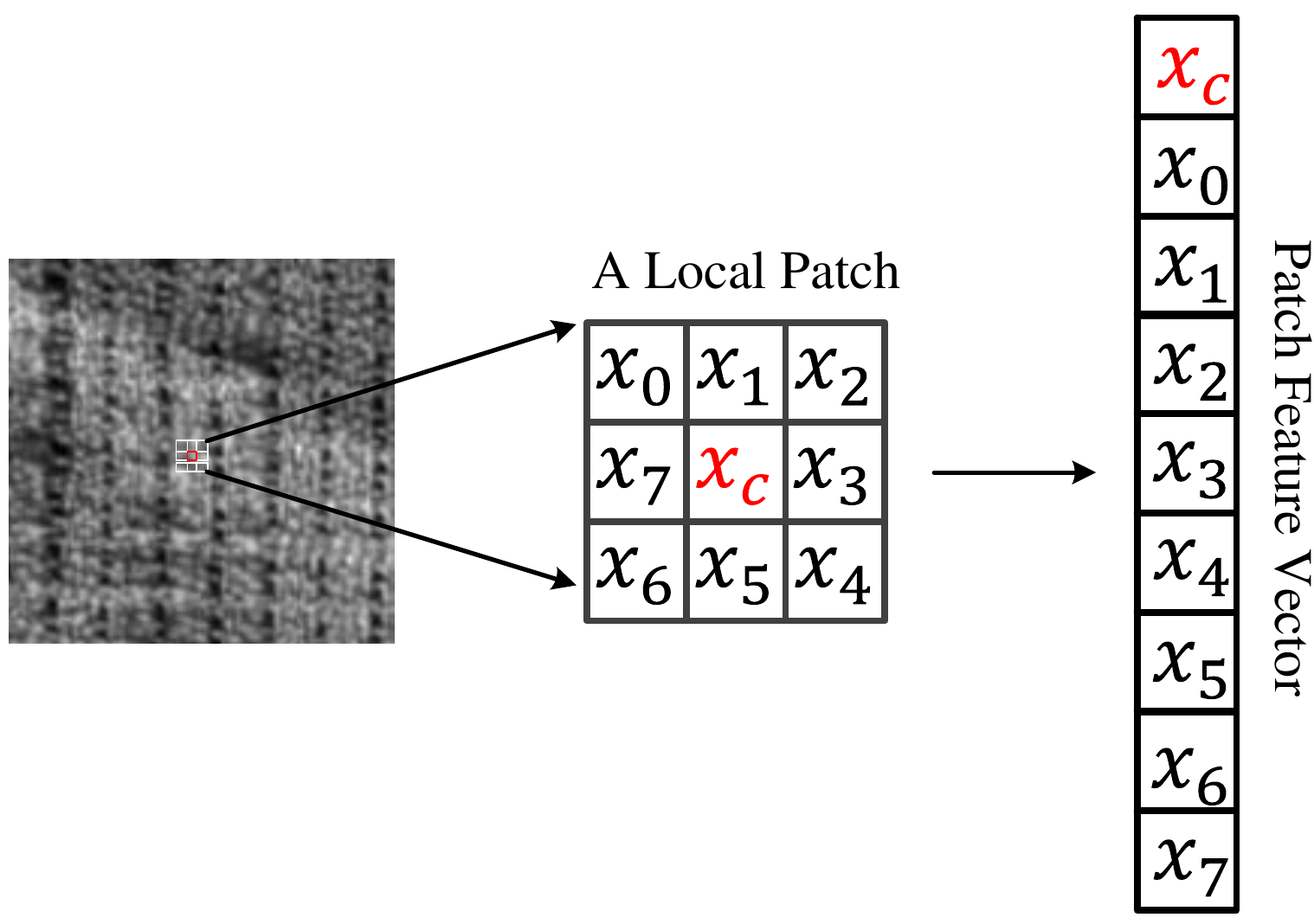}
\caption{Illustration for the Patch Descriptor proposed in \cite{Varma09}: the raw intensity vector is used directly as the local representation.}
\label{fig:Patch}
\end {figure}

\textbf{(4) Maximum Response (MR8) Filters} of Varma and Zisserman \cite{Varma05} consist of 38 root filters but only 8 filter responses. The filter bank contains filters at multiple orientations but their outputs are pooled by recording only the maximum filter response across all orientations, in order to achieve rotation invariance. The root filters are a subset of the LM Filters \cite{leung2001representing} of Fig.~\ref{fig:LMfilters}, retaining the two rotational symmetry filters, the edge filter, and the bar filter at 3 scales and 6 orientations. Recording only the maximum response across orientations reduces the number of responses from 38 to 8 (3 scales for 2 anisotropic filters, plus 2 isotropic), resulting the so called MR8 filter bank.

Realizing the shortcomings of Leung and Malik's method \cite{leung2001representing}, Varma and Zisserman \cite{Varma05} attempted to improve the classification of a single texture sample image under unknown imaging conditions, bypassing the registration step, instead learning 2D textons by aggregating filter responses over different images. Experimental results \cite{Varma05} showed that MR8 outperformed the LM Filters and S Filters, indicating that detecting better features and clustering in a lower dimensional feature
space can be advantageous. The best results for MR8 are
$97.4\%$ obtained with a dictionary of $2440$ textons and a Nearest Neighbor Classifier (NNC) \cite{Varma05}. Later, Hayman \emph{et al.} \cite{hayman2004significance} showed that SVM could further enhance the texture classification performance of MR8 features, giving a $98.5\%$ classification rate for the same setup used for texton representation.

\textbf{(5) Patch Descriptors} of Varma and Zisserman \cite{Varma09} challenged the dominant role of the filter banks \cite{mellor2008locally,randen1999filtering} in texture analysis, and instead developed a simple Patch Descriptor, keeping the raw pixel intensities of a square neighborhood to form a feature vector, as illustrated in Fig. \ref{fig:Patch}. By replacing the filter responses such as LM Filters \cite{randen1999filtering}, S Filters \cite{Schmid2001constructing} and MR8 \cite{Varma05} with the Patch Descriptor in texture classification, Varma and Zisserman \cite{Varma09} observed very good classification performance using extremely compact neighborhoods ($3\times3$), and that for any fixed size of neighborhood the Patch Descriptor leads to superior classification compared to filter banks with the same support.

Two variants of the Patch Descriptor, the
Neighborhood Descriptor and the MRF Descriptor, were developed. For the Neighborhood Descriptor, the central pixel is discarded and only the neighborhood vector is used for texton representation. Instead of ignoring the central pixel, the MRF Descriptor explicitly models the joint distribution of the central
pixels and its neighbors. The best result $98.0\%$ is given by the MRF Descriptor using a
$7\times7$ neighborhood with 2440 textons and 90 bins and a NNC classifier. Note that the dimensionality of this MRF representation is very high:  $2440\times90$. A clear limitation of the Patch, Neighborhood and MRF Descriptors is sensitivity to nearly any change (brightness, rotation, affine \emph{etc.}). Varma and Zisserman \cite{Varma09} adopted the method of finding the dominant orientation of a patch and measuring the neighborhood relative to this orientation to achieve rotation invariance, and reported a $97.8\%$ classification rate on the UIUC dataset. It is worth mentioning that finding the dominant orientation for each patch is computationally expensive.

\begin {figure}[!t]
\centering
\includegraphics[width=0.45\textwidth]{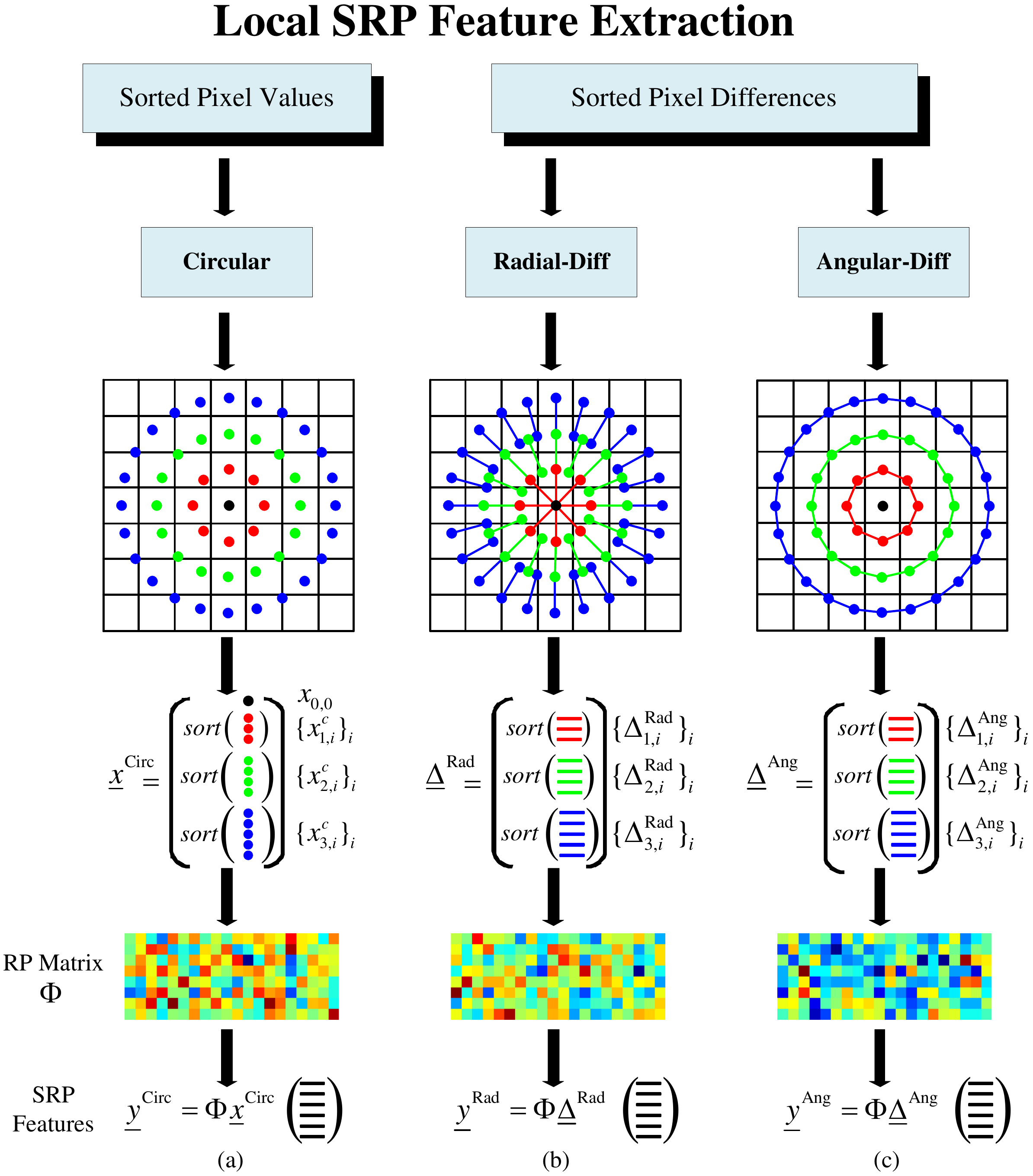}
\caption{An illustration of SRP descriptor: extracting SRP features on an example local image patch of size $7\times7$. (a) Sorting pixel intensities; (b,c) Sorting pixel differences.}
\label{fig:SRP}
\end {figure}

\textbf{(6) Random Projection (RP) and Sorted Random Projection (SRP)} features of Liu and Fieguth \cite{LiuFieguthPAMI} were inspired by theories of sparse representation and compressed sensing \cite{candes2006near,donoho2006compressed}. Taking advantage of the sparse nature of textured images, a small set of random features is extracted from local image patches by projecting the local patch feature vectors to a lower dimensional feature subspace. The random projection is a fixed, distance-preserving embedding capable of alleviating the curse of dimensionality \cite{Baraniuk2008simple,Giryes2016Deep}. The random features are embedded into BoW to perform texture classification. It has been
shown that the performance of RP features is superior to that of the Patch Descriptor with equivalent neighborhoods \cite{LiuFieguthPAMI}; a clear indication that the RP matrix preserves the
salient information contained in the local patch and that performing classification
in a lower feature space is advantageous. The best result $98.5\%$ is achieved using a
$17\times17$ neighborhood with 2440 textons and a NNC classifier.

Like the Patch Descriptors, the RP features remain sensitive to image rotation. To further improve robustness, Liu \emph{et al.} \cite{LiuFieguthPR,liu2011sorted} proposed sorting the RP features, as illustrated in Fig.~\ref{fig:SRP}, whereby rings of pixel values are sorted, without any reference orientation, ensuring rotation invariance. Two kinds of local features are used, one based on raw intensities and the other on gradients (radial differences and angular differences). Random functions of the sorted local features are taken to obtain SRP features. It was shown that SRP outperformed RP significantly for robust texture classification \cite{liu2011sorted,LiuFieguthPR}, producing state of the art classification results on CUReT ($99.4\%$) KTHTIPS ($99.3\%$), and UMD ($99.3\%$) with a SVM classifier \cite{liu2011sorted,liu2015fusing}.

\textbf{(7) Local Binary Patterns} of Ojala \emph{et al.} \cite{Ojala1996Comparative} marked the beginning of the LBP methodology, followed by the simpler rotation invariant version of Pietik\"{a}inen \emph{et al.} \cite{Pietikainen2000Rotation},  and later ``uniform'' patterns to reduce feature dimensionality \cite{Ojala02}.

\begin {figure}[!t]
\centering
\includegraphics[width=0.5\textwidth]{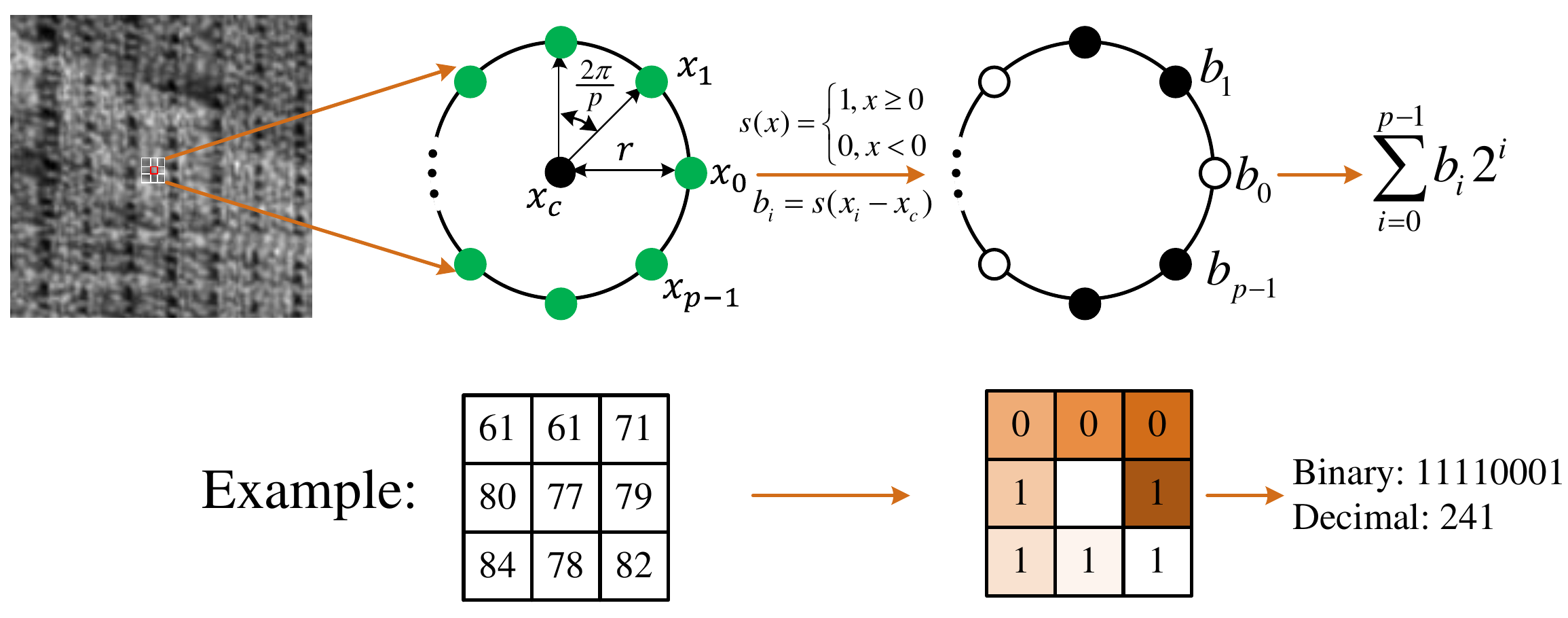}
\caption{A circular neighborhood used to derive an LBP code: a central pixel $x_c$ and its $p$ circularly and evenly spaced neighbors on a circle of radius $r$.}
\label{fig:LBPSampling}
\end {figure}

Texture representation generally requires the analysis of patterns in local pixel neighborhoods, which are comprehensively described by their joint distribution. However, stable estimation of joint distributions is often infeasible, even for small neighborhoods, because of the combinatorics of joint distributions.  Considering the joint distribution:
\begin{equation}\label{eqn:lbpjoint}
    g({x}_{c}, {x}_{0},\ldots, {x}_{p-1})
\end{equation}
of center pixel ${x}_{c}$ and $\{{x}_n\}_{n=0}^{p-1}$, $p$ equally spaced pixels on a circle of radius $r$, Ojala \emph{et al.} \cite{Ojala02} argued that much of the information in this joint distribution is conveyed by the joint distribution of differences:
\begin{equation}\label{eqn:diffjoint}
    g({x}_{0}-{x}_{c}, {x}_{1}-{x}_{c},\ldots, {x}_{p-1}-{x}_{c}).
\end{equation}
The size of the joint histogram was greatly minimized by keeping only the {\em sign} of each difference, as illustrated in Fig.~\ref{fig:LBPSampling}.

A certain degree of rotation invariance is achieved by cyclic shifts of the LBPs, \emph{i.e.,} grouping together those LBPs that are actually rotated versions of the same underlying pattern. Since the dimensionality of the representation (which grows exponentially with $p$) is still high, Ojala \emph{et al.} \cite{Ojala02} introduced a uniformity measure to identify $p(p-1)+2$ uniform LBPs and classified all remaining nonuniform LBPs under a single group. By changing parameters $p$ and $r$, we can derive LBP for any quantization of the angular
space and for any spatial resolution, such that multiscale analysis can be accomplished by combining multiple operators of varying $r$.  The most prominent advantages of LBP are its invariance to monotonic gray scale change, very low computational complexity, and ease of implementation.

\begin {figure}[!t]
\centering
\includegraphics[width=0.5\textwidth]{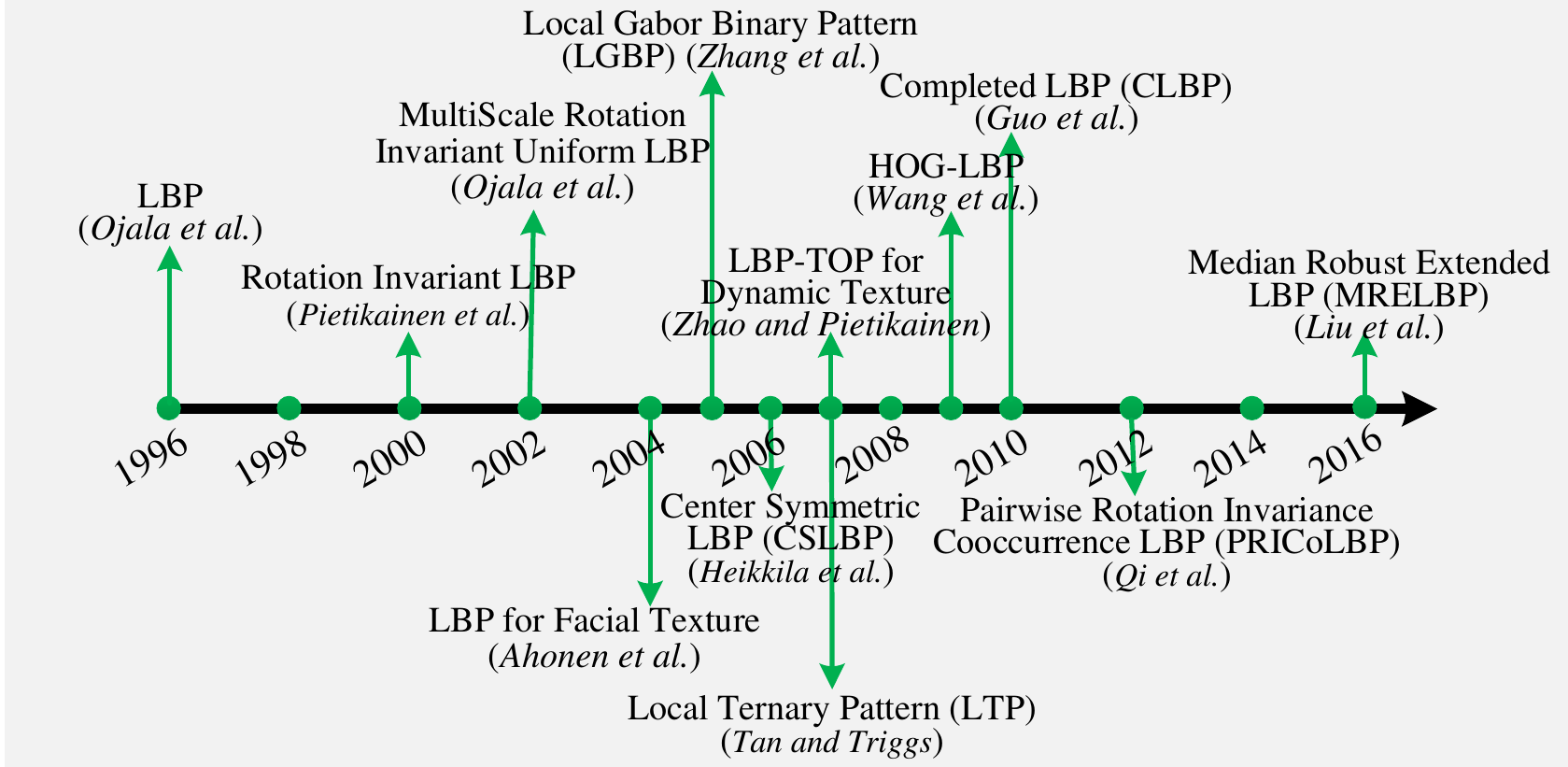}
\caption{LBP and its representative variants (see text for discussion).}
\label{fig:LBPhistory}
\end {figure}

Since \cite{Ojala02}, LBP started to receive increasing attention in computer vision and pattern recognition, especially texture and facial analysis, with the LBP milestones presented in Fig.~\ref{fig:LBPhistory}. As Gabor filters and LBP provide complementary information (LBP captures small and fine details, Gabor filters encode appearance information over a broader range of scales),  Zhang \emph{et al.} \cite{Zhang2005Local} proposed Local Gabor Binary Pattern (LGBP) by extracting LBP features from images filtered by Gabor filters of different scales and orientations, to enhance the representation power, followed by subsequent Gabor-LBP approaches \cite{Huang2011local,liu2017local,Pietikainen11}. Additional important LBP variants include LBP-TOP, proposed by Zhao and Pietik\"{a}inen \cite{Zhao07}, a milestone in using LBP for dynamic texture analysis; the Local Ternary Patterns (LTP) of Tan and Triggs \cite{Tan2007Enhanced}, introducing a pair
of thresholds and a split coding scheme which allows for encoding pixel similarity; the Local Phase Quantization (LPQ)
by Ojansivu \emph{et al.} \cite{Ojansivu2008Blur,Ojansivu2008Rotation} quantizing the Fourier transform phase
in local neighborhoods which is, by design, tolerant to most common types of
image blurs; the Completed LBP (CLBP) of Guo \emph{et al.} \cite{Guo10TIP},
encoding not only the signs but also the magnitudes of local differences; and the Median Robust Extended LBP (MRELBP) of Liu \emph{et al.} \cite{LiuTIP16} which enjoys high distinctiveness, low computational complexity, and strong robustness to image rotation and noise.

LBP has also led to compact and efficient binary feature descriptors designed for image matching, with noticeable ones including Binary Robust Independent Elementary Features (BRIEF) \cite{Calonder12BRIEF}, Oriented FAST and Rotated BRIEF (ORB) \cite{Rublee2011ORB}, Binary Robust Invariant Scalable Keypoints (BRISK) \cite{Leutenegger11BRISK} and Fast Retina Keypoint (FREAK) \cite{Alahi12FREAK}. These binary descriptors provide a comparable matching performance with the widely used region descriptors such as SIFT \cite{lowe2004distinctive} and SURF \cite{Bay2006SURF}, but are fast to compute and have significantly lower memory requirements, especially suitable for applications on resource constrained devices.

In summary, for large datasets with rotation variations and no significant illumination related variations, LBP \cite{Ojala02} could serve as an effective and efficient approach for texture classification. However, in the presence of significant illumination variations, significant affine transformations, or noise corruption, LBP
fails to meet the expected level of performance. MRELBP \cite{LiuTIP16}, a recent LBP variant, has been demonstrated to outperform LBP significantly, with near perfect classification performance on two small benchmark datasets (Outex\_TC10 $100\%$ and Outex\_TC12 $99.8\%$) \cite{LiuTIP16}, and which obtained the best overall performance in a recent experimental survey \cite{liu2017local} evaluating robustness in multiple classification challenges.  In general, LBP-based features work well in situations when limited training data are available; learning
based approaches like MR8, Patch Descriptors and DCNN based representations, which require large amount of training samples, are significantly outperformed by LBP based ones.

After over 20 years of developments, LBP is no longer just a simple texture operator, but has laid the foundation for a direction of research dealing with local image and video descriptors. A large number of LBP variants have been proposed to improve its robustness and to increase its discriminative power and applicability to different types of problems, and  interested readers are referred to excellent surveys \cite{Huang2011local,liu2017local,Pietikainen11}. Recently, although CNN based methods are beginning to dominate, LBP research remains active, as evidenced by significant recent work \cite{Guo16,sulc2014fast,Ryu15,Levi2015emotion,Lu2017Simultaneous,Juefei2016Local,Zhai2015Face,Ding2016Multi}.

\begin {figure}[!t]
\centering
\includegraphics[width=0.48\textwidth]{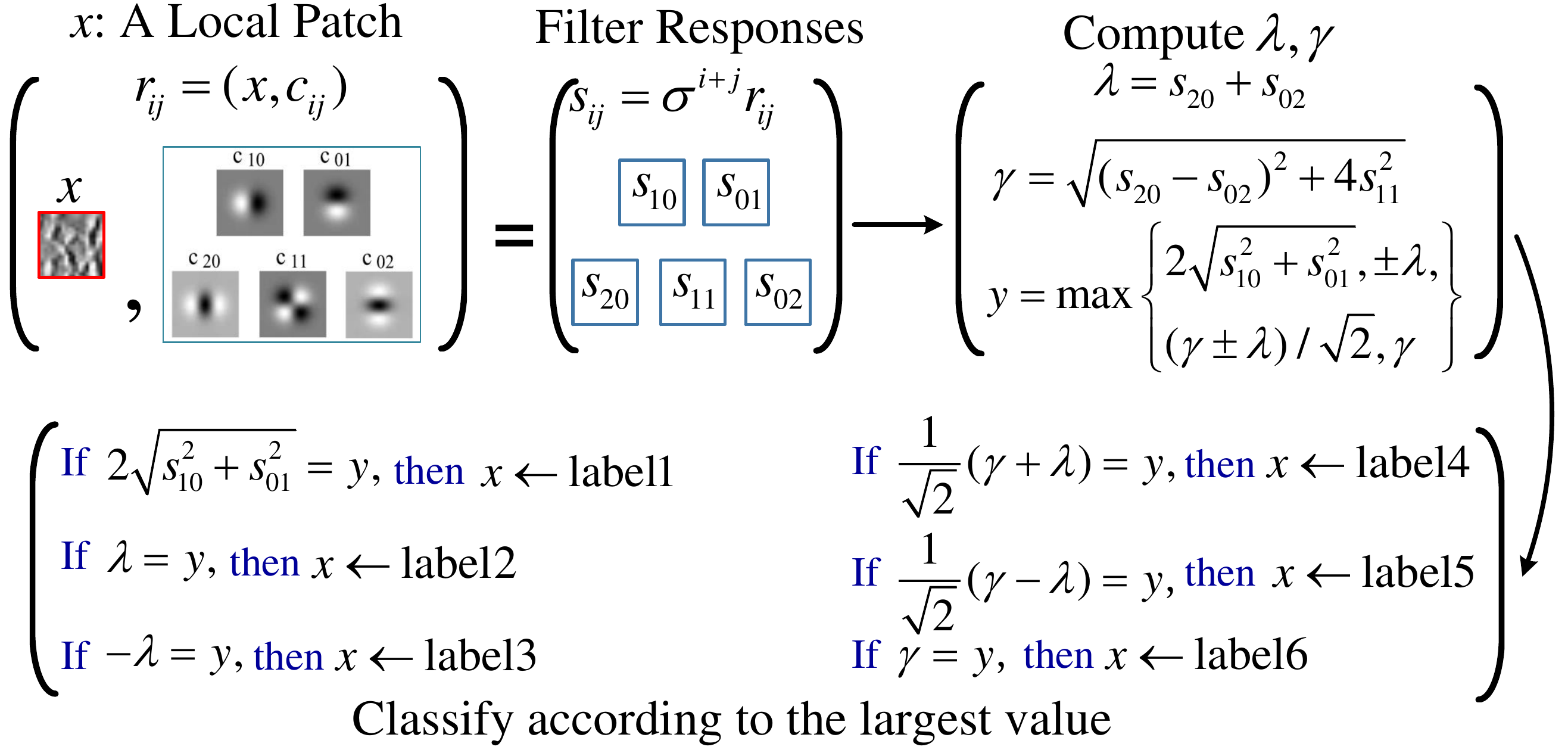}
\caption{Illustration of the calculation of BIF features.}
\label{fig:BIF}
\end {figure}

\textbf{(8) Basic Image Features (BIF)} approach \cite{Crosier10} is similar to LBP \cite{Ojala02}, in that it is based upon a predefined codebook rather than one learned from training. It therefore shares the advantages of LBP over methods based on codebook learning with clustering. In contrast with LBP, BIF probes an image locally using Gaussian derivative filters \cite{griffin2010symmetry,griffin2009basic} whereas LBP computes the differences between a pixel and its neighbors.  Derivative of Gaussians (DtG), consisting of first and second order derivatives of the Gaussian filter, can effectively detect the local basic and symmetry structure of an image, and allows achieving exact rotation invariance \cite{Freeman1991Design}.
BIF feature extraction is summarized in Fig.~\ref{fig:BIF}: each pixel in the image is filtered by the DtG filters, and then labeled as the maximizing class.  A simple six dimensional BIF histogram can be used as a global texture representation, however the histogram over these six categories produces too coarse a representation, therefore others (\emph{e.g.}, Crosier and Griffin \cite{Crosier10}) have performed multiscale analysis and calculated joint histograms over multiple scales. Multiscale BIF features achieved very good classification performance on CUReT ($98.6\%$), UIUC ($98.8\%$) and KTHTIPS ($98.5\%$) \cite{Crosier10}, with a NNC classifier.

\begin {figure}[!t]
\centering
\includegraphics[width=0.1\textwidth]{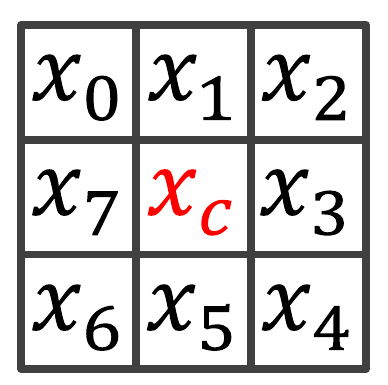}
\caption{First order square symmetric neighborhood for WLD computation.}
\label{fig:WLD}
\end {figure}

\textbf{(9) Weber Law Descriptor (WLD)} \cite{Chen09} is based on the fact that human perception of a pattern depends not only on the change of a stimulus but also on the original intensity of the stimulus.  The WLD consists of two components: differential excitation and orientation. For a small patch of size $3\times3$, shown in Fig.~\ref{fig:WLD}, the differential excitation is the relative intensity ratio
\begin{equation*}
\xi(x_{c}) = \textrm{arctan}\left(\frac{\sum_{i=0}^{7}{(x_{i}-x_{c}})}{x_{c}}\right)
\end{equation*}
and the orientation component is derived from the local gradient orientation
\begin{equation*}
\theta(x_{c})=\textrm{arctan}\frac{x_{7}-x_{3}}{x_{5}-x_{1}}.
\end{equation*}
Both $\xi$ and $\theta$ are quantified into a 2D histogram, offering a global representation. Clearly the use of multiple neighborhood sizes supports a multiscale generalization. Though computationally efficient, WLD features fail to meet the expected level of performance for texture recognition.

\subsubsection{Fractal Based Descriptors}
\label{subsubsec:Fractal}
Fractal Based Descriptors present a mathematically well founded alternative to dealing with scale \cite{Mandelbrot1983Fractal}, however they have not become popular as texture features due to their lack of discriminative power \cite{Varma2007Locally}. Recently, inspired by the BoW approach, researchers revisited the fractal method and proposed the MultiFractal Spectrum (MFS) method \cite{xu2009viewpoint,xu2009combining,Xu10}, invariant to viewpoint changes, nonrigid deformations and local affine illumination changes.

The basic MFS method was proposed in \cite{xu2009viewpoint}, where MFS was first defined for simple image features, such as intensity, gradient and Laplacian of Gaussian (LoG).  A texture image is first transformed into $n$ feature maps such as intensity, gradient or LoG filter features.  Each map is clustered into $k$ clusters (\emph{i.e.} $k$ codewords) via \emph{k}means. Then, a codeword label map is obtained and is decomposed into $k$ binary feature maps: those pixels assigned to codeword $i$ are labeled with 1 and the remainder as 0. For each binary feature map, the box counting algorithm \cite{Xu10} is used to estimate a fractal dimension feature. Thus, a total of $k$ fractal dimension features are computed for each feature map, forming a $k$D feature vector (referred to as a fractal spectrum) as the global representation of the image. Finally, for the $n$ different feature maps, $n$ fractal spectrum feature vectors are concatenated as the MFS feature. The MFS representation demonstrated invariance to a number of geometrical changes such as viewpoint changes, nonrigid surface changes and reasonable robustness to illumination changes. However, since
it is based on simple features (intensities and gradients) and has very low dimension,
it has limited discriminability, and gives classification rates $92.3\%$ and $93.9\%$ on datasets UIUC and UMD respectively.

Later MFS was improved by generalizing the simple image intensity and gradient features with SIFT \cite{xu2009combining}, wavelets \cite{Xu10}, and LBP \cite{quan2014lacunarity}. For instance, the Wavelet based MFS (WMFS) features archived significantly improved classification performance on UIUC ($98.6\%$) and UMD ($98.7\%$). The downside of the MFS approach is that it requires high resolution images to obtain sufficiently stable features.

\subsection{\textbf{Codebook Generation}}
\label{Sec:CodebookGene}
Texture characterization requires the analysis of spatially repeating patterns, which suffice to characterize textures and the pursuit of which has had important implications in a series of practical problems, such as dimensionality reduction, variable decoupling, and biological modelling \cite{Olshausen1997Sparse,Zhu2005Textons}. The extracted set of local texture features is versatile, and yet overly redundant \cite{leung2001representing}. It can therefore be expected that a set of prototype features (\emph{i.e.} codewords or textons) must exist which can be used to create global representations of textures in natural images \cite{leung2001representing,Okazawa2015Image,Zhu2005Textons}, in a similar way as in speech and language (such as words, phrases and sentences).

There exist a variety of methods for codebook generation. Certain approaches, such as LBP \cite{Ojala02} and BIF \cite{Crosier10}, which we have already discussed, use predefined codebooks, therefore entirely bypassing the codebook learning step.

For approaches requiring a learned codebook, \emph{k}means clustering \cite{Lazebnik05,leung2001representing,LiuFieguthPAMI,Varma09,Zhang07} and Gaussian Mixture Models (GMM) \cite{Cimpoi14,Cimpoi2016deep,Lategahn10,Jegou2012Aggregating,Perronnin10,sharma2016local} are the most popular and successful methods. GMM modeling considers both cluster centers and covariances,
which describe the location and spread/shape of clusters, whereas \emph{k}means clustering cannot capture overlapping distributions in the feature space as it considers only distances to cluster centers, although generalizations to \emph{k}means with multiple prototypes per cluster can allow this limitation to be relaxed. The GMM and \emph{k}means  methods learn a codebook in an unsupervised manner, but some recent approaches focus on building more discriminative ones \cite{Yang2008Unifying,Winn2005Object}.

In addition, another significant research thread is reconstruction based codebook learning \cite{aharon2006rm,peyre2009sparse,Skretting2006Texture,wang2010locality}, under the assumption that natural images admit a sparse decomposition
in some redundant basis (\emph{i.e.,} dictionary or codebook). These methods focus on learning nonparametric redundant dictionaries that facilitate a sparse representation of the data and minimize the reconstruction error of the data.  Because discrimination is the primary goal of texture classification, researchers have proposed to construct discriminative dictionaries
that explicitly incorporate category specific information \cite{mairal2008discriminative,mairal2009supervised}.

Since the codebook is used as the basis for encoding feature vectors, codebook generation is often interleaved with feature encoding, described next.

\begin {figure}[!t]
\centering
\includegraphics[width=0.49\textwidth]{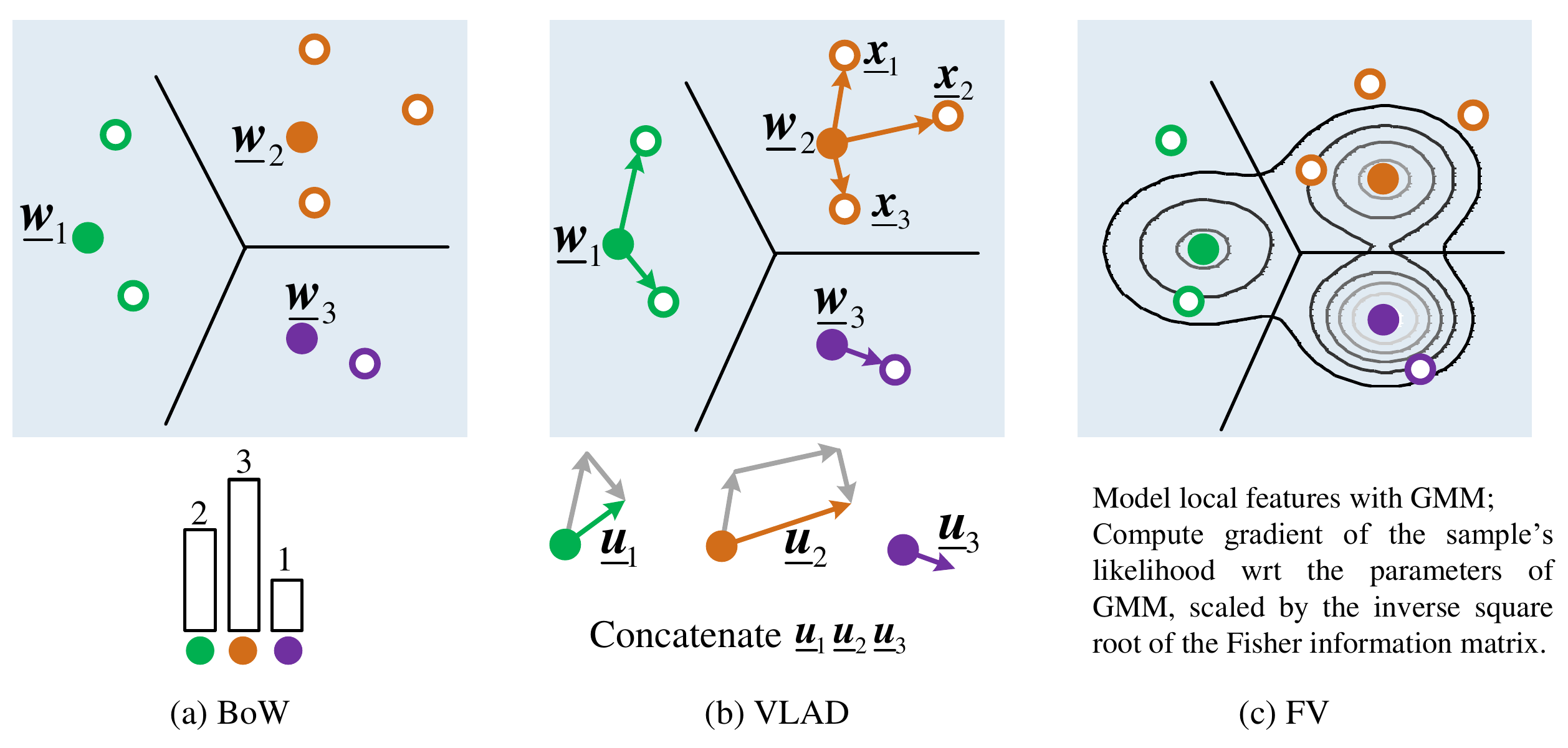}
\caption{Contrasting the ideas of BoW, VLAD and FV. (a) BoW: Counting the number of local features assigned to
each codeword. It encodes the zero order statistics of the distribution of local descriptors. (b) VLAD: Accumulating the differences of local features assigned to each codeword. (c) FV: The Fisher vector extends the BOW by encoding higher order statistics (first and second order), retaining information about the fitting error of the best fit.}
\label{fig:BoWvsVLADvsFV}
\end {figure}
\subsection{\textbf{Feature Encoding}}
\label{Sec:FeatEncode}

As illustrated in Fig.~\ref{fig:TextureRepresentation}, a given image is transformed into a pool of local texture features, from which a global image representation is derived by feature encoding with the generated codebook.
In the field of texture classification, we group commonly-used encoding strategies into three
major categories:
\begin{itemize}
\renewcommand{\labelitemi}{$\circ$}
\item Voting based  \cite{leung2001representing,Varma05,Van2008Kernel,Van2010Visual},
\item Fisher Vector based \cite{Jegou2012Aggregating,Cimpoi2016deep,Perronnin10,Sanchez13}, and
\item Reconstruction based \cite{mairal2008discriminative,mairal2009supervised,Olshausen1996Emergence,peyre2009sparse,wang2010locality}.
\end{itemize}
Comprehensive comparisons of encoding methods in image classification can be found in \cite{Chatfield2011Devil,Cimpoi14,huang2014feature}.

\textbf{Voting based methods.} The most intuitive way to quantize a local feature is to assign it to its nearest codeword in the codebook, also referred to as hard voting  \cite{leung2001representing,Varma05}.  A histogram of the quantized local descriptors can be computed by counting the number of local features assigned to each codeword; this histogram constitutes the baseline BoW representation (as illustrated in Fig.~\ref{fig:BoWvsVLADvsFV} (a)) upon which other methods can improve. Formally, it starts by learning a codebook $\{{\textbf{\emph{w}}}_i\}_{i=1}^{K}$, usually by \emph{k}means clustering.
Given a set of local texture descriptors $\{{\textbf{\emph{x}}}_i\}_{i=1}^{N}$ extracted from an image, the encoding representation of some descriptor ${\textbf{\emph{x}}}$ via hard voting is
\begin{eqnarray}\label{eqn:hardvote}
{\textbf{\emph{v}}}(i)=\left\{\begin{array}{ll}
1, & \textrm{if} \;\; i=\textrm{argmin}_{j}(\|{\textbf{\emph{x}}}-{\textbf{\emph{w}}}_j\|)\\
0, & \textrm{otherwise}.
\end{array}\right.
\end{eqnarray}
The histogram of the set of local descriptors is to aggregate all encoding vectors $\{{\textbf{\emph{v}}}_i\}_{i=1}^{N}$ via sum pooling. Hard voting overlooks codeword uncertainty, and may label image features by nonrepresentative codewords. In an improvement to this hard voting scheme, soft voting \cite{Ahonen2007Soft,Ren2013Noise,Ylioinas2013Constructing,Van2008Kernel,Van2010Visual} employs several nearest codewords to encode each local feature in a soft manner, such that the weight of each assigned codeword is an inverse function of the distance from the feature, for some kernel definition of distance. Voting based methods yield a histogram representation
of dimensionality $K$, the number of bins in the histogram.

\textbf{Fisher Vector based methods.} By counting the number of occurrences of codewords,
the standard BoW histogram representation encodes the zeroth-order statistics of the distribution of
descriptors, which is only a rough approximation of the probability density distribution of the local features. The Fisher vector extends the histogram approach by encoding additional information
about the distribution of the local descriptors. Based on the original FV encoding  \cite{Perronnin2007Fisher}, improved versions were proposed \cite{Cinbis2016Approximate,Perronnin10} such as the Improved FV (IFV) \cite{Perronnin10} and VLAD \cite{Jegou2012Aggregating}.
We briefly describe IFV \cite{Perronnin10} here, since to the best of
our knowledge it achieves the best performance in texture classification \cite{Cimpoi14,Cimpoi15,Cimpoi2016deep,sharma2016local}. Theory and practical issues regarding FV encoding can be found in \cite{Sanchez13}.

IFV encoding learns a soft codebook with GMM, as shown in Fig.~\ref{fig:BoWvsVLADvsFV} (c). An IFV encoding of a local feature is computed by assigning it to each codeword, in turn, and computing the gradient of the soft assignment
with respect to the GMM parameters\footnote{The derivative to weights, which is considered to make little contribution to the performance, is removed in IFK \cite{Perronnin10}.}. The IFV encoding dimensionality is $2DK$, where $D$ is the dimensionality of the feature space and $K$ is the number of Gaussian mixtures.  BoW can be considered a special case of FV in the case where the gradient computation is restricted to the mixture weight parameters
of the GMM. Unlike BoW, which requires a large codebook size, FV can be computed from a much smaller codebook (typically 64 or 256) and therefore at a lower computational cost at the codebook learning step.  On the other hand, the resulting dimension of the FV encoding vector (\emph{e.g.} tens of thousands) is usually significantly higher than BoW  (thousands), which makes it unsuitable for nonlinear classifiers, however it offers good performance even with simple linear classifiers.

The VLAD encoding scheme proposed by J\'{e}gou \emph{et al.} \cite{Jegou2012Aggregating}
can be thought of as a simplified version of FV, in that it typically uses \emph{k}means, rather than GMM, and records only first-order statistics rather than second order.  In particular, it records the residuals (the difference between the local features and the codewords), as shown in Fig.~\ref{fig:BoWvsVLADvsFV} (b).

\begin {figure}[!t]
\centering
\includegraphics[width=0.45\textwidth]{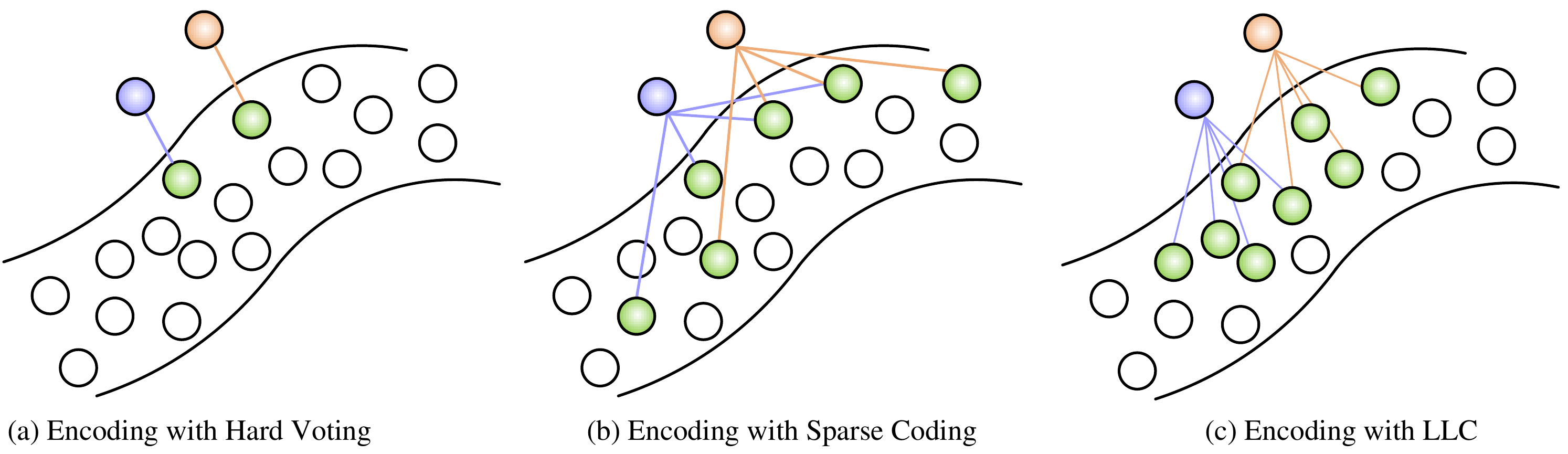}
\caption{Contrasting the ideas of hard voting, sparse coding, and LLC. }
\label{fig:SparseCodingvsLLC}
\end {figure}
\textbf{Reconstruction based methods.}
Reconstruction based methods aim to obtain an information-preserving encoding vector that allows for the reconstruction of a local feature with a small number of codewords. Typical methods include sparse coding and Local constraint Linear Coding (LLC), which are contrasted in Fig.~\ref{fig:SparseCodingvsLLC}. Sparse coding was initially proposed \cite{Olshausen1996Emergence} to model natural image statistics, then to texture classification \cite{Dahl2011Learning,mairal2008discriminative,mairal2009supervised,peyre2009sparse,Skretting2006Texture} and later to other problems such as image classification \cite{Yang2009Linear} and face recognition \cite{Wright2009Robust}.

In sparse coding, a local feature ${\textbf{\emph{x}}}$ can be well
approximated by a sparse decomposition ${\textbf{\emph{x}}}\approx\textbf{W}{\textbf{\emph{v}}}$ over the learned codebook $\textbf{W}=[{\textbf{\emph{w}}}_1,{\textbf{\emph{w}}}_2,
\ldots{\textbf{\emph{w}}}_K]$, by leveraging the sparse nature of the underlying image \cite{Olshausen1996Emergence}. A sparse encoding can be solved as
\begin{equation}\label{eqn:SparseCoding}
    \textrm{argmin}_{\textbf{\emph{v}}}{\|\textbf{\emph{x}}-
    \mathbf{W}\textbf{\emph{v}}\|^2_2}\quad s.t.
    \quad\|\textbf{\emph{v}}\|_{0}\leq s.
\end{equation}
where $s$ is a small integer denoting the sparsity level, limiting the number of nonzero entries in ${\textbf{\emph{v}}}$, measured as $\|{\textbf{\emph{v}}}\|_0$. Learning a redundant codebook that facilitate a sparse representation of the local features is important in sparse coding \cite{aharon2006rm}.
Methods in \cite{mairal2008discriminative,mairal2009supervised,peyre2009sparse,Skretting2006Texture} are based on learning $C$ class-specific codebooks, one for each texture class and approximating each local feature using a constant sparsity $s$. The $C$ different codebooks provides $C$ different reconstruction
errors, which can then be used as classification features. In \cite{peyre2009sparse,Skretting2006Texture}, the class specific codebooks were optimized for reconstruction, but significant
improvements have been shown by optimizing for discriminative power instead \cite{Dahl2011Learning,mairal2008discriminative,mairal2009supervised}, an approach which is, however, associated with high computational cost, especially when the number of texture classes $C$ is large.

Locality constrained linear coding (LLC) \cite{wang2010locality}
projects each local descriptor ${\textbf{\emph{x}}}$ down to the \emph{local} linear subspace spanned by $q$ codewords in the codebook of size $K$ closest to it (in Euclidean distance), resulting in a $K$ dimensional encoding vector whose entries are all zero except for the indices of the $q$ codewords closest to ${\textbf{\emph{x}}}$. The projection of ${\textbf{\emph{x}}}$ down to the span of its $q$ closest codewords is solved via
\begin{eqnarray*}\label{eqn:LLC}
    \textrm{argmin}_{\textbf{\emph{v}}}{\|\textbf{\emph{x}}-
    \mathbf{W}\textbf{\emph{v}}\|^2_2} + \lambda \sum_{k=1}^K{\left(\textbf{\emph{v}}(i)exp\frac{\|\textbf{\emph{x}}-\textbf{\emph{w}}_i\|_2}{\sigma}\right)^2}
    \\
    s.t.  \quad\sum_{k=1}^K{\textbf{\emph{v}}(i)}=1,
\end{eqnarray*}
where $\lambda$ is a small regularization constant and $\sigma$ adjusts the weight decay speed.

In summary, reconstruction based coding has received significant attention since sparse coding was applied for visual classification \cite{mairal2008discriminative,mairal2009supervised,peyre2009sparse,Skretting2006Texture,wang2010locality}.
A theoretical study for the success of sparse coding
over vector quantization can be found in \cite{Coates2011Importance}.

\begin {figure*}[!t]
\centering
\includegraphics[width=0.95\textwidth]{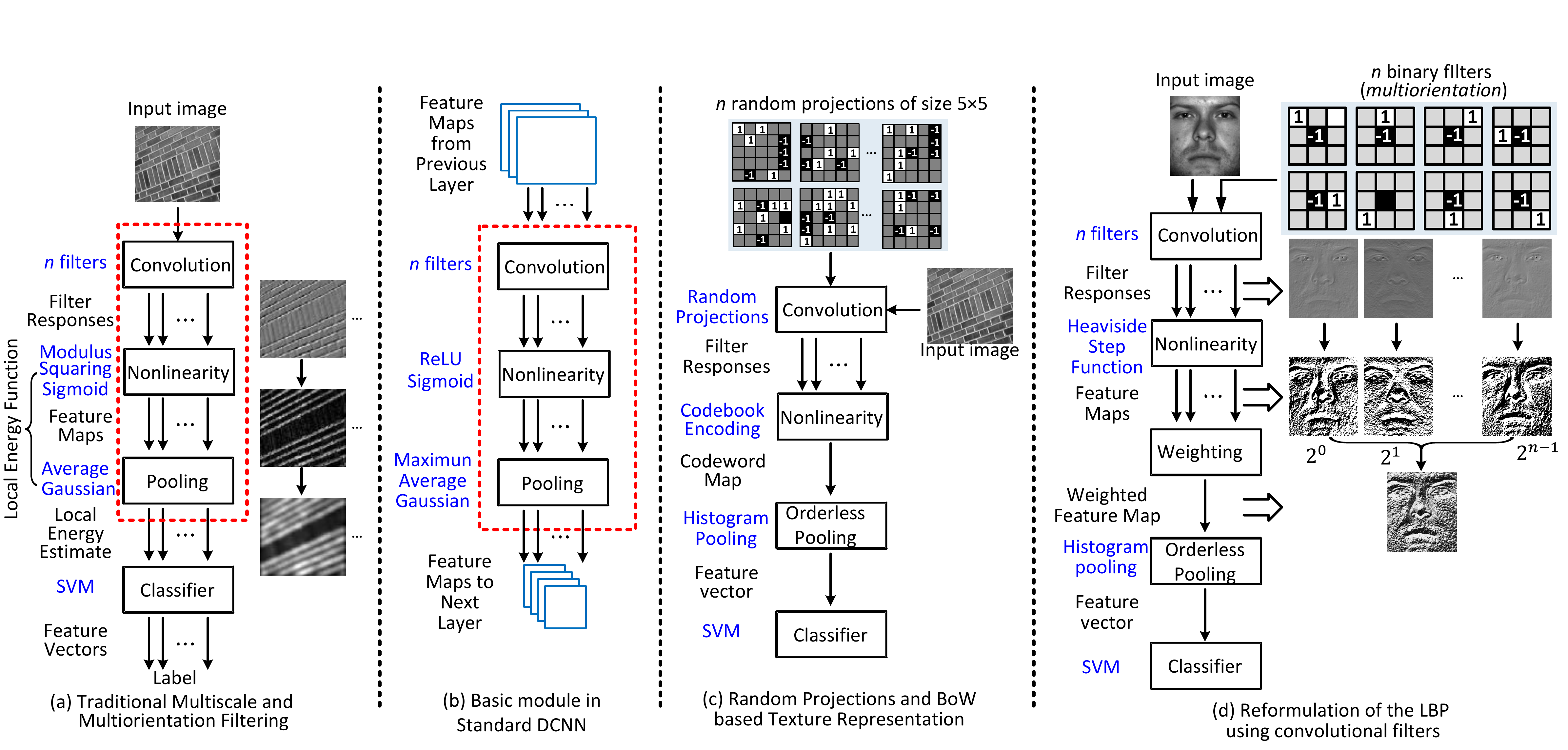}
\caption{Contrasting classical filtering based texture features, CNN, BoW and LBP.}
\label{fig:FrameWorkCompare}
\end {figure*}

\subsection{\textbf{Feature Pooling and Classification}}
\label{Sec:FeatPool}
The goal of feature pooling \cite{Boureau2010Theoretical} is to integrate or combine the coded feature vectors $\{\textbf{\emph{v}}_i\}_i,\textbf{\emph{v}}_i\in\mathbb{R}^{d}$ of a given image into a final compact global representation ${\textbf{\emph{y}}}_i$ which is more robust to image transformations and noise. Commonly used pooling methods include sum pooling, average pooling and max pooling \cite{leung2001representing,Varma09,wang2010locality}. Boureau \emph{et al.} \cite{Boureau2010Theoretical} presented a theoretical analysis of
average pooling and max pooling, and showed that max pooling may be well suited to
sparse features. The authors also proposed softer max pooling methods by using a smoother estimate of the expected max-pooled feature and demonstrated improved performance. Another noticeable pooling method is the mix-order max pooling method which considers the information of visual word occurrence frequency \cite{Liu2011Defense}.

Specifically, let $\textbf{V}=[\textbf{\emph{v}}_1,...,\textbf{\emph{v}}_N]\in\mathbb{R}^{d\times N}$ denote the coded features from $N$ locations. For $\textbf{\emph{u}}$ denoting a row of $\textbf{V}$, $\textbf{\emph{u}}$ is reduced to a single scalar by some operation (sum, average, max), reducing $\textbf{V}$ to a \emph{d}-dimensional feature vector. Realizing that pooling over the entire image disregards all information regarding spatial dependencies, Lazebnik \emph{et al.} \cite{lazebnik2006beyond} proposed a simple Spatial Pyramid Pooling (SPM) scheme by partitioning the image into increasingly fine subregions and computing histograms of
local features found inside each subregion via average or max pooling. The final global representation is a concatenation of all histograms extracted from subregions, resulting in a higher dimensional representation that preserves more spatial information \cite{timofte2012training}.

Given a pooled feature, a given texture sample can be classified. Many classification approaches are possible \cite{Jain2000statistical,Webb2011}, although Nearest Neighbor Classifier (NNC) and Support Vector Machine (SVM) are the most widely-used classifiers for the BoW representation.
Different distance measures may be used, such as the EMD distance \cite{Lazebnik05,Zhang07}, KL divergence and the widely-used Chi Square distance \cite{LiuFieguthPAMI,Varma09}. For high dimensional BoW features, as with SPM features and multilevel histograms, histogram intersection kernel SVM \cite{Grauman2005Pyramid,lazebnik2006beyond,Maji2008Classification} is a good and efficient choice.
For very high-dimensional features, as with IFV and VLAD, linear SVM may represent a better choice \cite{Jegou2012Aggregating,Perronnin10}.

\begin{table*}[!t]
\caption {CNN based texture representation}\label{Tab:MethodCNN}
\centering
\renewcommand{\arraystretch}{1}
\setlength\arrayrulewidth{0.2mm}
\setlength\tabcolsep{2pt}
\resizebox*{17cm}{!}{
\begin{tabular}{lp{10cm}}
\Xhline{1.5pt}
\scriptsize Approach & \scriptsize Highlights \\
\Xhline{1.5pt}
\scriptsize  Using Pretrained Generic CNN Models \cite{Cimpoi2016deep} (Section \ref{sec:Pretrained})
& \scriptsize Traditional feature encoding and pooling;
New pooling such as bilinear pooling \cite{lin2016visualizing,Lin2017Bilinear} and LFV \cite{Song2017Locally} \\
\hline
\scriptsize  $\quad\bullet$ AlexNet \cite{Krizhevsky12}
& \scriptsize Achieved breakthrough image classification result on ImageNet;
The historical turning point of feature representation from handcrafted to CNN.  \\
\hline
\scriptsize  $\quad\bullet$ VGGM \cite{chatfield2014return,Cimpoi2016deep}
& \scriptsize Similar complexity as AlexNet, but better texture classification performance.  \\
\hline
\scriptsize  $\quad\bullet$ VGGVD \cite{simonyan2014very}
& \scriptsize Much deeper than AlexNet; Much Larger model size
than AlexNet and VGGM; Much better texture recognition performance than AlexNet and VGGM. \\
\scriptsize  $\quad\bullet$ GoogleNet \cite{GoogLeNet2015}
& \scriptsize Much deeper than AlexNet; Small pretrained model size; Not often used in texture classification.  \\
\hline
\scriptsize  $\quad\bullet$ ResNet \cite{He2016ResNet}   & \scriptsize Significantly deeper than VGGVD;
Smaller model size (ResNet 101) than AlexNet.  \\
\Xhline{1.5pt}
\scriptsize  Using Finetuned CNN Models (Section \ref{subsec:Finetuned})  & \scriptsize End-to-end learning \\
\hline
\scriptsize  $\quad\bullet$ TCNN \cite{Andrearczyk2016using}
& \scriptsize Using global average pooling; Combining outputs from multiple CONV layers.  \\
\hline
\scriptsize  $\quad\bullet$ BCNN \cite{lin2015bilinear,lin2016visualizing} & \scriptsize Introducing a novel and orderless bilinear feature pooling method; Generalizing Fisher Vector and VLAD; Good representation ability; Very high feature dimensionality. \\
\hline
\scriptsize  $\quad\bullet$ Compact BCNN \cite{Gao2016Compact}  & \scriptsize  Adopting Random Maclaurin Projection or Tensor Sketch Projection to reduce the dimensionality of bilinear features (\emph{e.g.} from 262144 ($512^2$) to 8192); Maintain similar performance to BCNN; \\
\hline
\scriptsize  $\quad\bullet$ FASON \cite{Dai2017FASON} & \scriptsize Combining the ideas of TCNN \cite{Andrearczyk2016using} and  Compact BCNN \cite{Gao2016Compact}. \\
\hline
\scriptsize  $\quad\bullet$ NetVLAD \cite{Arandjelovic2016netvlad}   & \scriptsize  Plugging a VLAD like layer in a CNN network at the last CONV layer. \\
\hline
\scriptsize  $\quad\bullet$ DeepTEN \cite{Zhang2017Deep}   & \scriptsize  Similar to NetVLAD \cite{Arandjelovic2016netvlad}, integrating an encoding layer on top of CONV layers; Generalizing orderless pooling methods such as VLAD and FV in a CNN trained end to end. \\
\Xhline{1.5pt}
\scriptsize  Texture Specific Deep Convolutional Models (Section \ref{subsec:TextureSpecific})   & \scriptsize \\
\hline
\scriptsize  $\quad\bullet$ ScatNet \cite{Bruna13} & \scriptsize Use Gabor wavelets for comvolution; Mathematical interpretation of CNNs; Features being stable to deformations and preserving high frequency information;  \\
\hline
\scriptsize  $\quad\bullet$ PCANet \cite{Chan15} & \scriptsize  Inspired by ScatNet \cite{Bruna13}, using PCA filters to replace Gabor wavelets;Using LBP and histogramming as feature pooling; No local invariance. \\
\Xhline{1.5pt}
\end{tabular}
}
\end{table*}

\section{CNN based Texture Representation}
\label{sec:CNN}
A large number of CNN-based texture representation methods have been proposed
in recent years since the record-breaking image classification result \cite{Krizhevsky12} achieved in 2012.
A key to the success of CNNs is their ability to leverage large labeled datasets to learn high quality features.
Learning CNNs, however, amounts to estimating millions of parameters and requires a very large number of
annotated images, an issue which rather constrains the applicability of CNNs in problems with limited training data.
A key discovery, in this regard, was that CNN features pretrained on very large datasets
were found to transfer well to many other problems, including texture analysis, with a relatively
modest adaptation effort \cite{chatfield2014return,Cimpoi2016deep,Girshick2014Rich,Oquab2014Learning,sharif2014cnn}.
In general, the current literature on texture classification includes examples of both employing pretrained
generic CNN models or performing finetuning
for specific texture classification tasks.

In this survey we will classify CNN based texture representation methods into three categories,
and which form the basis of the following three sections:
\begin{itemize}
\renewcommand{\labelitemi}{$\circ$}
\item using pretrained generic CNN models,
\item using finetuned CNN models, and
\item using handcrafted deep convolutional networks.
\end{itemize}
These representations have had a widespread influence in image understanding;
representative examples of each of these are given in Table \ref{Tab:MethodCNN}.

\subsection{\textbf{Using Pretrained Generic CNN Models}}
\label{sec:Pretrained}
Given the behavior of CNN transfer, the success of pretrained CNN models lies in the feature extraction and encoding steps. Similar to Section \ref{sec:BoW}, we will describe first some commonly used networks for
pretraining and then the feature extraction process.

\textbf{(1) Popular Generic CNN Models}
can serve as good choices for extracting features, including AlexNet \cite{Krizhevsky12},
VGGNet \cite{simonyan2014very}, GoogleNet \cite{GoogLeNet2015}, ResNet \cite{He2016ResNet} and DenseNet \cite{Huang2016Densely}. Among these networks, AlexNet was
proposed the earliest, and in general the others are deeper and more complex. A full review of these networks is beyond the scope of this paper, and we refer
readers to the original papers \cite{He2016ResNet,Huang2016Densely,Krizhevsky12,simonyan2014very,
GoogLeNet2015} and to excellent surveys \cite{bengio2013representation,chatfield2014return,Gu2017Recent,lecun2015deep,Liu2018DeepLearning} for additional details.
Briefly, as shown in Fig. \ref{fig:FrameWorkCompare} (b), a typical CNN repeatedly applies the following three operations:
\begin{enumerate}
\item Convolution with a number of linear filters,
\item Nonlinearities, such as sigmoid or rectification,
\item Local pooling or subsampling.
\end{enumerate}
These three operations are highly related to traditional filter bank methods widely used in texture analysis \cite{randen1999filtering}, as shown in Fig.~\ref{fig:FrameWorkCompare} (a), with the key differences that the CNN filters are learned directly from data rather than handcrafted, and that CNNs have a hierarchical architecture learning increasingly abstract levels of representation. These three operations are also closely related
to the RP approach (Fig.~\ref{fig:FrameWorkCompare} (c)) and the LBP
(Fig.~\ref{fig:FrameWorkCompare} (d)).

Several large-scale image datasets are usually used for CNN pretraining. Among them the commonly used ImageNet dataset, with 1000 classes and 1.2 million images \cite{russakovsky2015imagenet}, and the scene-centric MITPlaces dataset \cite{Zhou2014Learning,Zhou2017Places}.

Comprehensive evaluations of the feature transfer effect of CNNs for the purpose of  texture classification have been conducted in  \cite{Cimpoi14,Cimpoi15,Cimpoi2016deep,Napoletano2017}, with the following critical insights.  During model transfer, features extracted from different
layers exhibit different classification performance. Experiments
confirm that the fully-connected layers of the CNN, whose role is primarily that of classification, tend to exhibit relatively worse generalization ability and  transferability, and therefore would need retraining or finetuning on the transfer target. In contrast the convolutional layers, which act more as feature extractors, with coarser convolutional layers acting as progressively more abstract features, generally transfer well. That is, the convolutional descriptors are substantially less committed to a specific dataset than the fully connected descriptors.  As a result, the source training set is relevant to classification accuracy on different datasets, and the similarity of the source and target plays a critical role when using a pretrained CNN model \cite{bell2015material}.  Finally, from \cite{Cimpoi15,Cimpoi2016deep,Napoletano2017} it was found that deeper models transfer better, and that the deepest convolutional  descriptors give the best performance, superior to the fully-connected descriptors, when proper encoding techniques are employed (such as FVCNN$\leftarrow$CNN features with Fisher Vector encoder).

\textbf{(2) Feature Extraction:}
A CNN can be viewed as a composition $f_L\circ\cdots\circ f_2\circ f_1$ of $L$ \emph{layers}, where the output of each layer $\textbf{X}^l=(f_l\circ\cdots\circ f_2\circ f_1)(\textbf{I})$ consists of $D^l$ feature maps of size $W^l\times H^l$.
The $D^l$ responses at each spatial location form a $D^l$ dimensional feature vector. The network
is called convolutional if all the layers are implemented as
filters, in the sense that they act locally and uniformly
on their input. From bottom to top layers, the image undergoes convolution,
and the receptive field of these convolutional filters and the number of feature channels increases,
whereas the size of the feature maps decreases. Usually, the last several layers of a typical CNN are \emph{fully connected} (FC) because, if seen as filters, their support is
the same as the size of the input $\textbf{X}^{l-1}$, and therefore lack
locality.

The most straightforward approach to CNN based texture classification is to extract the descriptor from the fully connected layers of the
network \cite{Cimpoi15,Cimpoi2016deep}, \emph{e.g.}, the FC6 or FC7
descriptors in AlexNet \cite{Krizhevsky12}. The fully connected layers are pretrained discriminatively, which can be
either an advantage or a disadvantage, depending on whether the information that they captured can be transferred to the
domain of interest \cite{chatfield2014return,Cimpoi2016deep,Girshick2014Rich}. The fully connected descriptors have a global
receptive field and are usually viewed as global features suitable for classification with an SVM classifier.
In contrast, the convolutional layers of a CNN can be used as filter banks to extract local features \cite{Cimpoi15,Cimpoi2016deep,Gong2014Multi}. Compared with the global fully-connected descriptors, lower level convolutional descriptors are more robust to image transformations such as translation and occlusion. In \cite{Cimpoi15,Cimpoi2016deep}, the features
are extracted as the output of a convolutional layer, directly from the linear filters (excluding ReLU and max
pooling, if any), and are combined with traditional encoders for global representation. For instance, the last convolutional layer of VGGVD (very deep with 19 layers) \cite{simonyan2014very} yields a set of 512 descriptor vectors;
in \cite{Cimpoi14,Cimpoi15,Cimpoi2016deep} four types of CNNs were considered for feature extraction.

\begin {figure*}[!t]
\centering
\includegraphics[width=0.98\textwidth]{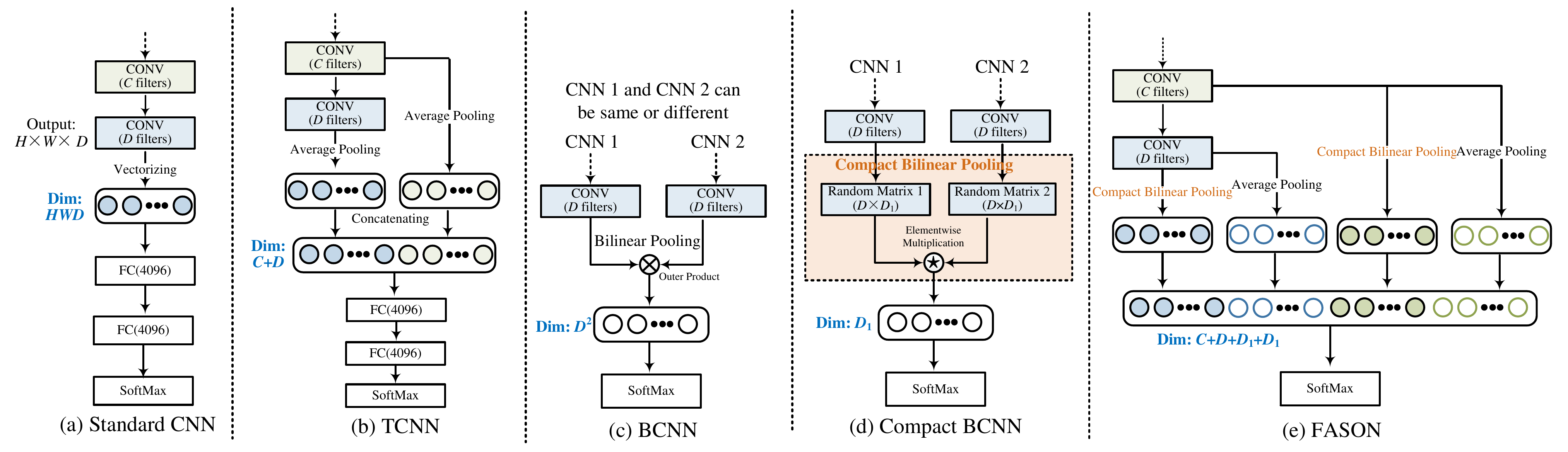}
\caption{Comparison of Fine Tuned CNNs: (a) standard CNN, (b) TCNN \cite{Andrearczyk2016using}, (c) BCNN \cite{Lin2017Bilinear}, (d) Compact Bilinear Pooling \cite{Gao2016Compact}, and (e) FASON \cite{Dai2017FASON}.}
\label{fig:FineTunedCNNs}
\end {figure*}

\textbf{(3) Feature Encoding and Pooling:}
A set of features extracted from convolutional or fully connected layers resembles a set of texture features as described in Section~\ref{Sec:LocalDes}, so the traditional feature encoding methods discussed in Section~\ref{Sec:FeatEncode} can be directly
employed.

\begin {figure}[!t]
\centering
\includegraphics[width=0.4\textwidth]{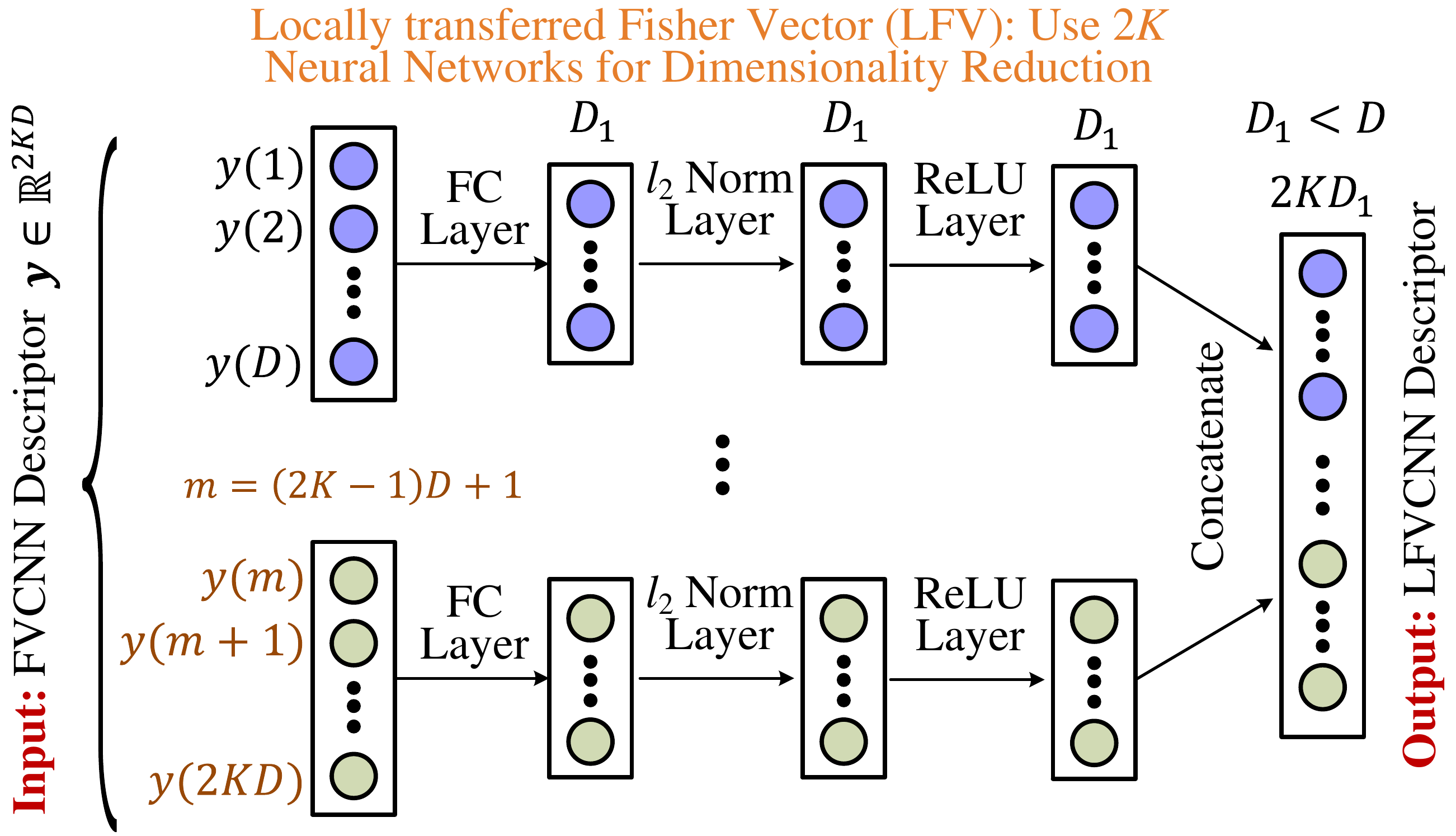}
\caption{Locally transferred Fisher Vector (LFV): use $2K$ neural networks for dimensionality reduction of FVCNN descriptor.}
\label{fig:LFV}
\end {figure}

In \cite{Cimpoi2016deep}, Cimpoi \emph{et al.} evaluated several encoders, \emph{i.e.} standard BoW \cite{leung2001representing}, LLC \cite{wang2010locality}, VLAD \cite{Jegou2012Aggregating} and IFV  \cite{Perronnin10} (reviewed in Section \ref{Sec:FeatEncode}), for CNN features, and showed that the best performance is achieved by IFV. It has been reported that VGGVD+IFV with a linear SVM classifier produced consistently near perfect classification performance on several texture datasets: KTHTIPS ($99.8\%$), UIUC ($99.9\%$, UMD ($99.9\%$) and ALOT ($99.5\%$)), as summarized in Table \ref{Tab:Results}. In addition, it obtained significant improvement on very challenging datasets like KTHTIPS2b ($81.8\%$), FMD ($79.8\%$) and DTD ($72.3\%$). However, it only achieved $80.0\%$ and $82.3\%$ on Outex\_TC10 and Outex\_TC12 respectively, which are significantly worse than the near perfect performance of MRELBP on these two datasets \cite{liu2017local}; a clear indicator that DCNN based features
require large amount of training samples and that they lack local invariance. Song \emph{et al.} \cite{Song2017Locally} proposed a neural network to transform the FVCNN descriptors into a lower dimensional representation. As shown in Fig.~\ref{fig:LFV}, locally transferred FVCNN (LFVCNN) descriptors are obtained by passing the $2KD$ dimensional FVCNN descriptors of images through a multilayer neural network consisting of fully connected, $l_2$ normalization layers, and ReLU layers.
LFVCNN achieved state of the art results on KTHTIPS2b ($82.6\%$), FMD ($82.1\%$) and DTD ($73.8\%$), as shown in Table \ref{Tab:Results}.

Recently, Gatys \emph{et al.} \cite{Gatys2015Texture} showed
that the Gram matrix representations extracted from various layers of VGGNet \cite{simonyan2014very} can be inverted for texture synthesis. The work of Gatys \emph{et al.} \cite{Gatys2015Texture} ignited a renewed interest in texture synthesis \cite{Ulyanov17Improved}. Notably, the Gram matrix representation
used in their approach is identical to the bilinear pooling of CNN features of Lin \emph{et al.} \cite{lin2015bilinear}, which were proved to be good for texture recognition in \cite{lin2016visualizing}. Like the traditional encoders introduced in Section \ref{Sec:FeatEncode}, the bilinear feature pooling is an orderless representation of the input image and hence
is suitable for modeling textures.  The Bilinear CNN (BCNN) descriptors are obtained by computing the outer product of each feature $\textbf{\emph{x}}^l_i$ with itself,  reordered into feature vectors, and subsequently pooled via sum to obtain the final global representation. The dimension of the bilinear descriptor is $(D^l)^2$, which is very high (\emph{e.g.} $512^2$).  It was shown in \cite{lin2016visualizing,Lin2017Bilinear} that the texture classification performance of BCNN and FVCNN was virtually identical, indicating that bilinear pooling is as good as the Fisher vector pooling for texture recognition. It was also found that the BCNN descriptor of the last convolutional layer performed the best, in agreement with \cite{Cimpoi2016deep}.

\subsection{\textbf{Using Finetuned CNN Models:}}
\label{subsec:Finetuned}
Pretrained CNN models, discussed in Section \ref{sec:Pretrained}, have achieved impressive performance in texture recognition, however training in these methods is a multistage pipeline that involves feature extraction, codebook generation, feature encoding and classifier training. Consequently, these methods cannot take advantage of utilizing the full capability of neural networks in representation learning. Generally finetuning CNN models on task-specific training datasets (or learning from scratch if large-scale task-specific datasets are available) is expected to improve on already strong performance achieved by pretrained CNN models \cite{chatfield2014return,Girshick2014Rich}. When using a finetuned CNN model, the global image representation is usually generated in an end-to-end manner; that is, the network will produce a final visual representation without additional explicit encoding or pooling steps, as illustrated in Fig.~\ref{fig:BoW}. When finetuning a CNN, the last fully connected layer is modified to have $B$ nodes corresponding to the number of classes in the target dataset.
The nature of the datasets used in finetuning is important to learning discriminative CNN features.  The pretrained CNN model is capable of discriminating images of different objects or scene classes, but may be less effective in discerning the difference between different textures (material types) since an image in ImageNet may contain different types of textures (materials). The size of the dataset used in finetuning matters as well, since too small a dataset may be inadequate for complete learning.

To the best of our knowledge, the behaviour of a finetuned large-scale CNN like VGGNet \cite{simonyan2014very} or training it from scratch using a texture dataset have not been fully explored, almost certainly due to the fact that a large texture dataset on the scale of ImageNet \cite{russakovsky2015imagenet} or MITPlaces \cite{Zhou2014Learning} does not exist. Most existing texture datasets are small, as discussed later in Section \ref{sec:Texturedatasets}, and according to \cite{Andrearczyk2016using,lin2016visualizing} finetuning a VGGNet \cite{simonyan2014very} or AlexNet \cite{Krizhevsky12} on existing texture datasets leads to negligible performance improvement.  As shown in Fig.~\ref{fig:FineTunedCNNs} (a), for a typical CNN like VGGNet \cite{simonyan2014very}, the output of the last convolutional layer is reshaped into a single feature vector (spatially sensitive) and fed into fully connected layers (\emph{i.e.,} order sensitive pooling). The global spatial information is necessary for analyzing the global shapes of objects, however it has been realized \cite{Andrearczyk2016using,Cimpoi2016deep,Gatys2015Texture,lin2016visualizing,Zhang2017Deep} that it is not of great importance for analyzing textures due to the need for orderless representation. The FVCNN descriptor shows higher recognition performance than FCCNN, even if the pretrained VGGVD model is finetuned on the
texture dataset (\emph{i.e.,} the finetuned FCCNN descriptor) \cite{Cimpoi2016deep,lin2016visualizing}. Therefore, an orderless feature pooling from the output of a convolution layer is desirable for end-to-end learning. In addition, orderless pooling does not require an input image to be of a fixed size, motivating a series of innovations in designing novel CNN architectures for texture recognition \cite{Andrearczyk2016using,Arandjelovic2016netvlad,Dai2017FASON,Lin2017Bilinear,
Zhang2017Deep}.

A Texture CNN (TCNN) based on AlexNet, as illustrated in Fig.~\ref{fig:FineTunedCNNs} (b), was developed in \cite{Andrearczyk2016using}. It simply utilizes global average pooling to transform a field of descriptor $\textbf{X}^l\in\mathbb{R}^{W^l\times H^l\times D^l}$ at a given convolutional layer $l$ of a CNN into a $D^l$ dimension vector which is connected to a fully connected layer. TCNN has fewer parameters and
lower complexity than AlexNet. In addition, Andrearczyk and Whelan \cite{Andrearczyk2016using} proposed to fuse the global average pooled vector of an intermediate convolutional layer and that of the last convolutional layer via concatenation and introduced to later fully connected layers, a combination which resembles the hypercolumn feature developed in \cite{Hariharan2015Hypercolumns}. Andrearczyk and Whelan \cite{Andrearczyk2016using} observed that finetuning a network that was pretrained on a texture-centric dataset achieves better results on other
texture datasets compared to a network pretrained on an object-centric dataset of the
same size, and that the size of the dataset on which the network is pretrained or finetuned predominantly influences the
performance of the finetuning. These two observations suggest that a very large texture dataset could bring a significant contribution to CNNs applied to texture analysis.

In BCNN \cite{Lin2017Bilinear}, as shown in Fig.~\ref{fig:FineTunedCNNs} (c), Lin \emph{et al.} proposed to replace the fully connected layers with an orderless bilinear pooling layer, which was discussed in Section \ref{sec:Pretrained}. This method was successfully applied to texture classification in \cite{lin2016visualizing} and obtained slightly better results than FVCNN, however the representational power of bilinear features
comes at the cost of very high dimensional feature representations, which induce substantial computational burdens and require large amounts of training data, motivating several improvements on BCNN. Gao \emph{et al.} \cite{Gao2016Compact} proposed compact bilinear pooling, as shown in Fig. \ref{fig:FineTunedCNNs} (d), which utilizes Random Maclaurin Projection or Tensor Sketch Projection to reduce the dimensionality of bilinear
representations while still maintaining similar performance to the full BCNN feature \cite{Lin2017Bilinear} with a $90\%$ reduction in the number of learned parameters. To combine the ideas in \cite{Andrearczyk2016using} and \cite{Gao2016Compact}, Dai \emph{et al.} \cite{Dai2017FASON} proposed an effective fusion network called FASON (First And Second Order information fusion Network) that combines first and second order information flow, as illustrated in Fig.~\ref{fig:FineTunedCNNs} (e). These two types of features were generated from different convolutional layers and concatenated to form a single feature vector which was connected to a fully connected softmax layer for end to end training.
In \cite{Kong2017Low}, Kong and Fowlkes proposed to represent
the bilinear features as a matrix and applied a low rank bilinear classifier. The resulting classifier can be evaluated
without explicitly computing the bilinear feature map
which allows for a large reduction in the computational time as
well as decreasing the effective number of parameters to be
learned.

There are some works attempting to integrate CNN and VLAD or FV pooling in an end to end manner. In \cite{Arandjelovic2016netvlad}, a NetVLAD network was proposed by plugging a VLAD-like layer into a CNN network at the last convolutional layer and allows training end to end. The model was
initially designed for place recognition, however when applied to texture classification by Song \emph{et al.} \cite{Song2017Locally} it was found that the classification performance was inferior to FVCNN. Similar to NetVLAD \cite{Arandjelovic2016netvlad}, a Deep Texture Encoding Network (DeepTEN) was introduced in \cite{Zhang2017Deep} by integrating an encoding layer on top of convolutional layers, also generalizing orderless pooling methods such as VLAD and FV in a CNN trained end to end.

\subsection{\textbf{Using Handcrafted Deep Convolutional Networks}}
\label{subsec:TextureSpecific}

\begin {figure}[!t]
\centering
\includegraphics[width=0.48\textwidth]{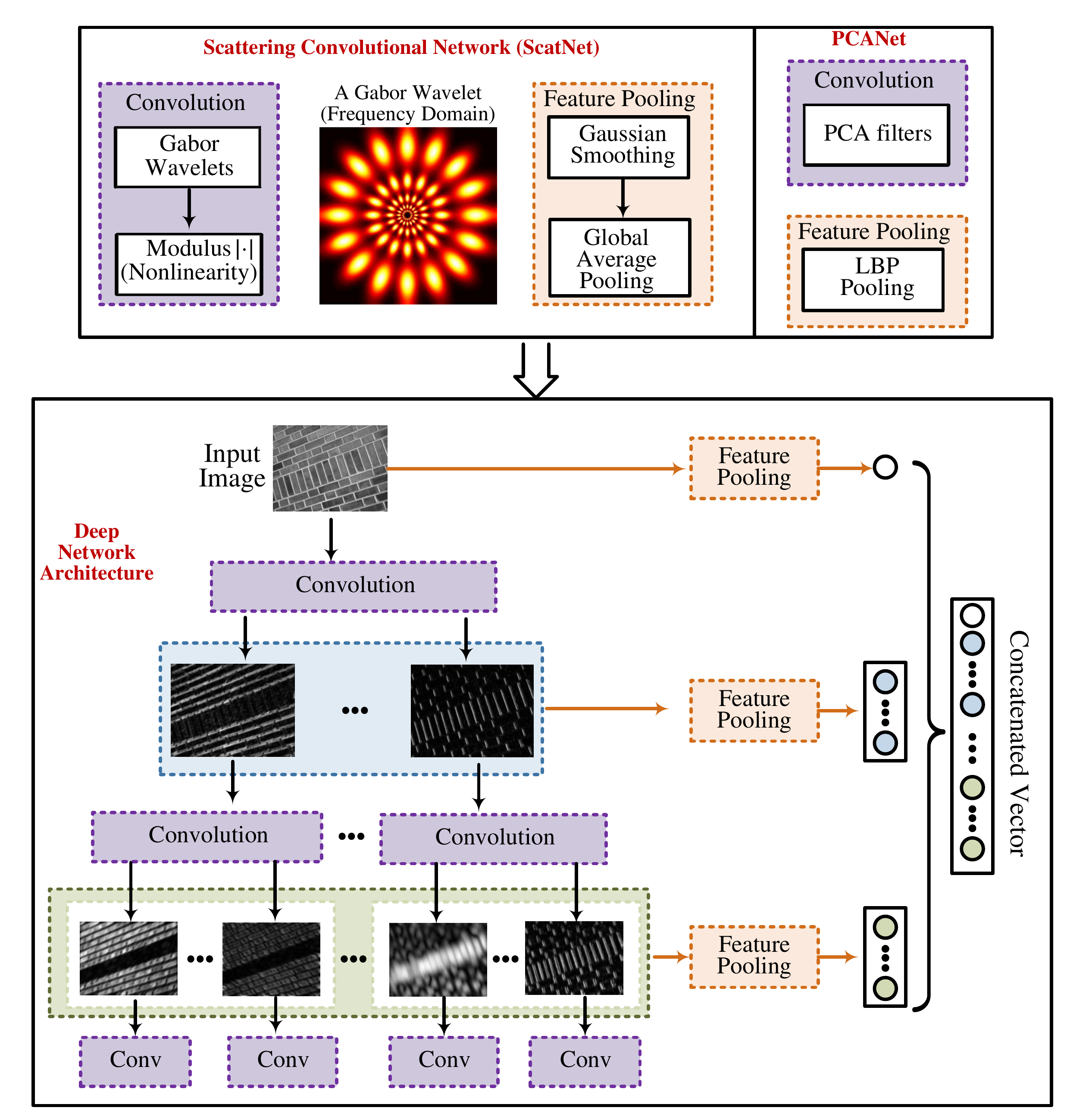}
\caption{Illustration of two similar handcrafted deep convolutional networks: ScatNet \cite{Bruna13} and PCANet \cite{Chan15}.}
\label{fig:ScatNet}
\end {figure}

In addition to the CNN based methods reviewed in Sections \ref{sec:Pretrained} and  \ref{subsec:Finetuned}, some ``handcrafted'' \footnote{Note that ``handcrafted'' commonly used for traditional features is somewhat imprecise, because many traditional features like Gabor filters are biologically or psychologically inspired.} deep convolutional networks \cite{Bruna13,Chan15} deserve attention. Recall that a standard CNN architecture (as shown in Fig.~\ref{fig:FrameWorkCompare} (b)) consists of multiple \emph{trainable} building blocks stacked on top of one another followed by a supervised classifier. Each block generally
consists of three layers:  a convolutional filter bank layer,
a nonlinear layer, and a feature pooling layer. Similar to the CNN architecture, Bruna and Mallat \cite{Bruna13} proposed a highly influential Scattering convolution Network (ScatNet), as illustrated in Fig.~\ref{fig:ScatNet}.

The key difference from CNN, where the convolutional filters are learned from data, is that the convolutional filters in ScatNet are predetermined --- they are simply wavelet filters, such as Gabor or Haar wavelets, and no learning is required. Moreover, the ScatNet usually cannot go as deep as a CNN; Bruna and Mallat \cite{Bruna13} suggested two convolutional layers, since the energy of the third layer scattering coefficients is negligible. Specifically, as can be seen in Fig.~\ref{fig:ScatNet}, ScatNet cascades wavelet transform convolutions with modulus nonlinearity and averaging poolers. It is shown in \cite{Bruna13} that ScatNet computes translation-invariant image representations which are stable to deformations and preserve high frequency information for recognition. As shown in Fig.~\ref{fig:ScatNet}, the average pooled feature vector from each stage is concatenated to form the global feature representation of an image, which is input into a simple PCA classifier for recognition, and which has demonstrated very high performance in texture recognition \cite{Bruna13,Sifre12,Sifre13,SifreRigid,liu2017local}. It achieved very high classification performance on Outex\_TC10 ($99.7\%$), Outex\_TC12 ($99.1\%$), KTHTIPS ($99.4\%$), CUReT ($99.8\%$), UIUC ($99.4\%$) and UMD ($99.7\%$) \cite{Bruna13,Sifre13,liu2017local}, but performed poorly on even challenging datasets like DTD ($35.7\%$). A downside of ScatNet is that the feature extraction stage is very time consuming, although the dimensionality of the global representation feature is relatively low (several hundreds). ScatNet has been extended to achieve rotation and scale invariance  \cite{Sifre12,Sifre13,SifreRigid} and applied to other problems besides texture such as object recognition \cite{Oyallon2015Deep}. Importantly, the mathematical analysis of ScatNet explains important properties of CNN architectures, and it is one of the few works that provides detailed theoretical understanding of CNNs.

Fig. \ref{fig:ScatNet} contrasts ScatNet and PCANet, proposed by Chan \emph{et al.} \cite{Chan15}, a very simple convolutional network based on trained PCA filters, instead of predefined Gabor wavelets, and LBP encoding \cite{Ojala02} and histogramming for feature pooling.  Two simple variations of PCANet, RandNet and LDANet, were also introduced in \cite{Chan15}, sharing the same topology as
PCANet, but their convolutional filters are either random filters as in \cite{LiuFieguthPAMI}
or learned from Linear Discriminant Analysis (LDA). Compared with ScatNet, feature extraction in PCANet is much faster, but with weaker invariance and texture classification performance \cite{liu2017local}.

\section{Attribute-Based Texture Representation }
\label{sec:Attributes}
\begin {figure}[!t]
\centering
\includegraphics[width=0.4\textwidth]{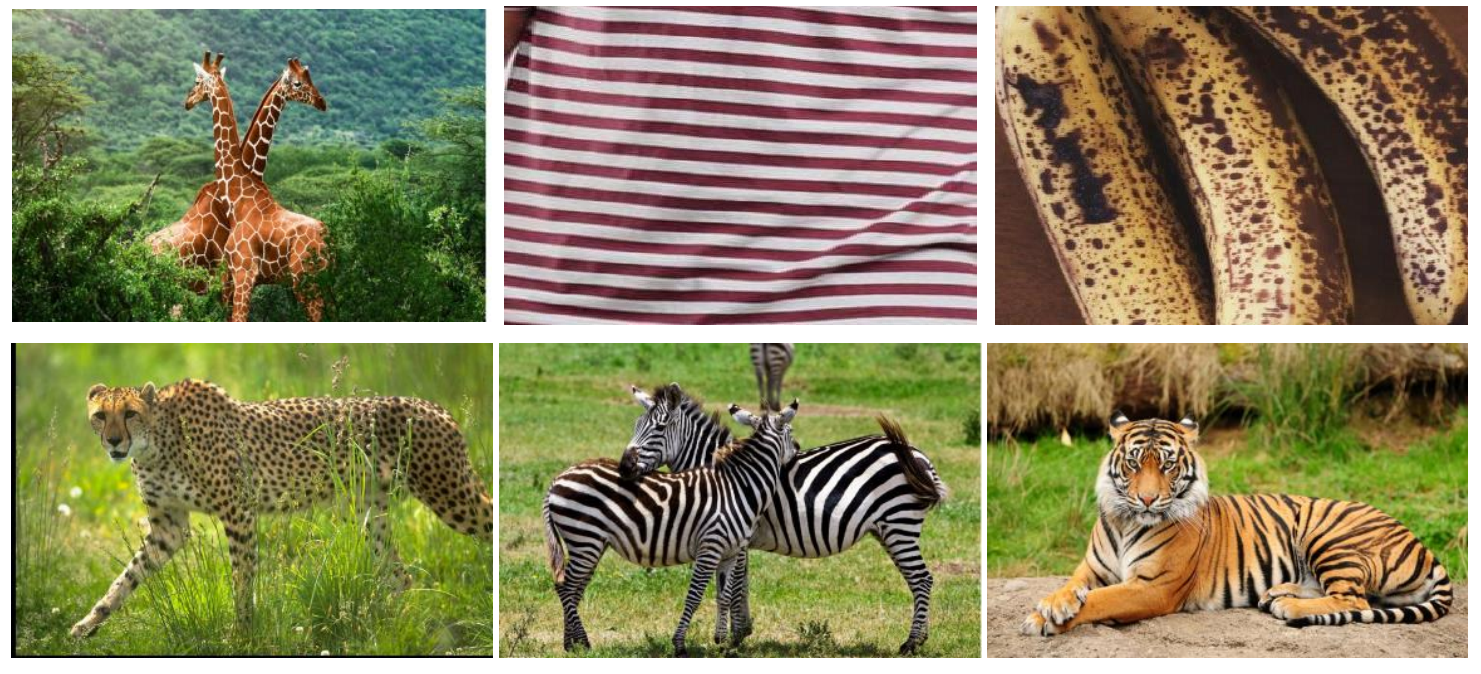}
\caption{Objects with rich textures in our daily life. Visual texture attributes like \emph{mesh}, \emph{spotted}, \emph{striated}, \emph{spotted} and \emph{striped} provide detailed and vivid descriptions of objects.}
\label{fig:Attributes}
\end {figure}

In recent years, the recognition of texture categories has been extensively studied and has shown substantial progress, partly thanks to the texture representations reviewed in Sections~\ref{sec:BoW} and~\ref{sec:CNN}.  Despite the rapid progress, particularly with the development of deep learning techniques, we remain far from reaching the goal of comprehensive scene
understanding \cite{krishna2016visual}. Although the traditional goal was to recognize
texture categories based on their perceptual differences or their material types, textures have other properties, as shown in Fig.~\ref{fig:Attributes}, where we may speak of a \emph{banded} shirt, a \emph{striped} zebra, and a \emph{striped} tiger. Here, \emph{banded} and \emph{striped} are referred to as visual texture attributes \cite{Cimpoi14}, which describe texture patterns using human-interpretable semantic words. With texture attributes, the textures shown back in Fig.~\ref{fig:TextureSamples} (d) might all be described as \emph{braided}, falling into a single category in the Describable
Textures Dataset (DTD) database \cite{Cimpoi14}.

The study of visual texture attributes \cite{Bormann2016Texture,Cimpoi14,matthews2013enriching} was motivated by the significant interest raised by visual attributes \cite{farhadi2009describing,patterson2014sun,parikh2011relative,kumar2011describable}. Visual attributes allow the describing of objects in significantly greater detail than a category label and are therefore important towards reaching the goal of comprehensive scene understanding \cite{krishna2016visual}, which would support important
applications such as detailed image search, question answering, and robotic interactions. Texture attributes are an important component of visual attributes, particularly for objects that are best characterized by a pattern. It can support advanced image search applications, such as more specific queries in image search engines (\emph{e.g.} a \emph{striped} skirt, rather than just any skirt). The investigation of texture attributes and detailed semantic texture description offers a significant opportunity to close the semantic gap in texture
modeling and to support applications that require fine grained texture description.
Nevertheless, there are only several papers \cite{Bormann2016Texture,Cimpoi14,matthews2013enriching} investigating the texture attributes thus far, and there is no systematic study yet attempted.

There are three essential issues in studying texture attribute based representation:
\begin{enumerate}
\item The identification of a universal texture attribute vocabulary that can describe a wide range of textures;
\item The establishment of a benchmark texture dataset, annotated by semantic attributes;
\item The reliable estimation of texture attributes from images, based on low level texture representations, such as the methods reviewed in Sections~\ref{sec:BoW} and~\ref{sec:CNN}.
\end{enumerate}
Tamura \emph{et al.} \cite{tamura1978textural} proposed a set of six attributes for describing textures: coarseness, contrast, directionality,
line-likeness, regularity and roughness. Amadasun and King \cite{Amadasun1989Textural} refined
this idea with the five attributes of coarseness, contrast, business,
complexity, and strength. Later, Bhushan \emph{et al.} \cite{bhushan1997texture} studied texture attributes from the perspective of psychology, asking subjects to cluster a collection of 98 texture adjectives according to similarity and identified eleven major clusters.

Recently, inspired by the work in \cite{bhushan1997texture,farhadi2009describing,parikh2011relative,kumar2011describable}, Matthews \emph{et al.} \cite{matthews2013enriching} attempted to enrich texture analysis with semantic attributes. They identified eleven commonly-used texture attributes\footnote{Blemished, bumpy, lined, marbled, random, repetitive, speckled,
spiralled, webbed, woven, and wrinkled} by selecting a single adjective from each of the eleven clusters identified
by Bhushan \emph{et al.} \cite{bhushan1997texture}. Then, with the eleven texture attributes, they released a publicly available human-provided labeling of over 300 classes of texture from the Outex database \cite{Ojala2002Outex}. For each texture image, instead of asking a subject to simply identifying the
presence or absence of each texture attribute, Matthews \emph{et al.} \cite{matthews2013enriching} proposed a framework of pairwise comparison, in which a subject was shown two texture images simultaneously and prompted
to choose the image exhibiting more of some attribute, motivated by the use of relative attributes \cite{parikh2011relative}.

After performing a screening process on the 98 adjectives identified by Bhushan \emph{et al.} \cite{bhushan1997texture}, Cimpoi \emph{et al.}~\cite{Cimpoi14} obtained a texture attribute vocabulary of 47 English adjectives and collected a dataset providing 120 example images for each attribute. They furthermore provide a comparison of BoW- and CNN-based texture representation methods for attribute estimation, demonstrating that texture
attributes are excellent texture descriptors, transferring between datasets.
Bormann \emph{et al.} \cite{Bormann2016Texture} introduced a set of seventeen human comprehensible attributes (seven color and ten structural) for color texture characterization. They also collected a new database named Robotics Domain Attributes Database (RDAD) for the indoor service robotics context. They compared five low level texture representation approaches for attribute prediction, and found that not all objects can be described very well with the seventeen attributes. Clearly, which attributes are best suited for a precise description of different object and texture classes deserves further attention.

\section{Texture Datasets and Performance}
\label{sec:Texturedatasets}
\subsection{\textbf{Texture Datasets}}
\label{sec:datasets}
Datasets have played an important role throughout the history of visual recognition research. They have been one of the most important factors for the considerable progress in the field, not only as a common ground for measuring and comparing performance of competing algorithms but also pushing the field towards increasingly complicated and challenging problems. With the rapid development of visual recognition approaches,  datasets have become progressively more challenging, evidenced by the fact that the recent large scale ImageNet dataset \cite{russakovsky2015imagenet} has enabled breakthroughs in visual recognition research. In the big data era, it becomes urgent to further enrich texture datasets to promote future research.  In this section, we discuss existing  texture image datasets that have been released and commonly used by the research community for texture classification, as summarized in Table~\ref{Tab:datasets}.

\begin{table*}[!t]
\caption {Summary of commonly-used texture databases.}\label{Tab:datasets}
\centering
\renewcommand{\arraystretch}{1.2}
\setlength\arrayrulewidth{0.2mm}
\setlength\tabcolsep{1pt}
\resizebox*{18cm}{!}{
\begin{tabular}{!{\vrule width1.2bp}c|c|c|c|c|c|c|c|c|c|c|c|c|c|c|c!{\vrule width1.2bp}}
\Xhline{1pt}
\scriptsize   \shortstack [c] {No.} & \scriptsize   \shortstack [c] {Texture\\Dataset} & \scriptsize   \shortstack [c] {References} & \scriptsize   \shortstack [c]
{Total \\ Images} & \scriptsize   \shortstack [c]
{Texture \\ Classes} & \scriptsize   \shortstack [c] {Image \\ Size }
& \scriptsize   \shortstack [c] {Gray or \\  Color}
& \scriptsize   \shortstack [c] {Imaging \\ Environment}
& \scriptsize   \shortstack [c] {Illumination \\ Changes}
& \scriptsize   \shortstack [c] {Rotation \\ Changes}
& \scriptsize   \shortstack [c] {Viewpoint \\ Changes}
 & \scriptsize   \shortstack [c]{Scale \\ Changes}
& \scriptsize   \shortstack [c]{Image \\ Content}
& \scriptsize   \shortstack [c]{Instances or \\ Categories}
& \scriptsize   \shortstack [c]{Year}
& \scriptsize   \shortstack [c]{Download\\Link} \\
\hline
\scriptsize $1$ &\scriptsize   Brodatz & \scriptsize \cite{Brodatz66} & \scriptsize  $ 111 $& \scriptsize  $111$
& \scriptsize $640\times640$& \scriptsize  Gray & \scriptsize  Controlled
& \scriptsize & \scriptsize  & \scriptsize  & \scriptsize
& \scriptsize  Objects& \scriptsize   Instances& \scriptsize $ 1966$
& \scriptsize  \cite{BrodatzDwon} \\
\hline
\scriptsize $2$ &\scriptsize   VisTex & \scriptsize $-$ & \scriptsize   $167 $& \scriptsize $ 167$
& \scriptsize $786\times512$& \scriptsize  Color
& \scriptsize  Wild
& \scriptsize & \scriptsize $\surd$ & \scriptsize  & \scriptsize
& \scriptsize  Objects& \scriptsize   Instances& \scriptsize $ 1995$
& \scriptsize  \cite{VisTexDown} \\
\hline
\scriptsize $3$ &\scriptsize   CUReT & \scriptsize \cite{dana1999reflectance}& \scriptsize   $5612$ & \scriptsize  $92$
& \scriptsize $200\times200$& \scriptsize  Color
& \scriptsize  Controlled
& \scriptsize $\surd$ & \scriptsize Small  & \scriptsize $\surd$ & \scriptsize
& \scriptsize  Materials& \scriptsize  Instances& \scriptsize  $1999$
& \scriptsize  \cite{CUReTDown} \\
\hline
\scriptsize $4$ &\scriptsize   Outex & \scriptsize \cite{Ojala2002Outex}& \scriptsize   $8640$ & \scriptsize  $320$
& \scriptsize $746\times 538$& \scriptsize  Color
& \scriptsize  Controlled
& \scriptsize $\surd$ & \scriptsize $\surd$  & \scriptsize  & \scriptsize
& \scriptsize  Materials & \scriptsize  Instances& \scriptsize  $2002$
& \scriptsize  \cite{OutexDown} \\
\hline
\scriptsize $5$ &\scriptsize   KTHTIPS & \scriptsize \cite{hayman2004significance,KTdoc} & \scriptsize   $810$ & \scriptsize  $10$
& \scriptsize $200\times 200$& \scriptsize  Color
& \scriptsize  Controlled
& \scriptsize $\surd$ & \scriptsize Small  & \scriptsize Small  & \scriptsize $\surd$
& \scriptsize  Materials & \scriptsize  Instances& \scriptsize  $2004$
& \scriptsize  \cite{KTHTIPSDown} \\
\hline
\scriptsize $6$ &\scriptsize   UIUC & \scriptsize  \cite{Lazebnik05}& \scriptsize   $1000$ & \scriptsize  $25$
& \scriptsize $640 \times 480$& \scriptsize  Gray
& \scriptsize  Wild
& \scriptsize $\surd$ & \scriptsize $\surd$  & \scriptsize $\surd$  & \scriptsize $\surd$
& \scriptsize  Materials & \scriptsize  Instances& \scriptsize  $2005$
& \scriptsize  \cite{UIUCDown} \\
\hline
\scriptsize $7$ &\scriptsize   KTHTIPS2a & \scriptsize \cite{caputo2005class,KT2doc} & \scriptsize    $4608$ & \scriptsize  $11$
& \scriptsize $200 \times 200$& \scriptsize  Color
& \scriptsize  Controlled
& \scriptsize $\surd$ & \scriptsize Small  & \scriptsize Small  & \scriptsize $\surd$
& \scriptsize  Materials & \scriptsize  Categories& \scriptsize  $2006$
& \scriptsize  \cite{KTHTIPSDown} \\
\hline
\scriptsize $8$ &\scriptsize   KTHTIPS2b  & \scriptsize \cite{caputo2005class,KT2doc}  & \scriptsize   $4752$ & \scriptsize  $11$
& \scriptsize $200 \times 200$& \scriptsize  Color
& \scriptsize  Controlled
& \scriptsize $\surd$ & \scriptsize Small  & \scriptsize Small  & \scriptsize $\surd$
& \scriptsize  Materials & \scriptsize  Categories& \scriptsize  $2006$
& \scriptsize  \cite{KTHTIPSDown} \\
\hline
\scriptsize $9$ &\scriptsize   UMD & \scriptsize \cite{xu2009viewpoint} & \scriptsize   $1000$ & \scriptsize  $25$
& \scriptsize $1280 \times 960$& \scriptsize  Gray
& \scriptsize  Wild
& \scriptsize $\surd$ & \scriptsize $\surd$  & \scriptsize $\surd$  & \scriptsize $\surd$
& \scriptsize  Objects & \scriptsize  Instances& \scriptsize  $2009$
& \scriptsize  \cite{UMDDown} \\
\hline
\scriptsize $10$ &\scriptsize   ALOT  & \scriptsize \cite{Burghouts09} & \scriptsize   $25000$ & \scriptsize  $250$
& \scriptsize $1536 \times 1024$& \scriptsize  Color
& \scriptsize  Controlled
& \scriptsize $\surd$ & \scriptsize   & \scriptsize   & \scriptsize
& \scriptsize  Materials & \scriptsize  Instances& \scriptsize  $2009$
& \scriptsize  \cite{ALOTDown} \\
\hline
\scriptsize  $11$ &\scriptsize   RawFooT & \scriptsize \cite{Cusano2016Evaluating}  & \scriptsize $3128$  & \scriptsize $68$
& \scriptsize  $800\times800$ & \scriptsize  Color
& \scriptsize Controlled
& \scriptsize  $\surd$  & \scriptsize    & \scriptsize  & \scriptsize
& \scriptsize  Materials  & \scriptsize  Instances  & \scriptsize  $2016$
& \scriptsize  \cite{RFTdown} \\
\hline
\scriptsize $12$ &\scriptsize   FMD & \scriptsize  \cite{sharan2009material,sharan2013recognizing}& \scriptsize   $1000$ & \scriptsize  $10$
& \scriptsize $512 \times 384$& \scriptsize  Color
& \scriptsize  Wild
& \scriptsize $\surd$ & \scriptsize $\surd$  & \scriptsize & \scriptsize
& \scriptsize  Materials & \scriptsize  Categories & \scriptsize  $2009$
& \scriptsize  \cite{FMDDown} \\
\hline
\scriptsize $13$ &\scriptsize   DreTex & \scriptsize \cite{oxholm2012scale} & \scriptsize   $40000$ & \scriptsize  $20$
& \scriptsize $200 \times 200$& \scriptsize  Color
& \scriptsize  Controlled
& \scriptsize $\surd$ & \scriptsize $\surd$  & \scriptsize  & \scriptsize $\surd$
& \scriptsize  Materials & \scriptsize  Instances& \scriptsize  $2012$
& \scriptsize  \cite{DrexelDown} \\
\hline
\scriptsize $14$ &\scriptsize   UBO2014 & \scriptsize \cite{weinmann2014material}  & \scriptsize $1915284$  & \scriptsize $7$
& \scriptsize $400\times400$  & \scriptsize  Color
& \scriptsize Synthesis
& \scriptsize  $\surd$  & \scriptsize  $\surd$   & \scriptsize $\surd$  & \scriptsize
& \scriptsize Materials   & \scriptsize  Categories  & \scriptsize  $2014$
& \scriptsize  \cite{UBOdown} \\
\hline
\scriptsize $15$ &\scriptsize   OpenSurfaces & \scriptsize \cite{bell2013opensurfaces} & \scriptsize   $10422$ & \scriptsize  $22$
& \scriptsize unfixed & \scriptsize  Color
& \scriptsize  Wild
& \scriptsize $\surd$ & \scriptsize $\surd$  & \scriptsize $\surd$ & \scriptsize $\surd$
& \scriptsize  Materials & \scriptsize  Clutter& \scriptsize  $2013$
& \scriptsize  \cite{OSDown} \\
\hline
\scriptsize $16$ &\scriptsize   DTD  & \scriptsize \cite{Cimpoi14} & \scriptsize   $5640$ & \scriptsize  $47$
& \scriptsize unfixed & \scriptsize  Color
& \scriptsize  Wild
& \scriptsize $\surd$ & \scriptsize $\surd$  & \scriptsize  & \scriptsize $\surd$
& \scriptsize  Attributes & \scriptsize  Categories& \scriptsize  $2014$
& \scriptsize  \cite{DTDDown} \\
\hline
\scriptsize $17$ &\scriptsize   MINC & \scriptsize  \cite{bell2015material} & \scriptsize   $2996674$ & \scriptsize  $23$
& \scriptsize unfixed & \scriptsize  Color
& \scriptsize  Wild
& \scriptsize $\surd$ & \scriptsize $\surd$  & \scriptsize $\surd$ & \scriptsize $\surd$
& \scriptsize  Materials & \scriptsize  Clutter& \scriptsize  $2015$
& \scriptsize  \cite{MINCDown} \\
\hline
\scriptsize $18$ &\scriptsize   MINC2500 & \scriptsize  \cite{bell2015material} & \scriptsize   $57500$ & \scriptsize  $23$
& \scriptsize $362 \times 362$& \scriptsize  Color
& \scriptsize  Wild
& \scriptsize $\surd$ & \scriptsize $\surd$  & \scriptsize $\surd$  & \scriptsize $\surd$
& \scriptsize  Materials & \scriptsize  Clutter & \scriptsize  $2015$
& \scriptsize  \cite{MINCDown} \\
\hline
\scriptsize $19$ &\scriptsize   GTOS & \scriptsize \cite{Xue2017differential}  & \scriptsize $34243$  & \scriptsize $40$
& \scriptsize  $240 \times 240$ & \scriptsize  Color
& \scriptsize \shortstack [c] {Partially \\ Controlled}
& \scriptsize  $\surd$ & \scriptsize  $\surd$  & \scriptsize $\surd$ & \scriptsize
& \scriptsize  Materials  & \scriptsize  Instances  & \scriptsize  $2016$
& \scriptsize  \cite{GTOSdown} \\
\hline
\scriptsize $20$ &\scriptsize   LFMD & \scriptsize \cite{Wang20164D}  & \scriptsize  $1200$ & \scriptsize $12$ & \scriptsize  $3787\times2632$ & \scriptsize  Color
& \scriptsize Uncontrolled
& \scriptsize  $\surd$ & \scriptsize    & \scriptsize & \scriptsize
& \scriptsize  Materials  & \scriptsize Categories   & \scriptsize  $2016$
& \scriptsize  \cite{LFdown} \\
\hline
\scriptsize $21$ &\scriptsize   RDAD & \scriptsize \cite{Bormann2016Texture}  & \scriptsize $1488$  & \scriptsize $57$
& \scriptsize $2592\times1944$  & \scriptsize  Color
& \scriptsize Uncontrolled & \scriptsize   & \scriptsize  $\surd$  & \scriptsize $\surd$  & \scriptsize
& \scriptsize  Objects   & \scriptsize  Instances  & \scriptsize  $2016$
& \scriptsize  \cite{RDADdown} \\
\Xhline{1pt}
\end{tabular}
}
\end{table*}

The Brodatz texture database \cite{BrodatzDwon}, derived from the Brodatz Album \cite{Brodatz66}, is the earliest, the most widely used and the most famous texture database. It has a relatively large number of
classes (111), with each class having only one image. Many texture representation approaches exploit the Brodatz database for evaluations \cite{Kim2002Support,LiuFieguthPAMI,Ojala02,Pun2003Log,
randen1999filtering,Valkealahti1998Reduced}, however in most  cases the entire database is not utilized, except in some recent studies \cite{Georgescu2003Mean,Lazebnik05,liu2017local,Picard1993Real,Zhang07}. The database has been criticized because of the lack of intraclass variations such as scale, rotation, perspective and illumination.

The Vision Texture Database (VisTex) \cite{Liu2005Promise,VisTexDown} is an early and well-known database. Built by the MIT Multimedia Lab, it has 167 classes of textures, each with only one image. The VisTex textures are imaged under natural lighting conditions, and have extra visual cues such as shadows, lighting, depth, perspective, thus closer in appearance to real-world images.  VisTex is often used for texture synthesis or segmentation, but rarely for image-level texture classification.

Since 2000, texture recognition has evolved to classifying real world
textures with large intraclass variations due to changes in camera pose and illumination, leading to the development of a number of benchmark texture datasets based on various real-world material instances. Among these, the most famous and widely used is the Columbia-Utrecht Reflectance and Texture (CUReT) dataset \cite{dana1999reflectance}, with 61 different material textures  taken under varying image conditions in a controlled lab environment. The effects of specularities, interreflections, shadowing, and
other surface normal variations are evident, as shown in Fig.~\ref{fig:TextureSamples} (a). CUReT is a considerable improvement over Brodatz, where all such effects are absent. Based on the original CUReT, Varma and Zisserman \cite{Varma05} built a subset for texture classification, which became the widely used benchmark to assess classification performance. CUReT has limitations of no significant scale change for most of the textures and limited in-plane rotation. Thus, a discriminative texture feature without rotation invariance can achieve high recognition rates \cite{Bruna13}.

\begin {figure}[!t]
\centering
\includegraphics[width=0.45\textwidth]{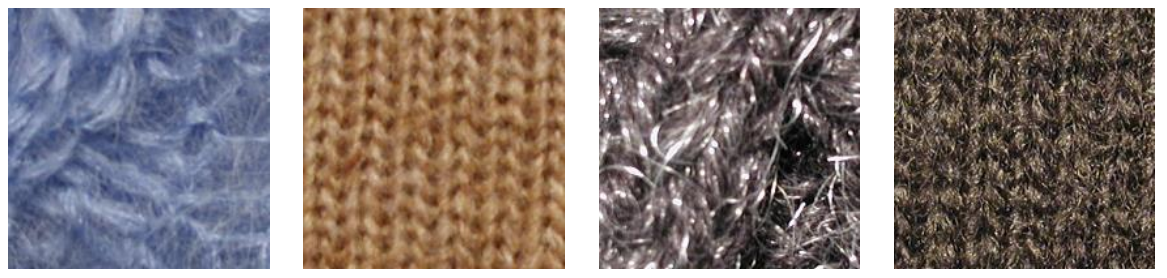}
\caption{Image examples from one category in KTHTIPS2.}
\label{fig:KTHTIPS2}
\end {figure}

Noticing the limited scale invariance in CUReT, researchers from the Royal Institute of Technology (KTH) introduced a dataset called ``KTH Textures under varying Illumination, Pose, and Scale'' (KTHTIPS) \cite{hayman2004significance,KTHTIPSDoc} by imaging ten CUReT materials at three different illuminations, three different poses, and nine different distances, but with significantly fewer settings for lighting and viewing angle than CUReT.  KTHTIPS was created to extend CUReT in two directions: (i) by providing variations in scale (as shown in Fig. \ref{fig:KTHTIPS2}), and (ii) by imaging different samples of the CUReT materials in different settings. This supports the study of recognizing different samples of the CUReT materials; for instance, does training on CUReT enable good recognition performance on KTHTIPS? Despite pose variations, KTHTIPS rotation variations are rather limited.

Experiments with Brodatz or VisTex used different nonoverlapping
subregions from the same image for training and testing; experiments with CUReT or KTHTIPS used different subsets of the images imaged from the identical sample for training and testing. KTHTIPS2 was one of the first datasets to offer considerable variations within each class. It groups textures not only by instance, but also by the type of material (\emph{e.g.,} wool). It is built on KTHTIPS and provides a considerable extension by imaging four physical, planar samples of each of eleven materials \cite{KTHTIPSDoc}.

The Oulu Texture (Outex) database was collected by the Machine Vision Group at the University of Oulu \cite{Ojala2002Outex}. It has the largest number of different texture classes (320), with each class having images photographed under three illuminations and nine rotation angles, but with limited scale variations. Based on Outex, a series of benchmark test suites were derived for evaluations of texture classification or segmentation algorithms \cite{Ojala2002Outex}. Among them, two benchmark datasets Outex\_TC00010 and Outex\_TC00012 \cite{Ojala02} designated for testing rotation and illumination invariance, appear commonly in papers.

The UIUC (University of Illinois Urbana-Champaign) dataset collected by Lazebnik \emph{et al.} \cite{Lazebnik05} contains 25 texture classes, with each class having 40 uncalibrated, unregistered images. It has significant variations in scale and viewpoint as well as nonrigid deformations (see Fig.~\ref{fig:TextureSamples} (b)), but has less severe illumination variations than CUReT. The challenges of this database are that there are few sample images per class, but with significant variations within classes.  Though UIUC improves over CUReT in terms of large intraclass variations, it is much smaller than CUReT both in the number of classes and the number of images per class. The UMD (University of Maryland) dataset \cite{xu2009viewpoint} also contains 25 texture classes; similar to UIUC, it has significant viewpoint and scale variations and uncontrolled illumination conditions. As textures are imaged under variable truncation, viewpoint, and illumination, the UIUC and the UMD have
stimulated the creation of texture representations that are invariant to significant viewpoint changes.

The Amsterdam Library of Textures (ALOT) database \cite{Burghouts09} consists of 250 texture classes. It was collected under controlled lab environment at eight different lighting conditions. Although it has a much larger number of texture classes than UIUC or UMD, it has little scale, rotation and viewpoint variations and is therefore not a very challenging dataset. The Drexel Texture (DreTex) dataset \cite{oxholm2012scale} contains 20 different textures, each of which was imaged approximately 2000 times under different (known) illumination directions, at multiple distances, and with different in-plane and out of plane rotations. It contains stochastic and regular textures.

The Raw Food Texture database (RawFooT), has been specially designed to
investigate the robustness of texture representation methods with respect
to variations in the lighting conditions \cite{Cusano2016Evaluating}. It consists of 68 texture classes of raw food,
with each class having 46 images acquired under 46 lighting conditions
which may differ in the light direction, in the illuminant color, in its intensity, or in a combination of these factors.
It has no variations in rotation, viewpoint and scale.

Due to the rapid progress of texture representation approaches, the performance of many methods on the datasets described above are close to saturation, with KTHTIPS2b being an exception due
to its increased complexity. However, most datasets introduced above make the simplifying
assumption that textures fill images, and often there is limited intraclass variability, due to a single or limited number of instances, captured under controlled scale, viewpoint and illumination.
In recent years, researchers have set their sights on
more complex recognition problems where textures appear under poor viewing conditions,
low resolution, and in realistic cluttered backgrounds.
The Flickr Material Database (FMD) \cite{sharan2009material,sharan2013recognizing} was built to address some of these limitations, by collecting many different object instances from the Internet grouped in 10 different material
categories, with examples shown in Fig.~\ref{fig:TextureSamples} (e).
The FMD \cite{sharan2009material} focuses on identifying materials such as plastic, wood, fiber and glass.
The limitations of the FMD dataset is that its size is quite small, containing only 10 material classes with 100 images in each class.

The UBO2014 dataset \cite{weinmann2014material} contains 7 material categories, with each having
12 different physical instances. Each material instance was measured by a full bidirectional texture function with 22,801 images
(a sampling of 151 viewing and 151 lighting directions), resulting in a total of more than 1.9 million
synthesized images. This synthesized material dataset allows classifying materials under complex real world scenarios.

\begin {figure}[!t]
\centering
\includegraphics[width=0.4\textwidth]{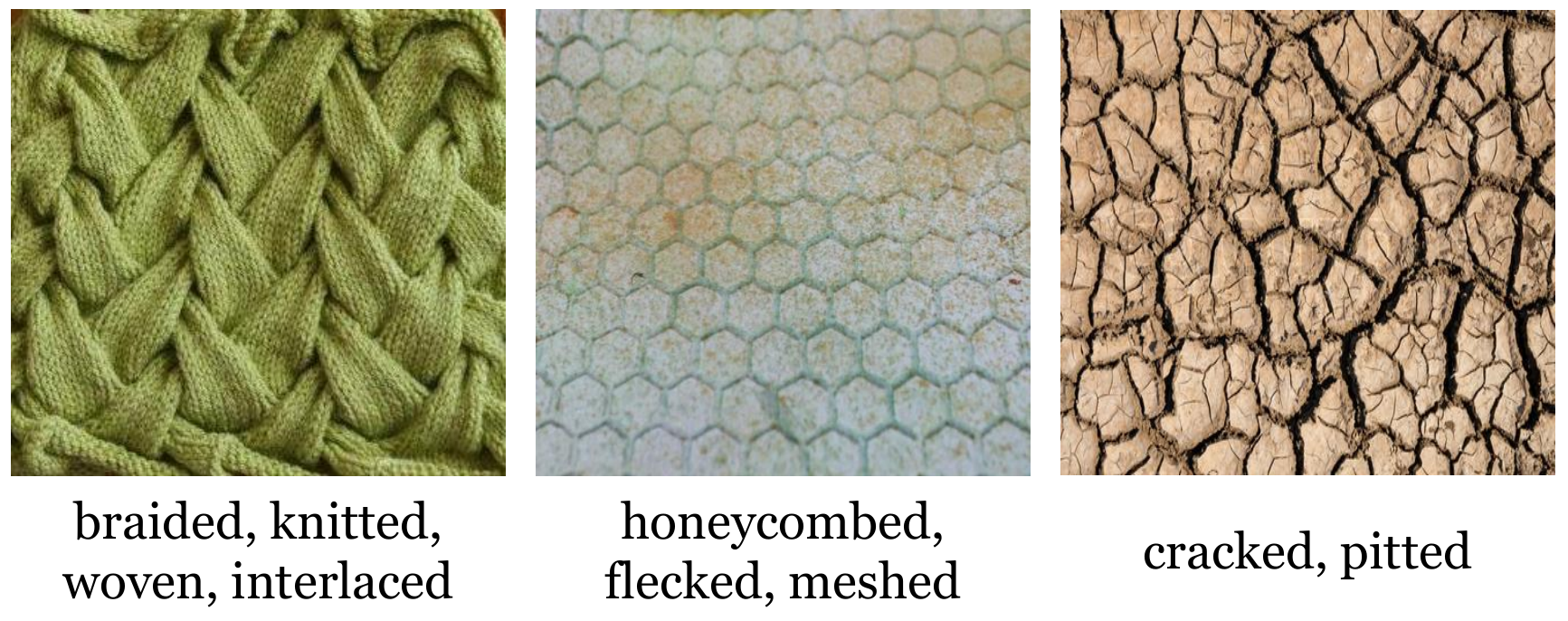}
\caption{Describing textures with attributes:  The goal of DTD is to understand and generate automatically human interpretable descriptions such as the examples above.}
\label{fig:DTDAttributes}
\end {figure}

Motivated by recent interests in visual attributes \cite{farhadi2009describing,patterson2014sun,parikh2011relative,kumar2011describable},
Cimpoi \emph{et al.} \cite{Cimpoi14} identified a vocabulary of $47$ texture attributes based on the seminal work of Bhusan \emph{et al.} \cite{bhushan1997texture} who studied the relationship between
commonly used English words and the perceptual properties of textures, identifying a set of words sufficient to describing a wide variety of texture patterns. These human interpretable texture attributes can vividly characterize textures,  as shown in Fig. \ref{fig:DTDAttributes}. Based on the 47 texture attributes,
they introduced a corresponding DTD dataset consisting of 120 texture images per attribute,
by downloading images from the Internet in an effort to
support directly real world applications.  The large intraclass variations in the DTD are different from traditional texture datasets like CUReT, UIUC and UMD, in the sense that the images shown in Fig.~\ref{fig:TextureSamples} (d) all belong to the {\em braided} class, whereas in a traditional sense
these textures should belong to rather different texture categories.

\begin {figure}[!t]
\centering
\includegraphics[width=0.4\textwidth]{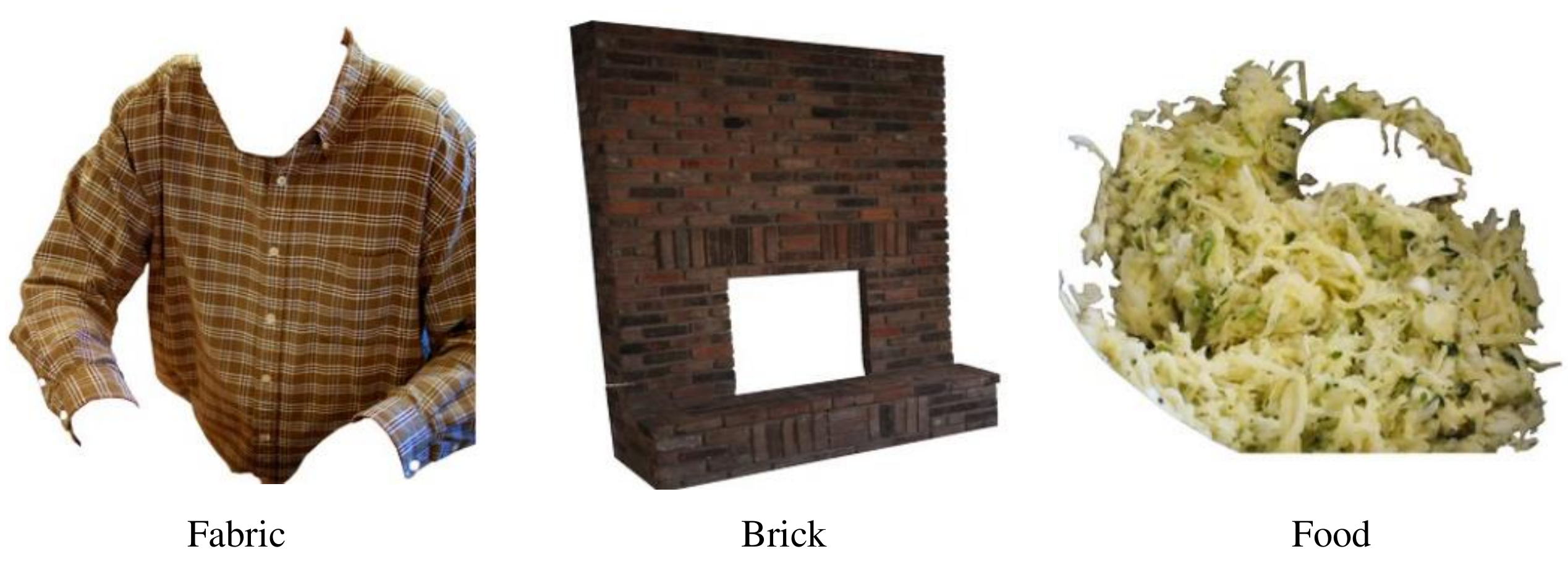}
\caption{Examples of material segments in the OpenSurfaces dataset.}
\label{fig:OS}
\end {figure}

Subsequent to FMD, Bell \emph{et al.} \cite{bell2013opensurfaces}
released OpenSurfaces (OS) which has over 20,000 images from consumer photographs, each containing a
number of high-quality texture or material segments. Images in OS have real world context,
in contrast to prior databases where each image belong to one texture category and the texture fills the whole image.
OS has over 100,000 segments (as shown shown in Fig.~\ref{fig:OS}) that can support a variety of applications.
Many, but not all, of these segments are annotated with material names,
the viewpoint, reflectance, the object names and scene class. The number of segments in each material category can also be highly unbalanced in the OS.

\begin {figure}[!t]
\centering
\includegraphics[width=0.45\textwidth]{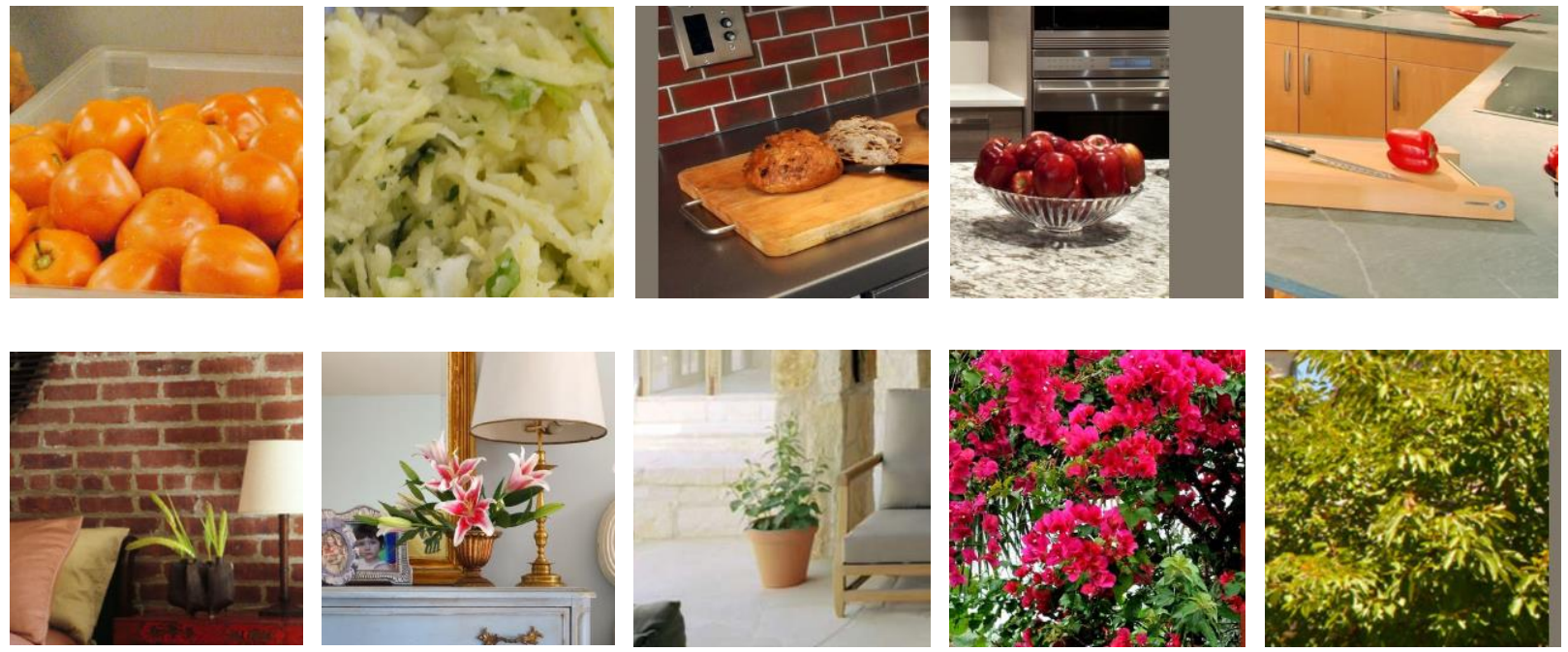}
\caption{Image samples from the MINC database. The first row are images from the {\em food} category, while the second row are images from {\em foliage}.}
\label{fig:MINCimages}
\end {figure}

Using the OS dataset as the seed, Bell \emph{et al.} \cite{bell2015material}
introduced a large material dataset named the Materials in Context Database (MINC) for material
recognition and segmentation in the wild, with samples shown in Fig.~\ref{fig:MINCimages}.
MINC has a total of 3 million material samples from 23
different material categories.
MINC is more
diverse, has more samples in each category, and is much larger than previous datasets.
Bell \emph{et al.} concluded that a large and well-sampled dataset such as MINC
is key for real-world material recognition and segmentation.

Concurrent to the work by
Bell \emph{et al.} \cite{bell2015material}, Cimpoi \emph{et al.} \cite{Cimpoi2016deep} derived a new dataset
from OS to conduct a study of material and describable texture attribute
recognition in clutter. Since not all segments in OS have a complete set of annotations,
Cimpoi \emph{et al.} \cite{Cimpoi2016deep} selected a subset of segments annotated with
material names, annotated the dataset with eleven texture attributes, and removed those material classes containing fewer than 400 segments.
Similarly, the Robotics Domain Attributes Database (RDAD) \cite{Bormann2016Texture} contains 57 categories
 of everyday indoor object and surface textures labeled with a set of seventeen
 texture attributes, collected to addresses the target domain of everyday objects and surfaces that a service robot
might encounter.

Wang \emph{et al.} \cite{Wang20164D} introduced a new light-field dataset of materials, called the Light-Field Material Database (LFMD).  Since light-fields can capture multiple viewpoints in a single shot, they implicitly contain reflectance information, which should be helpful in material recognition.  The goal of LFMD is to investigate whether 4D light-field information improves the performance of material recognition.

Finally, Xue \emph{et al.} \cite{Xue2017differential} built a material database named the
Ground Terrain in Outdoor Scenes (GTOS) to study the use of spatial and angular
reflectance information of outdoor ground terrain for
material recognition. It consists of over 30,000 images covering 40 classes of outdoor ground terrain
under varying weather and lighting conditions.

\begin {figure*}[!t]
\centering
\includegraphics[width=0.8\textwidth]{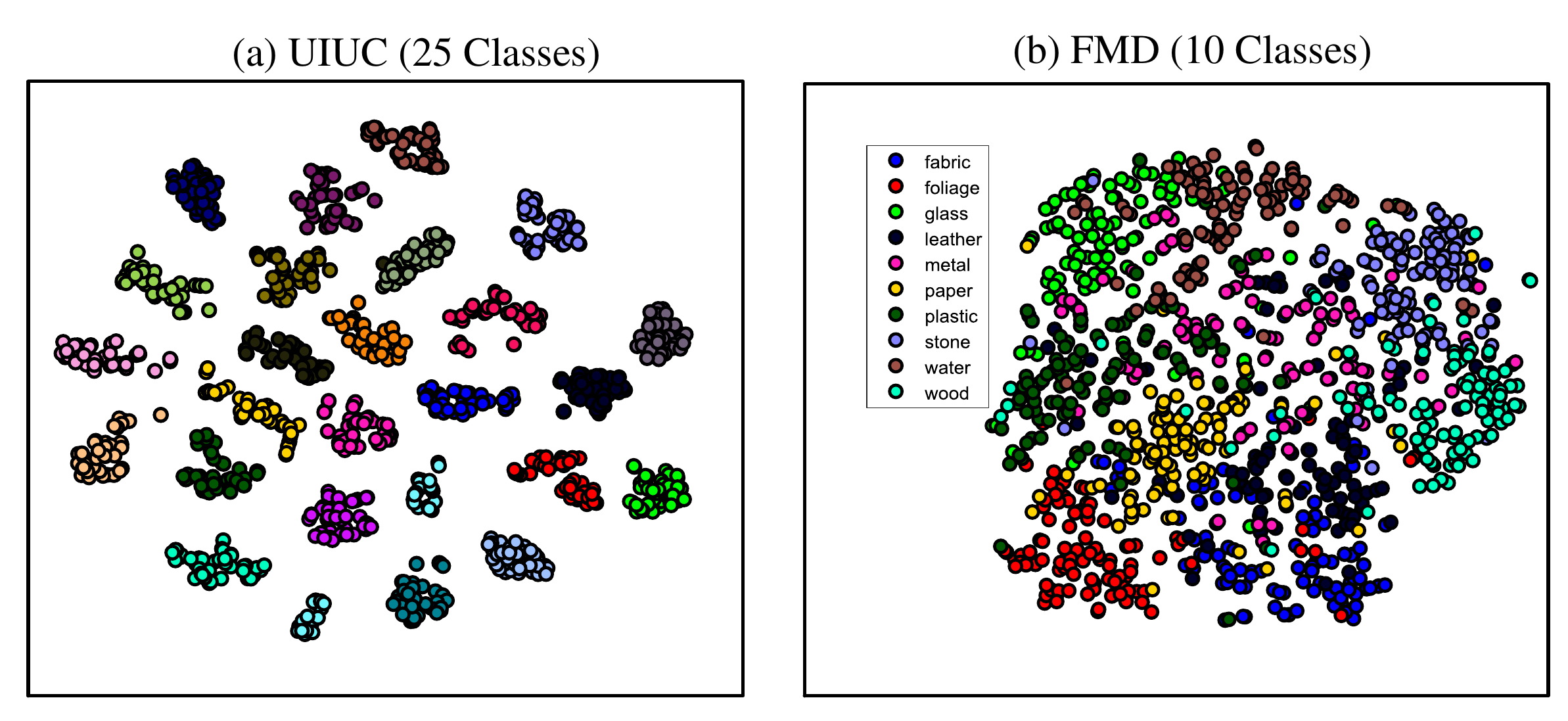}
\caption{t-distributed Stochastic Neighbor Embedding (tSNE) \cite{Maaten2008Visualizing} of textures from the IFV encoding of the VGGVD features
\cite{Cimpoi2016deep} from (a) the UIUC dataset and (b) the FMD dataset.
Clearly the classes in UIUC are more separable than those in FMD.}
\label{fig:gscatter}
\end {figure*}

\subsection{\textbf{Performance}}
\label{sec:Performance}

Table \ref{Tab:Results} presents a performance summary of representative methods applied to
popular benchmark texture datasets.  It is clear that major improvements have come from more powerful local texture descriptors
such as MRELBP \cite{liu2017local,LiuTIP16}, ScatNet \cite{Bruna13} and
CNN-based descriptors \cite{Cimpoi2016deep} and from advanced feature encoding methods like IFV \cite{Perronnin10}.
With the advance in CNN architectures, CNN-based texture representations have quickly demonstrated their strengths in texture classification, especially for recognizing textures with very large appearance
variations, such as in KTHTIPS2b, FMD and DTD.

Off-the-shelf CNN based descriptors, in combination with IFV feature encoding,
have advantages in nearly all of the benchmark datasets, except for Outex\_TC10 and Outex\_TC12, where
texture descriptors, such as MRELBP \cite{liu2017local,LiuTIP16} and ScatNet, \cite{Bruna13} that have rotation and gray scale invariances, give perfect accuracies, revealing one of the limitations of
CNN based descriptors in being sensitive to image degradations. Despite the usual advantages of CNN based methods, it is at a cost of very high computational complexity and memory requirements.
We believe that traditional texture descriptors, like the efficient LBP and
robust variants such as MRELBP, still
have merits in cases when real-time computation is a priority or when
robustness to image degradation is needed \cite{liu2017local}.

As can be seen from Table \ref{Tab:Results}, currently the highest classification scores on Outex\_TC10,
Outex\_TC12, CUReT, KTHTIPS, UIUC, UMD and ALOT are nearly perfect, in excess of 99.5\%, and quite a few texture representation approaches can achieve more than $99.0\%$ accuracy on these datasets.
Since the influential work by Cimpoi \emph{et al.} \cite{Cimpoi14,Cimpoi15,Cimpoi2016deep}, who reported near perfect classification accuracies with pretrained CNN features for texture classification, subsequent representative
CNN based approaches have not reported results on these datasets because performance is saturated and because the
datasets are not large enough
to allow finetuning to obtain improved results.
The FMD, DTD and KTHTIPS2b are undoubtedly more challenging than
other texture datasets, for example the UIUC and FMD texture category separation shown in  Fig.~\ref{fig:gscatter}, and these more challenging datasets appear more frequently in recent works.
However, since the IFV encoding of
VGGVD descriptors \cite{Cimpoi2016deep}, the progress on these three datasets has been slow,
with incremental improvements in accuracy and efficiency obtained
by building more complex or deeper CNN architectures.

As can be observed from Table \ref{Tab:Results}, LBP type methods (LBP \cite{Ojala02},
MRELBP \cite{LiuTIP16} and BIF \cite{Crosier10}) which
adopt a predefined codebook have a much more efficient feature extraction step than the remaining methods listed. For those BoW based methods which require codebook learning, since the codebook learning, feature encoding, and pooling process are similar, the distinguishing factors are the computation and feature dimensionality of the local texture descriptor. Among commonly-used local texture
descriptors, those approaches first detecting local regions of interest followed by local descriptors, such as
SIFT, RIFT and SPIN \cite{Lazebnik05,Zhang07}, are among the slowest and have relatively high dimensionality.  For the CNN based
methods developed in \cite{Cimpoi14,Cimpoi15,Cimpoi2016deep}, CNN feature extraction is performed
on multiple scaled versions of the original texture image, which requires more computational time.
In general, CNN pretraining and finetuning is efficient, whereas CNN model training is time consuming.
From \cite{liu2017local}, ScatNet is computationally expensive at the feature extraction stage, though
it has medium feature dimensionality.
Finally, at the feature classification stage linear SVM is significantly faster than kernel SVM.

\begin{table*}[!t]
\begin{sideways}
\begin{minipage}{\textheight}
\caption{Performance ($\%$) summarization of some representative methods on popular benchmark texture datasets. All methods used the same splitting strategy for training and testing on each dataset. Specifically, for KTHTIPS2, one image per class is used for training and the remaining three for testing. For Brodatz, please see \cite{Lazebnik05}; for Outex\_TC10 and Outex\_TC12 please see \cite{liliuECCV16}. For DTD, 80 images per class are randomly selected for training and the remaining 40 for testing. For all other  datasets, half of the samples per class are chosen for training and the remaining half for testing. Results are averaged over a number of random partitionings of training and testing data.  All listed results are quoted from the original papers, except those marked with ($\star$) from \cite{Zhang07}, and those marked ($\diamond$) from \cite{liliuECCV16}. For interested readers, more results on LBP variants can be found in the recent survey \cite{liliuECCV16,liu2017local}. Those dimensions with $(\dagger)$ denote feature dimension before the SoftMax layer. For Brodatz, KTHTIPS2, FMD and DTD, the highest classification score is highlighted; for all other datasets classification scores higher than $99\%$ are highlighted.}\label{Tab:Results}
\centering
\renewcommand{\arraystretch}{1.3}
\setlength\arrayrulewidth{0.2mm}
\setlength\tabcolsep{2pt}
\resizebox*{!}{16cm}{
\begin{tabular}{!{\vrule width1.5bp}c|c|c|c!{\vrule width1.5bp}c|c|c|c|c!{\vrule width1.5bp}c|c|c|c|c|c|c|c|c|c|c!{\vrule width1.5bp}}
\Xhline{1.5pt}
\multicolumn{4}{!{\vrule width1.5bp}c!{\vrule width1.5bp}}{\footnotesize \textbf{Method Info} } & \multicolumn{5}{c!{\vrule width1.5bp}}{\footnotesize \textbf{Texture Representation and Classification} }
& \multicolumn{11}{c!{\vrule width1.5bp}}{\footnotesize \textbf{Results Reported on Popular Benchmark Texture Datasets} }\\
\hline
\multicolumn{2}{!{\vrule width1.5bp}c|}{\footnotesize \textbf{Method} } & \scriptsize  \shortstack [c] {Published in} & \scriptsize Described in & \scriptsize  \shortstack [c] {Local \\ Representation}& \scriptsize  \shortstack [c] {Codebook \\ Generation} & \scriptsize  \shortstack [c] {Feature \\ Encoding} & \scriptsize  \shortstack [c] {Feature \\ Classification} & \scriptsize  \shortstack [c] {Feature \\ Dimension} & \scriptsize  Outex\_TC10 & \scriptsize  Outex\_TC12
& \scriptsize  Brodatz & \scriptsize  CUReT & \scriptsize  KTHTIPS & \scriptsize  UIUC
& \scriptsize UMD  & \scriptsize  ALOT & \scriptsize   KTHTIPS2b
& \scriptsize  FMD & \scriptsize  DTD\\
\hline
\multirow{18}{*}{\rotatebox{90}{\scriptsize \textbf{Traditional BoW based Texture Representation}}} & \scriptsize   LBP \cite{Ojala02} & \scriptsize  TAPMI 2002 & \scriptsize Section \ref{subsubsec:Dense}& \scriptsize$\textrm{LBP}^{riu2}$&\scriptsize Predefined&\scriptsize BoW &\scriptsize Chi Square, NNC  &\scriptsize $210$ & \scriptsize \cellcolor{black!95}\color{black}{\textcolor{white}{ $99.7 (\diamond)$}}
& \scriptsize $92.1 (\diamond)$& \scriptsize $90.7 (\diamond)$  & \scriptsize $97.0 (\diamond)$
& \scriptsize $-$ & \scriptsize $88.4 (\diamond)$ & \scriptsize $96.2 (\diamond)$ & \scriptsize $94.2 (\diamond)$
& \scriptsize $-$ & \scriptsize  $-$ & \scriptsize $37.1 (\diamond)$ \\
\cline{2-20}
&\scriptsize   MR8 \cite{Varma05} & \scriptsize  IJCV 2005& \scriptsize Section \ref{subsubsec:Dense}& \scriptsize MR8 filters&\scriptsize \emph{k}means&\scriptsize BoW &\scriptsize  Chi Square, NNC  &\scriptsize $2440$ & \scriptsize $-$
& \scriptsize $-$ & \scriptsize  $-$ & \scriptsize $97.4$
& \scriptsize$-$ & \scriptsize $-$ & \scriptsize$-$  & \scriptsize$-$
& \scriptsize$-$  & \scriptsize   $-$& \scriptsize $-$  \\
\cline{2-20}
&\scriptsize   MR8 \cite{hayman2004significance} & \scriptsize  ECCV 2004& \scriptsize Section \ref{subsubsec:Dense}& \scriptsize MR8 filters&\scriptsize \emph{k}means&\scriptsize BoW &\scriptsize  Chi Square, SVM  &\scriptsize $2440$  & \scriptsize$-$
& \scriptsize $-$ & \scriptsize   & \scriptsize $98.5$
& \scriptsize$-$ & \scriptsize $-$ & \scriptsize $-$ & \scriptsize$-$
& \scriptsize $-$ & \scriptsize $-$  & \scriptsize $-$ \\
\cline{2-20}
&\scriptsize   Lazebnik  \emph{et al.} \cite{Lazebnik05} & \scriptsize  TPAMI 2005& \scriptsize Section \ref{subsubsec:Sparse}& \scriptsize \shortstack [c] {Corners, Blobs \\ SPIN, RIPF}&\scriptsize \emph{k}means &\scriptsize Signatures& \scriptsize EMD, NNC& \scriptsize \shortstack [c] {$40$ \\ Clusters}
& \scriptsize & \scriptsize$-$ & \scriptsize $88.2$
& \scriptsize $72.5(\star)$  & \scriptsize  $91.3(\star)$ & \scriptsize $96.0$ & \scriptsize$-$  & \scriptsize$-$
& \scriptsize $-$ & \scriptsize $-$  & \scriptsize $-$ \\
\cline{2-20}
&\scriptsize   Zhang \emph{et al.} \cite{Zhang07} & \scriptsize  IJCV 2007& \scriptsize Section \ref{subsubsec:Sparse}& \scriptsize \shortstack [c] {Corners, Blobs \\ SPIN, RIPF, SIFT}&\scriptsize \emph{k}means &\scriptsize Signatures& \scriptsize EMD, SVM& \scriptsize \shortstack [c] {$40$ \\ Clusters}
& \scriptsize $-$& \scriptsize$-$ & \scriptsize $95.4$
& \scriptsize $95.3$  & \scriptsize  $95.5$ & \scriptsize $98.7$ & \scriptsize  $-$& \scriptsize$-$
& \scriptsize $-$ & \scriptsize $-$  & \scriptsize $-$ \\
\cline{2-20}
&\scriptsize   MFS  \cite{xu2009viewpoint} & \scriptsize  IJCV 2009& \scriptsize Section \ref{subsubsec:Dense} &\scriptsize \shortstack [c] {Gaussian, Energy\\Laplacian}& \multicolumn{2}{|c|}{\scriptsize MFS Pooling }
& \scriptsize $l_1$, NNC & \scriptsize $78$ &\scriptsize $-$&\scriptsize $-$  & \scriptsize $-$ & \scriptsize $-$ & \scriptsize $-$ & \scriptsize $92.3$ & \scriptsize  $93.9$ & \scriptsize $-$
& \scriptsize $-$ & \scriptsize $-$  & \scriptsize $-$ \\
\cline{2-20}
&\scriptsize   OTF  \cite{xu2009combining} & \scriptsize  CVPR 2009& \scriptsize Section \ref{subsubsec:Dense}& \scriptsize \shortstack [c] {Multilevel\\Orientation Hist.}& \multicolumn{2}{|c|}{\scriptsize \shortstack [c] {Multiscale Decomposition \\ of MFS Vectors }}&\scriptsize RBF SVM&\scriptsize $1160$
& \scriptsize $-$& \scriptsize $-$& \scriptsize$-$& \scriptsize $-$ & \scriptsize$-$
& \scriptsize $97.4$ & \scriptsize  $98.5$  & \scriptsize$-$
& \scriptsize $-$ & \scriptsize $-$  & \scriptsize$-$  \\
\cline{2-20}
&\scriptsize   WMFS  \cite{Xu10} & \scriptsize  CVPR 2010& \scriptsize Section \ref{subsubsec:Dense}& \scriptsize \shortstack [c] {Multiorientation \\ Wavelet Pyramid}& \multicolumn{2}{|c|}{\scriptsize \shortstack [c] {MFS Pooling }}&\scriptsize RBF SVM&\scriptsize $103$
& \scriptsize $-$ & \scriptsize $-$ & \scriptsize $-$ & \scriptsize $-$ & \scriptsize $-$
& \scriptsize $98.6$ & \scriptsize  $98.7$ & \scriptsize $-$
& \scriptsize $-$ & \scriptsize  $-$ & \scriptsize $-$ \\
\cline{2-20}
&\scriptsize   Patch \cite{Varma09} & \scriptsize  TPAMI 2009& \scriptsize Section \ref{subsubsec:Dense} & \scriptsize Patch Vectors& \scriptsize \emph{k}means &\scriptsize BoW & \scriptsize Chi Square, NNC &\scriptsize \shortstack [c] {Dependent on \\ Dataset} & \scriptsize$-$  & \scriptsize$-$
& \scriptsize $92.9(\star)$ & \scriptsize $98.0$& \scriptsize $92.4(\star)$
& \scriptsize $97.8$ & \scriptsize  $-$& \scriptsize$-$
& \scriptsize $-$ & \scriptsize  $-$ & \scriptsize$-$  \\
\cline{2-20}
&\scriptsize   BIF \cite{Crosier10}& \scriptsize  IJCV 2010& \scriptsize Section \ref{subsubsec:Dense}& \scriptsize BIF Features&\scriptsize Predefined &\scriptsize BoW &\scriptsize Chi Square, NNC&\scriptsize $1296$
& \scriptsize  $-$ & \scriptsize $-$& \scriptsize$-$
& \scriptsize $98.6$  & \scriptsize $98.5$ & \scriptsize $98.8$ & \scriptsize$-$  & \scriptsize$-$
& \scriptsize  $-$& \scriptsize $-$  & \scriptsize$-$  \\
\cline{2-20}
&\scriptsize   RP \cite{LiuFieguthPAMI} & \scriptsize  TPAMI 2012& \scriptsize Section \ref{subsubsec:Dense}& \scriptsize Random Features& \scriptsize \emph{k}means &\scriptsize BoW & \scriptsize Chi Square, NNC&  \scriptsize \shortstack [c] {Dependent on \\ Dataset}
& \scriptsize  $-$ & \scriptsize$-$ & \scriptsize$-$
& \scriptsize $98.5$  & \scriptsize $-$  & \scriptsize $-$ & \scriptsize $-$ & \scriptsize$-$
& \scriptsize $-$ & \scriptsize  $-$ & \scriptsize$-$  \\
\cline{2-20}
&\scriptsize   SRP \cite{liu2011sorted} & \scriptsize  ICCV 2011& \scriptsize Section \ref{subsubsec:Dense}& \scriptsize SRP Features& \scriptsize \emph{k}means &\scriptsize BoW & \scriptsize Chi Square, SVM&\scriptsize \shortstack [c] {Dependent on \\ Dataset}
& \scriptsize  $-$ & \scriptsize$-$ & \scriptsize  $97.2$
& \scriptsize \cellcolor{black!95}\color{black}{\textcolor{white}{  $99.4$}}  & \scriptsize \cellcolor{black!95}\color{black}{\textcolor{white}{  $99.3$}}   & \scriptsize $98.6$ & \scriptsize \cellcolor{black!95}\color{black}{\textcolor{white}{  $99.3$}} & \scriptsize$-$
& \scriptsize $-$ & \scriptsize $-$  & \scriptsize $-$ \\
\cline{2-20}
&\scriptsize   Timofte \emph{et al.} \cite{timofte2012training} & \scriptsize BMVC 2012& \scriptsize Section \ref{subsubsec:Dense}& \scriptsize BIF Features&\scriptsize Predefined&\scriptsize  \shortstack [c] {MultiLevel \\ BoW} &\scriptsize \shortstack [c] {Reasoning \\ Collaborative} &\scriptsize $1780$
& \scriptsize $-$  & \scriptsize $-$& \scriptsize $97.3$
& \scriptsize $99.4$ & \scriptsize \cellcolor{black!95}\color{black}{\textcolor{white}{  $99.4$}}   & \scriptsize \cellcolor{black!95}\color{black}{\textcolor{white}{  $99.0$}} & \scriptsize \cellcolor{black!95}\color{black}{\textcolor{white}{  $99.5$}} & \scriptsize$-$
& \scriptsize $-$ & \scriptsize $55.8$  & \scriptsize $-$ \\
\cline{2-20}
&\scriptsize   Sharan \emph{et al.} \cite{sharan2013recognizing} & \scriptsize IJCV 2013& \scriptsize Section \ref{subsubsec:Dense}& \scriptsize \shortstack [c] {Eight Features \\ (Including SIFT) } & \scriptsize \emph{k}means &\scriptsize BoW &\scriptsize \shortstack [c] {Hist. \\ Intersection SVM }  &\scriptsize $1550$
& \scriptsize   $-$  & \scriptsize  $-$  & \scriptsize  $-$
& \scriptsize  $-$   & \scriptsize  $-$   & \scriptsize  $-$   & \scriptsize  $-$   & \scriptsize  $-$
& \scriptsize  $-$  & \scriptsize $57.1$  & \scriptsize  $-$  \\
\cline{2-20}
&\scriptsize   MRELBP  \cite{LiuTIP16} & \scriptsize TIP 2016& \scriptsize Section \ref{subsubsec:Dense}& \scriptsize MRELBP$^{riu2}$ &\scriptsize  Predefined &\scriptsize BoW &\scriptsize  Chi Square, SVM &\scriptsize  $800$
& \scriptsize  \cellcolor{black!95}\color{black}{\textcolor{white}{   $100 (\diamond)$}}  & \scriptsize  \cellcolor{black!95}\color{black}{\textcolor{white}{ $99.8 (\diamond)$}}   & \scriptsize $93.1(\diamond)$
& \scriptsize  \cellcolor{black!95}\color{black}{\textcolor{white}{  $99.0(\diamond)$}} & \scriptsize $-$ & \scriptsize $96.9(\diamond)$  & \scriptsize  \cellcolor{black!95}\color{black}{\textcolor{white}{ $99.4(\diamond)$}}
& \scriptsize  \cellcolor{black!95}\color{black}{\textcolor{white}{  $99.1(\diamond)$ }} & \scriptsize & \scriptsize $-$  & \scriptsize $44.9(\diamond)$ \\
\cline{2-20}
&\scriptsize   SIFT  \cite{lowe2004distinctive,Cimpoi2016deep} & \scriptsize IJCV 2016 & \scriptsize Section \ref{subsubsec:Dense}& \scriptsize Dense SIFT & \scriptsize GMM&\scriptsize IFV&\scriptsize  Linear SVM &\scriptsize $65536$ &\scriptsize $-$
& \scriptsize  $-$ & \scriptsize$-$   & \scriptsize  \cellcolor{black!95}\color{black}{\textcolor{white}{  $99.0$}}
& \scriptsize  \cellcolor{black!95}\color{black}{\textcolor{white}{  $99.5$}}    & \scriptsize $96.6$ & \scriptsize \cellcolor{black!95}\color{black}{\textcolor{white}{  $99.1$}} & \scriptsize$-$ & \scriptsize $70.8$
& \scriptsize $59.8$ & \scriptsize   $58.6$ \\
\Xhline{1.5pt}
\multirow{14}{*}{\rotatebox{90}{\scriptsize \textbf{CNN based Texture Representation}}} &\scriptsize   ScatNet  \cite{Bruna13,Sifre13} & \scriptsize \shortstack [c] {TPAMI 2013 \\ CVPR 2013} & \scriptsize   Section \ref{subsec:TextureSpecific} & \scriptsize \shortstack [c] {Gabor Wavelet \\ ScatNet} & \multicolumn{2}{|c|}{\scriptsize \shortstack [c] {Gaussian Smoothing \\ Average Pooling}  }&\scriptsize PCA Classifier &\scriptsize $596$
& \scriptsize  \cellcolor{black!95}\color{black}{\textcolor{white}{ $99.7 (\diamond)$ }}   & \scriptsize  \cellcolor{black!95}\color{black}{\textcolor{white}{ $99.1(\diamond)$ }}   & \scriptsize $84.5(\diamond)$&\scriptsize  \cellcolor{black!95}\color{black}{\textcolor{white}{  $99.8$}}
& \scriptsize  \cellcolor{black!95}\color{black}{\textcolor{white}{  $99.4$}} & \scriptsize  \cellcolor{black!95}\color{black}{\textcolor{white}{  $99.4$}} & \scriptsize  \cellcolor{black!95}\color{black}{\textcolor{white}{  $99.7$}}   & \scriptsize $98.3(\diamond)$ & \scriptsize $-$
& \scriptsize $-$ & \scriptsize  $35.7(\diamond)$ \\
\cline{2-20}
&\scriptsize   PCANet  \cite{Chan15} & \scriptsize TIP 2015& \scriptsize Section \ref{subsec:TextureSpecific}& \scriptsize \shortstack [c] {PCANet\\Stage 2}&\multicolumn{2}{|c|}{\scriptsize MultiBlock LBP Pooling } &\scriptsize Linear SVM &\scriptsize $32768$
& \scriptsize $-$  & \scriptsize $-$  & \scriptsize$-$
& \scriptsize \cellcolor{black!95}\color{black}{\textcolor{white}{  $99.6$}}  & \scriptsize $-$& \scriptsize  $-$ & \scriptsize$-$ & \scriptsize$-$
& \scriptsize $-$ & \scriptsize $-$  & \scriptsize $-$ \\
\cline{2-20}
&\scriptsize   AlexNet  \cite{Cimpoi2016deep} & \scriptsize IJCV 2016& \scriptsize Section \ref{sec:Pretrained}& \multirow{3}{*}{\scriptsize \shortstack [c] {Pretrained\\CONV features \\ (MultiScale \\ Inputs)}}  &\scriptsize GMM &\scriptsize IFV &\scriptsize Linear SVM &\scriptsize $32768$
& \scriptsize  $67.3(\diamond)$ & \scriptsize $72.3(\diamond)$& \scriptsize $98.2(\diamond)$
& \scriptsize $98.5$ & \scriptsize  \cellcolor{black!95}\color{black}{\textcolor{white}{  $99.6$}}  & \scriptsize \cellcolor{black!95}\color{black}{\textcolor{white}{  $99.2$}}  & \scriptsize\cellcolor{black!95}\color{black}{\textcolor{white}{  $99.7$}}
& \scriptsize  \cellcolor{black!95}\color{black}{\textcolor{white}{ $99.1(\diamond)$}} & \scriptsize $69.7$ & \scriptsize $67.2$  & \scriptsize $62.9$ \\
\cline{2-4}\cline{6-20}
&\scriptsize  VGGM  \cite{Cimpoi2016deep} & \scriptsize IJCV 2016& \scriptsize Section \ref{sec:Pretrained}& \scriptsize  &\scriptsize GMM &\scriptsize IFV &\scriptsize Linear SVM &\scriptsize  $65536$
& \scriptsize $72.8(\diamond)$ & \scriptsize $77.5(\diamond)$ & \scriptsize $98.6(\diamond)$
& \scriptsize $98.7$ &   \scriptsize\cellcolor{black!95}\color{black}{\textcolor{white}{  $99.8$}} & \scriptsize  \cellcolor{black!95}\color{black}{\textcolor{white}{  $99.6$}} &  \scriptsize\cellcolor{black!95}\color{black}{\textcolor{white}{  $99.9$}}
&  \scriptsize\cellcolor{black!95}\color{black}{\textcolor{white}{  $99.4(\diamond)$}}& \scriptsize $73.3$ & \scriptsize $73.5$  & \scriptsize $66.8$ \\
\cline{2-4}\cline{6-20}
&\scriptsize  VGGVD  \cite{Cimpoi2016deep} & \scriptsize IJCV 2016& \scriptsize Section \ref{sec:Pretrained}& \scriptsize &\scriptsize GMM &\scriptsize IFV &\scriptsize Linear SVM &\scriptsize  $65536$
& \scriptsize $80.0(\diamond)$  & \scriptsize $82.3(\diamond)$ & \scriptsize \cellcolor{black!95}\color{black}{\textcolor{white}{ $98.7(\diamond)$}}
& \scriptsize\cellcolor{black!95}\color{black}{\textcolor{white}{  $99.0$}} & \scriptsize\cellcolor{black!95}\color{black}{\textcolor{white}{ $99.8$ }}& \scriptsize \cellcolor{black!95}\color{black}{\textcolor{white}{  $99.9$}} & \scriptsize \cellcolor{black!95}\color{black}{\textcolor{white}{ $99.9$}}
& \scriptsize \cellcolor{black!95}\color{black}{\textcolor{white}{ $99.5(\diamond)$} }& \scriptsize
$81.8$ & \scriptsize $79.8$  & \scriptsize $72.3$ \\
\cline{2-20}
&\scriptsize  VGGVD  \cite{Cimpoi2016deep} & \scriptsize IJCV 2016& \scriptsize Section \ref{sec:Pretrained}& \scriptsize \shortstack [c] {Pretrained \\ FC features} &\scriptsize $-$ &\scriptsize $-$ &\scriptsize Linear SVM &\scriptsize  $4096$
& \scriptsize $-$  & \scriptsize $-$ & \scriptsize$-$
& \scriptsize $94.5$ & \scriptsize$97.9$ & \scriptsize  $97.0$ & \scriptsize $97.7$& \scriptsize $-$ & \scriptsize
$75.4$ & \scriptsize $77.4$  & \scriptsize $62.9$ \\
\cline{2-20}
&\scriptsize   BCNN \cite{lin2016visualizing} & \scriptsize CVPR 2016& \scriptsize Section \ref{sec:Pretrained}& \scriptsize \shortstack [c] {VGGVG\\CONV features} & \multicolumn{2}{|c|}{\scriptsize  Bilinear Pooling}& \scriptsize Linear SVM & \scriptsize $262144$
& \scriptsize $-$  & \scriptsize$-$ & \scriptsize$-$
& \scriptsize  $-$ & \scriptsize$-$ & \scriptsize $-$  & \scriptsize $-$
& \scriptsize $-$ & \scriptsize $77.9$ & \scriptsize $81.6$  & \scriptsize $72.9$ \\
\cline{2-20}
&\scriptsize   LFVCNN  \cite{Song2017Locally}  & \scriptsize CVPR 2017 & \scriptsize Section \ref{sec:Pretrained} & \scriptsize \shortstack [c] {VGGVG\\CONV features} &\scriptsize GMM &\scriptsize IFV &\scriptsize LFV classifier&\scriptsize $65536$ & \scriptsize $-$   & \scriptsize $-$   & \scriptsize$-$
& \scriptsize  $-$  & \scriptsize $-$ & \scriptsize $-$   & \scriptsize$-$ & \scriptsize $-$  &  \cellcolor{black!95} \scriptsize \color{black}{\textcolor{white}{ $82.6$}}
  & \cellcolor{black!95}\scriptsize \color{black}{\textcolor{white}{ $82.1$} } & \cellcolor{black!95}\scriptsize \color{black}{\textcolor{white}{ $73.8$ }} \\
\cline{2-20}
\cline{2-20}
&\scriptsize   ResNet   \cite{Zhang2017Deep}  & \scriptsize CVPR2017 & \scriptsize Section \ref{sec:Pretrained} & \scriptsize ResNet50&\scriptsize GMM &\scriptsize IFV &\scriptsize Linear SVM &\scriptsize $65536$
& \scriptsize $-$  & \scriptsize  $-$ & \scriptsize$-$
& \scriptsize  $-$ & \scriptsize$-$ & \scriptsize $-$  & \scriptsize $-$& \scriptsize
& \scriptsize $-$ & \scriptsize  $78.2$ & \scriptsize $-$ \\
\cline{2-20}
&\scriptsize   TCNN \cite{Andrearczyk2016using}   & \scriptsize PRL 2017  & \scriptsize Section \ref{subsec:Finetuned} & \scriptsize \shortstack [c] {AlexNet\\CONV layers} &\multicolumn{3}{|c|}{\scriptsize  Global Average Pooling, FC, SoftMax}  &\scriptsize $4096(\dagger)$
& \scriptsize  $-$ & \scriptsize  $-$ & \scriptsize$-$
& \scriptsize \cellcolor{black!95}\color{black}{\textcolor{white}{ $99.5$  } }& \scriptsize$-$ & \scriptsize  $-$ & \scriptsize $-$
& \scriptsize $-$ & \scriptsize$73.2$ & \scriptsize $-$  & \scriptsize $55.8$ \\
\cline{2-20}
&\scriptsize  \shortstack [c] {Compact\\BCNN} \cite{Gao2016Compact} & \scriptsize CVPR 2016    & \scriptsize Section \ref{subsec:Finetuned} & \scriptsize \shortstack [c] {VGGVD\\CONV layers}  &\multicolumn{3}{|c|}{\scriptsize  Compact Bilinear Pooling (CBP), SoftMax} &\scriptsize $8192(\dagger)$
& \scriptsize  $-$ & \scriptsize  $-$ & \scriptsize $-$
& \scriptsize $-$  & \scriptsize  $-$ & \scriptsize  $-$ & \scriptsize $-$ & \scriptsize $-$
& \scriptsize $-$ & \scriptsize  $-$ & \scriptsize $67.7$ \\
\cline{2-20}
&\scriptsize   FASON \cite{Dai2017FASON}   & \scriptsize CVPR 2017 & \scriptsize Section \ref{subsec:Finetuned} & \scriptsize  \shortstack [c] {VGGVD\\CONV layers} &\multicolumn{3}{|c|}{\scriptsize  CBP and Global Ave. Pooling, SoftMax} &\scriptsize $9216(\dagger)$
& \scriptsize $-$  & \scriptsize $-$  & \scriptsize$-$
& \scriptsize $-$  & \scriptsize $-$ & \scriptsize $-$   & \scriptsize $-$
& \scriptsize $-$ & \scriptsize $76.4$ & \scriptsize $-$  & \scriptsize $72.9$  \\
\cline{2-20}
&\scriptsize   DeepTEN   \cite{Zhang2017Deep} & \scriptsize CVPR 2017 & \scriptsize Section \ref{subsec:Finetuned} & \scriptsize ResNet50 &\multicolumn{3}{|c|}{\scriptsize  Texture Encoding Layer, SoftMax}&\scriptsize $4096(\dagger)$
& \scriptsize  $-$ & \scriptsize $-$  & \scriptsize$-$
& \scriptsize  $-$ & \scriptsize $-$& \scriptsize $-$  & \scriptsize$-$ & \scriptsize $-$
& \scriptsize $-$ & \scriptsize $80.2$  & \scriptsize $-$ \\
\Xhline{1.5pt}
\end{tabular}
}
\end{minipage}
\end{sideways}
\end{table*}

\section{Discussion and Conclusion}
\label{sec:Discussion}
The importance of texture representations lies in the fact that
they have extended to many different problems beyond that of textures themselves.
As a comprehensive survey on texture representations,
this paper has highlighted the recent achievements, provided some structural
categories for the methods according to their roles in feature representation, analyzed their merits and demerits,
summarized existing popular texture datasets, and discussed performance for the most representative
approaches. Almost any practical application is a compromise among conflicting requirements such as classification accuracy, robustness to image degradations, compactness and efficiency,
number of training data available, and cost and power consumption of
implementations. Although significant progress has been made, the following discussion identifies a number of promising directions for exploratory research.

\emph{Large Scale Texture Dataset Collection.}
The constantly increasing volume of image and video data creates new opportunities
and challenges. The complex variability of big image data reveals the
inadequacies of conventional handcrafted texture descriptors and brings opportunities for representation
learning techniques, such as deep learning, which aim at learning good representations automatically from data.
The recent success of deep learning in image classification and object recognition is
inseparable from the availability of large-scale annotated image datasets
such as ImageNet \cite{russakovsky2015imagenet} and MS COCO \cite{Lin2014Microsoft}.
However, deep learning based texture analysis has not kept pace with the rapid progress witnessed
in other fields, partially due to the
unavailability of a large-scale texture database.  As a result there is significant motivation for a good, large-scale texture dataset, which will significantly advance texture analysis.

\emph{More Effective and Robust Texture Representations.}
Despite significant progress in recent years most texture descriptors,
irrespective of whether handcrafted or learned, have not
been capable of performing at a level sufficient for real world textures.
The ultimate goal of the community is to develop texture representations that can accurately
and robustly discriminate massive image texture categories in all possible scenes, at a level comparable to the human visual system.
In practical applications, factors such as significant changes in illumination, rotation, viewpoint and scale, and
image degradations such as occlusions, image blur and random noise call for
more discriminative and robust texture representations.
Further input from psychological
research of visual perception and the biology of the human visual system would be welcome.

\emph{Compact and Efficient Texture Representations.}
There is a tension between the demands of big data and desire for highly compact
and efficient feature representations.  Thus, on the one hand, many existing texture
representations are failing to keep pace with the emerging ``big dimensionality'' \cite{Zhai2014Emerging},
leading to a pressing need for new strategies in dealing with scalability, high computational complexity, and storage.
On the other hand, there is a growing need for deploying highly compact
and resource-efficient feature representations on
platforms like low energy embedded vision sensors and handheld devices.
Many of the existing descriptors would similarly fail in these contexts, and the current general trend of deep CNN
architectures has been to develop deeper and more complicated networks, advances requiring massive data and power hungry GPUs,
not suitable to be deployed on mobile platforms that have limited resources.  As a result, there is a growing
interest in building compact and efficient CNN-based features
\cite{Howard2017MobileNets,Rastegari2016Xnor}. While CNNs generally outperform classical texture descriptors,
it remains to be seen which approaches will be most effective in resource-limited contexts, and whether some degree of
LBP / CNN hybridization might be considered, such as recent lightweight CNN architectures \cite{Lin2017Towards,Juefei2016Local}.

\emph{Reduced Dependence on Large Amounts of Data.}
There are many applications where texture representations are very useful and
 only limited amounts of annotated training data can be available, or where
collecting labeled training data is too expensive (such as visual inspection, facial micro-expression recognition, age
estimation and medical texture analysis).
Possible research could be the development of learnable
local descriptors requiring modest training data,
as in \cite{Duan2017Context,Lu2017Simultaneous}, or to explore effective transfer learning.

\emph{Semantic Texture Attributes.} Progress in image texture representation and understanding, while substantial, has so far been mostly
focused on low-level feature representation. However, in order to address advanced
human-centric applications, such as detailed image search and human-robotic interaction, low-level understanding will not be sufficient.
Future efforts should be devoted to go beyond
 texture identification and categorization, to develop semantic and easily describable texture attributes
 that can be well predicted with low-level texture representations, and to explore even
 fine-grained and compositional structure analysis of texture patterns.

\emph{Effect of Smaller Image Size.} Performance evaluation of texture descriptors is
usually done with texture datasets consisting of relatively large images.
For a large number of applications an ability to analyze small image sizes
at high speed is vital, including facial image analysis, interest region description, segmentation,
defect detection, and tracking. Many existing texture descriptors would fail in this respect, and it would be
important to evaluate the performance of new descriptors \cite{Schwartz2015}.

\section{Acknowledgments}
The authors would like to thank the pioneer researchers in
texture analysis and other related fields. The authors would also like to express their sincere appreciation to the associate editor and the reviewers for their comments and suggestions. This work has been supported by the Center for Machine Vision and Signal Analysis at the University of Oulu (Finland) and the National Natural Science Foundation of China under Grant 61872379.

\bibliographystyle{spbasic}      
\footnotesize
\bibliography{manuscriptbib}

\end{document}